%% file: main.tex
\documentclass[11pt, a4paper, logo]{lumia}

\usepackage[numbers,sort&compress]{natbib}
\definecolor{infrecred}{HTML}{D2071E}
\definecolor{abstrabg}{HTML}{FCDDDC}
\colorlet{abstradark}{infrecred}
\definecolor{reportaccent}{HTML}{BB0819}
\colorlet{reporttoc}{reportaccent}
\definecolor{cayenne}{HTML}{941100}
\hypersetup{allcolors=cayenne}
\definecolor{DarkRed}{HTML}{8B1E2D}
\definecolor{DarkTeal}{HTML}{236F6C}

\newcommand{\groupheading}[1]{%
    \textit{\textcolor{DarkRed}{#1}}%
}               %
\definecolor{lightblue}{HTML}{D0CFEA}

\input{notations}

\graphicspath{{fig/}{appendix/figures/}}
\usepackage{xurl}
\usepackage{amsfonts}
\usepackage{amsmath}
\usepackage{amssymb}
\usepackage{nicefrac}
\usepackage{microtype}
\usepackage{pdfpages}
\usepackage{makecell}
\usepackage{wrapfig}
\usepackage{float} % 这个包支持 [H] 强制原位插入
\usepackage{algorithm}
\usepackage{algorithmic}
\usepackage{multirow}
\usepackage{subcaption}
\usepackage{tikz}
\usetikzlibrary{calc,positioning,shapes,arrows.meta,fit,backgrounds}
\usepackage{xcolor}
\usepackage{graphicx}
\usepackage{booktabs}
\usepackage{adjustbox}
\usepackage{wrapfig}
\usepackage{adjustbox}
\usepackage{booktabs}
\usepackage{caption}
\usepackage{placeins}
\usepackage{longtable}
\newlength{\arxiviconheight}
\setheadertext{The Trinity}

\title{%
  \raisebox{-0.24\height}[0pt][0pt]{%
  }\hspace{-1pt}The Trinity:\\
  {\fontsize{14}{20}\selectfont
   Capability-Driven Recursive Self-Improvement for Multimodal Agents\par}%
}

\setheadertitle{The Trinity: Capability-Driven Recursive Self-Improvement for Multimodal Agents}

\author{%
  \shortstack[c]{%
    {\fontsize{13pt}{15pt}\selectfont
      The Trinity Research Team}%
    \\
    {\fontsize{10pt}{12pt}\selectfont
      \faGlobe\enspace
      \href{https://trinity-rsi.github.io/}{Project Page}
      \qquad
      \faGithub\enspace
      \href{https://github.com/InfRec-Cardinal/The-Trinity}{GitHub}
    }%
  }%
}

\begin{document}

\begin{abstract}
\input{sections/0_abstract}
\end{abstract}

\teaserfigure{%
\vskip4pt
\centering
\includegraphics[width=1.0\linewidth]{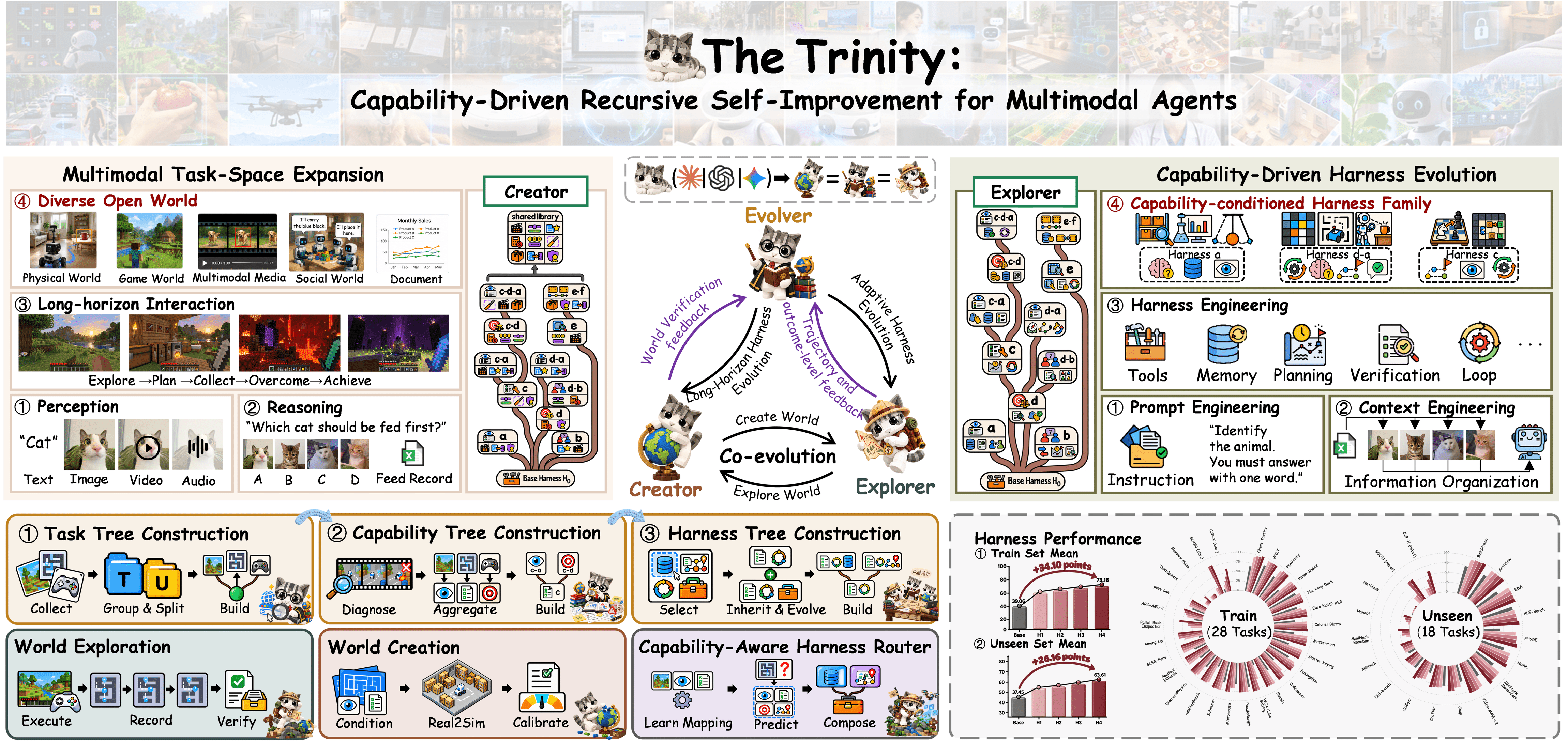}
{\captionsetup{font=small}
\captionof{figure}{\textbf{Overview of The Trinity.}
Creator, Explorer, and Evolver connect multimodal task-space expansion (left) with capability-driven Harness evolution (right) through verified feedback.
The lower panels summarize tree construction, world exploration and creation, and capability-aware routing; the charts report mean and benchmark-level Train/Unseen performance across RSI iterations.}
\label{fig:intro}}
\vskip4pt}

\maketitle

% A dedicated, linked contents page immediately after the cover.
\clearpage
\begingroup
\hypersetup{linkcolor=reporttoc,linktoc=section}
\setlength{\parskip}{0pt}
\setcounter{tocdepth}{3}
\tableofcontents
\endgroup
\clearpage

\input{sections/1_intro}
\input{sections/2_preliminaries}
\clearpage
\input{sections/3_method}

\input{sections/4_exp}

\input{sections/5_related_work}
\input{sections/6_sum}
\input{sections/7_diss}
\input{sections/8_authors}

\phantomsection
\addcontentsline{toc}{section}{References}
\bibliography{references}

\appendix
\input{sections/appendix}

\end{document}

%% file: notations.tex
\usepackage{xspace}

\theoremstyle{plain}

\newtheorem*{proposition*}{Proposition}

\theoremstyle{definition}

\theoremstyle{definition}

\def\eqref#1{equation~\ref{#1}}

\usepackage{xcolor}
\providecolor{reportaccent}{HTML}{BB0819}

\newcommand{\TrinityHarness}{\mathcal{H}}

\newcommand{\TrinityInitialHarness}{\TrinityHarness_0}

\newcommand{\TrinityNodeHarness}[2]{%
  \TrinityHarness_{\mathrm E,#1}^{(#2)}%
}

\newcommand{\TrinityHarnessTree}[1]{%
  \TrinityHarness^{(#1)}%
}

\newcommand{\TrinityCreatorHarness}[1]{%
  \TrinityHarness_{\mathrm{C}}^{(#1)}%
}

\newcommand{\TrinityCapabilityTree}{\mathcal{T}}

\newcommand{\TrinityTrainingTasks}{\mathcal{B}_{\mathrm{tr}}}

\newcommand{\TrinityUnseenTasks}{\mathcal{B}_{\mathrm{un}}}

\newcommand{\TrinityCapabilityVocabulary}{\mathcal{K}}

\newcommand{\TrinityTaskRequirements}[1]{K_{#1}}

\newcommand{\TrinityTaskPath}[1]{p_{#1}}

\newcommand{\TrinityCallout}[5][6pt]{%
  \par\smallskip
  \begingroup
  \setlength{\fboxsep}{0pt}%
  \noindent\colorbox[RGB]{#2}{%
    \hbox{%
      {\color[RGB]{#3}\vrule width 2.8pt}%
      \hspace{9.5pt}%
      \begin{minipage}{\dimexpr\linewidth-22.2pt\relax}
        \vspace{#1}%
        \normalfont\normalsize
        \textcolor[RGB]{#4}{#5}%
        \par\vspace{#1}%
      \end{minipage}%
      \hspace{9.5pt}%
      {\color[RGB]{#3}\vrule width 0.4pt}%
    }%
  }%
  \endgroup
  \par\smallskip
}

\newcommand{\TrinityPosition}[1]{%
  \TrinityCallout[7.6341pt]
    {247,228,234}{139,30,45}{0,0,0}%
    {\textcolor{DarkRed}{\textbf{Position.}} #1}%
}

\newcommand{\TrinityFinding}[2]{%
  \TrinityCallout
    {255,237,231}{139,30,45}{0,0,0}%
    {\textbf{#1} #2}%
}

%% file: sections/0_abstract.tex
\looseness=-1
Multimodal intelligence can be viewed as human-guided co-evolution between worlds and agents through interaction.
However, scaling this process faces two bottlenecks: constructing heterogeneous, verifiable environments is costly, and manual Harness iteration cannot keep pace with growing task diversity and complexity.
A key insight is that tasks with different modalities and surface forms can share the same underlying capability requirements.
We therefore introduce \textbf{The Trinity}, a recursive self-improvement framework coupling world generation with Harness evolution through three roles: Creator, Explorer, and Evolver.
Drawing on verified execution trajectories, Evolver shifts the basis of task organization from surface categories to shared requirements, reorganizing the original task tree into a hierarchical capability tree.
We introduce \textbf{Capability-Driven Real2Sim2Real}: for heterogeneous tasks at the same node, Creator abstracts their common requirements into simplified, verifiable simulated worlds for capability learning, while evaluation on original tasks measures transfer.
Along this capability tree, Explorer progressively evolves reusable capability modules through selective inheritance, forming a Harness tree whose validated updates accumulate across iterations within a fixed topology.
The resulting Harnesses specialize in different capabilities, motivating a \textbf{Capability-Aware Harness Router} that uses public task information to select compatible module configurations from the tree.
Our evolved Harnesses improve mean performance by 34 points across 28 training benchmarks and 26 points across 18 unseen benchmarks, including two real-world embodied tasks.

%% file: sections/1_intro.tex
\section{Introduction}
\label{sec:introduction}

\looseness=-1
Multimodal foundation models have rapidly advanced in understanding, reasoning, and generation across text, images, video, and audio~\citep{alayrac2022flamingo,achiam2023gpt,hurst2024gpt,team2023gemini,liu2023visual,bai2025qwen3,singh2025openai,Tang_2026_CVPR,liu2026all,tang2026varscaling,li2026coco,li2026videococo}, enabling agents to perceive, decide, and act in open environments~\citep{brohan2023rt,driess2023palm,liu2026world}. 
We view multimodal intelligence as an evolving loop: worlds provide tasks, humans guide learning, and agents interact with the world to produce verifiable outcomes~\cite{shi2026codehacker,zhang2026valg}. 
As tasks and environments grow increasingly complex, this loop faces two bottlenecks:
(1) Constructing large-scale, heterogeneous, and verifiable environments manually is costly, motivating automated world generation; and
(2) Agent capability is increasingly constrained by the Harness~\citep{lee2026meta,zhang2026self,lin2026agentic,lu2026muse,lee2026recursive,wu2026harnessdev}, whose manual iteration cannot keep pace with growing task complexity and diversity.

Inspired by Harness recursive self-improvement~\citep{zhang2026darwin,zhang2026jit,wu2026modularrsi} and agent--environment co-evolution~\citep{liu2026spade,wang2025embodied,yuan2026pace}, we ask whether this manual loop can improve continuously.
Generated worlds provide adaptive curricula, while verified interaction feedback guides Harness updates.

\looseness=-1
However, extending this process across heterogeneous text-only and multimodal tasks~\cite{cui2026earthverse,tang2026grouptom} presents a challenge. Current Harness RSI works focus on text-modality tasks (e.g., coding and math~\citep{zhang2026darwin,wu2026harnessdev,chen2026failed,lou2026autoharness}) and adopt task-specific evolution~\citep{zhou2026hierarchical,lee2026recursive,shao2026harness,zhang2026jit}. 
This approach is suitable when tasks share similar capability requirements, but can lead to Harness overfitting and poor generalization across heterogeneous tasks. 
Conversely, a global Harness struggles to accommodate conflicting task demands, including performance–efficiency trade-offs in embodied tasks, limiting its efficiency and adaptability.

\looseness=-1
A key insight is that tasks with different modalities and surface forms can share the same underlying capability requirements.
For example, although ARC-AGI-3~\citep{arcprize2026arcagi3} and Micromouse~\citep{ukmarsMicromouseRules} differ in task structure, both rely on accurate perception and state tracking: the former requires recognizing grid states and their changes, while the latter requires continuously updating environmental information during maze exploration. 
Grouping tasks by these shared capability requirements makes the underlying mechanisms a common learning objective, enabling generalization across tasks.

\TrinityPosition{Current RSI training relies heavily on task-specific optimization, limiting capability generalization across tasks. We advocate a shift from task-driven to capability-driven self-improvement, where shared capability requirements across tasks guide the learning and optimization of multiple specialized Harnesses. This shift highlights the role of shared capabilities in cross-task generalization and naturally motivates a capability-aware Router that matches tasks to suitable Harnesses based on their capability requirements.}

\looseness=-1
To operationalize this position, we introduce \textbf{The Trinity}, a recursive self-improvement framework coupling environment generation with agent evolution through Creator, Explorer, and Evolver.
Using verified trajectories, Evolver identifies task capability requirements and groups tasks sharing them, reorganizing the original task tree into a hierarchical capability tree.
Heterogeneous tasks at the same node share a learning objective, motivating \textbf{Capability-Driven Real2Sim2Real}: Creator abstracts their common requirements into simplified, verifiable environments, while evaluation on the original tasks measures capability transfer.
Along the capability tree, Explorer evolves a Harness tree through selective inheritance and local refinement, while Creator evolves at corresponding nodes and consolidates validated updates into a new Base Harness at the end of each iteration.

The Harnesses specialize in different capabilities, requiring task-dependent selection.
For Explorer, our \textbf{Capability-Aware Harness Router} uses verified execution experience and public task information to select a matching branch of Explorer's Harness tree.
For Creator, overlap within its Harness tree motivates merging complementary modules and removing near-duplicates to form a new Base Harness.
Explorer and Creator play distinct roles in RSI processes along the same capability tree.

\looseness=-1
\noindent\textbf{Contributions.}
\ding{182} \underline{\textbf{\textit{Capability-driven multimodal RSI.}}}
We make shared capability requirements the organizing principle of Harness evolution. Capability-Driven Real2Sim2Real turns these requirements into compact training worlds, yielding a reusable module-based Harness tree whose specialization motivates capability-aware routing.
\ding{183} \underline{\textbf{\textit{Capability--module priors for Harness design.}}}
We connect task-level capability requirements to reusable module configurations, using empirical associations and controlled interventions to examine which module support matters for a task.
\ding{184} \underline{\textbf{\textit{Efficient evolution and transfer.}}}
We show that targeting shared decision logic supports effective Harness improvement at low evolution cost, with transfer across heterogeneous tasks, unseen benchmarks, and real-world embodied settings.

%% file: sections/2_preliminaries.tex
% Keep the original local spacing while allowing normal pagination.
% The surrounding template's body font and margins are unchanged.
\begingroup
\setlength{\parskip}{0.3\baselineskip}
\setlength{\abovedisplayskip}{6pt}
\setlength{\belowdisplayskip}{6pt}
\setlength{\abovedisplayshortskip}{4pt}
\setlength{\belowdisplayshortskip}{4pt}
\section{Preliminaries}
\subsection{Multimodal Agentic Tasks}
\label{sec:formulation}
\looseness=-1
Long-horizon multimodal agentic tasks span interactive games~\citep{paglieri2025balrog}, GUI manipulation~\citep{xie2024osworld}, and embodied interaction~\citep{brohan2023rt,driess2023palm}. They combine perception, understanding, reasoning, and generation with memory, planning, and adaptation across observation--action sequences. Agents must retain evidence, revise hypotheses, and adjust plans as observations arrive. Because later decisions depend on earlier actions and feedback, misread observations or forgotten constraints can propagate across steps. Agent performance thus increasingly depends on effective Harness organization and coordination. Dynamically coordinating capabilities over long interactions and adapting to increasingly large, heterogeneous task sets remain key challenges.

\subsection{Agent Harness}
\label{sec:harness_formalization}
A Harness configuration $\phi$ coordinates a frozen model $f_\theta$ through six control-flow protocols:
\begin{equation}
\TrinityHarness_\phi = (\Pi_{\mathrm{er}},\mathbf{M},\mathbf{I},\mathbf{P},\mathbf{F},\mathbf{L}),
\qquad
\TrinityInitialHarness\equiv\TrinityHarness_{\phi_0}\quad\text{(Base)}.
\end{equation}
Given task $\tau$, observation $o_t$, public history $\xi_{<t}$, controller state $s_t$, and available resources $\mathcal{C}_\tau$:
\begin{equation}
\begin{aligned}
f_t &= \Pi_{\mathrm{er}}(o_t,\xi_{<t}),
& v_t &= \mathbf{M}(\xi_{<t},s_t,f_t),\\
\mathrm{Cand}_t &= \mathbf{I}(\tau,v_t),
& d_t &= \mathbf{P}(\tau,s_t,v_t,\mathrm{Cand}_t),\\
\mathcal{C}_t &= \mathbf{F}(\mathcal{C}_\tau,s_t,v_t,d_t),
& (s_{t+1},e_t,c_t) &= \mathbf{L}(s_t,v_t,d_t,\mathcal{C}_t).
\end{aligned}
\end{equation}
\looseness=-1
\textbf{Perception} extracts facts $f_t$; \textbf{Memory} builds a history-conditioned view $v_t$; \textbf{Inference} maintains compatible candidates; \textbf{Planning} ranks or prunes options into directive $d_t$; \textbf{Orchestration} identifies applicable resources; and \textbf{Loop} resolves execution $e_t$ and commitment $c_t$, including withholding action.
These roles are independent of task categories and capability-tree nodes. The equations describe functional roles and their data dependencies; individual module bindings are invoked under the runtime policy described in Section~\ref{sec:local_harness_evolution}.

\looseness=-1
Modules are pure functions of supplied public records, without hidden-state access or environment actions.
Only Memory may declare explicit \texttt{carry}; hypothesis refutation requires observed contradiction.
All protocols may emit typed readiness signals, while Loop resolves execution.
Orchestration assesses applicability, relevance, and trust within a fixed host-exposed tool set.
Evolution updates module configurations with $\theta$ fixed.

\looseness=-1
Capabilities and protocols have a many-to-many relationship: the tree groups shared requirements, while protocols locate module updates. Hypothesis consistency can involve Memory organizing evidence, Inference checking compatibility, and Planning ranking informative tests; the same Memory module can support several requirements. Explorer retains node modules; Creator uses the same tree structure and consolidates its next Base Harness. Evaluation-time skills are prepared separately (Section~\ref{sec:skill_preparation}); they are not stored, evolved, inherited, or routed in the tree.
\par
\endgroup

%% file: sections/3_method.tex
\section{The Trinity}
\label{sec:method}

\begin{figure}[htbp]
    \centering
    \includegraphics[width=1.0\linewidth]{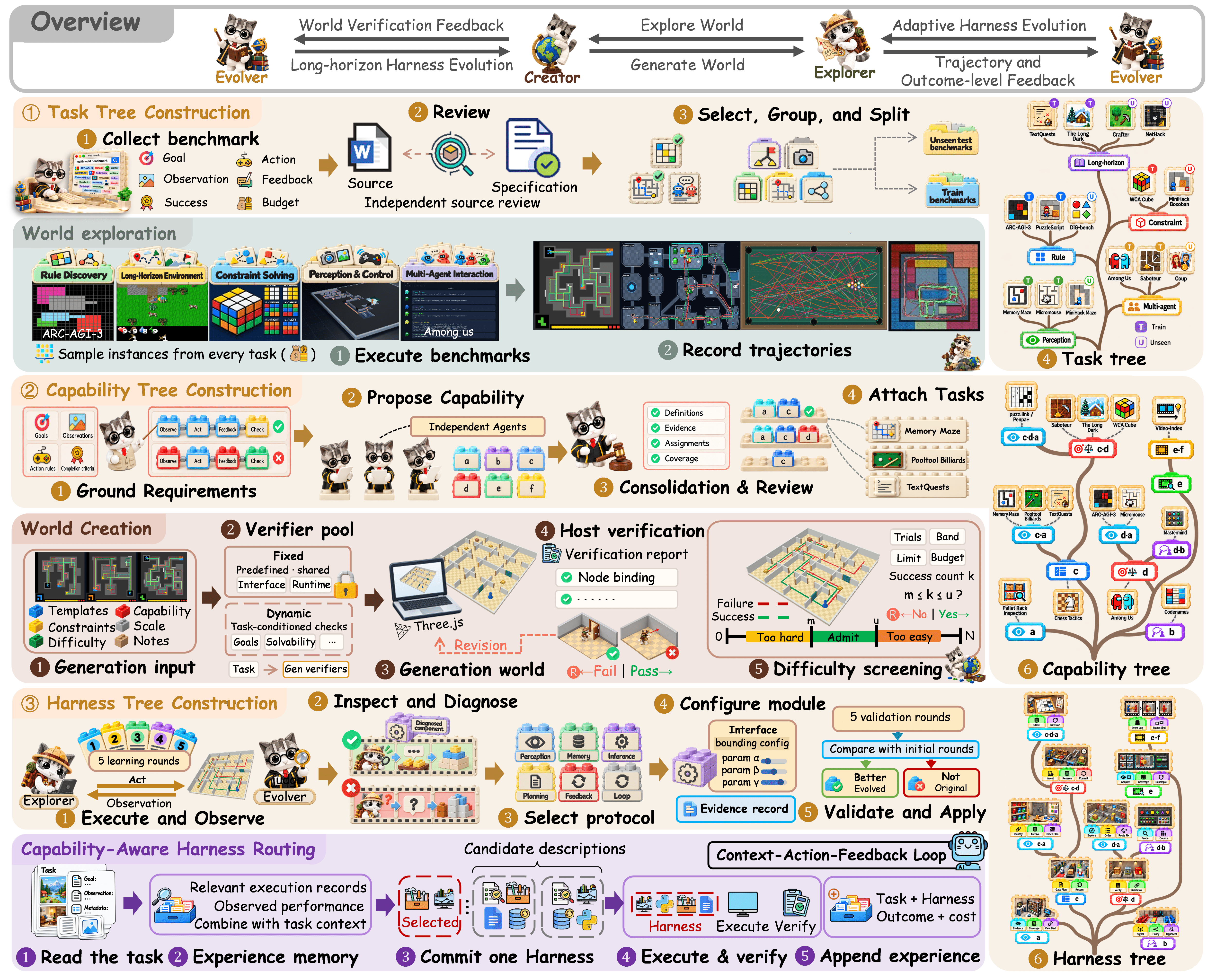}
    \caption{\textbf{Detailed pipeline of The Trinity.}
    Three stages build task, capability, and Harness trees.
    Worlds are verified and screened; Harnesses evolve through selective inheritance and local refinement.
    Routing selects one compatible Harness and stores verified feedback for later tasks.}
    \label{fig:pipeline}
\end{figure}

\subsection{Framework Overview}
\label{sec:framework_overview}
The Trinity couples environment generation and Harness evolution with a frozen backbone (Figure~\ref{fig:pipeline}). Explorer collects trajectories; Evolver organizes requirements from specifications and verified trajectories into a capability tree (Sections~\ref{sec:task_tree}--\ref{sec:capability_discovery}).
Through \textbf{Capability-Driven Real2Sim2Real}, Creator builds simplified, verifiable worlds; original-task evaluation measures transfer (Section~\ref{sec:world_creation}). Explorer selectively inherits and refines reusable modules, accumulating validated updates in a fixed-topology Harness tree (Sections~\ref{sec:harness_evolution} and~\ref{sec:rsi_method}). The \textbf{Capability-Aware Harness Router} selects a compatible branch using public task information and verified experience (Section~\ref{sec:router_method}).

\subsection{Stage 1: Task Tree Construction}
\label{sec:task_tree}
\paragraph{Benchmark Organization.}
Evolver normalizes public multimodal benchmarks into specifications of goals, observations, actions, feedback, task-completion criteria, and budgets, independently checked against their sources. A common observation--action interface preserves information access and scoring rules. Benchmarks are grouped into five requirement domains: rule discovery, constrained problem solving, perception and control, multi-agent interaction, and long-horizon environments. The training/unseen split is fixed before capability induction.

\paragraph{World Exploration.}
\label{sec:h0_exploration}
Explorer collects trajectories from sampled training instances using $\TrinityInitialHarness$ under fixed settings. Only training records support capability-tree construction, world generation, module learning, and Router training (Section~\ref{sec:router_method}). Unseen capability requirements are assessed only after the tree is fixed, for evaluation. Instance selectors and initial-state hashes are fixed in advance. Diagnostic reruns reset memory; completed runs are retained, while infrastructure failures are marked invalid and retried. Hash-chain checks verify execution records.

\subsection{Stage 2: Capability Tree Construction}
\label{sec:capability_discovery}
\label{sec:capability_tree}
Evolver derives requirements from task goals, permitted observations, action rules, and task-completion criteria, linking them to rules and trajectory steps satisfying or violating these criteria through rule-based checks. Host-private information supports only retrospective verification.

\paragraph{Requirement Organization.}
Using training-task specifications and trajectories, independent agents propose \textbf{capability categories} $\TrinityCapabilityVocabulary$ (Appendix~\ref{app:capability-tree}); a judge consolidates them, and reviewers independently check definitions, evidence, assignments, and coverage. Each training task $\tau$ receives a fixed requirement set $\TrinityTaskRequirements{\tau}\subseteq\TrinityCapabilityVocabulary$, including requirements already satisfied by $\TrinityInitialHarness$.

The tree is constructed from training tasks $\TrinityTrainingTasks$ only. A deterministic capability order $\prec$, fixed before construction, defines paths and the tree:
\begin{equation}
\TrinityTaskPath{\tau}=\operatorname{sort}_{\prec}(\TrinityTaskRequirements{\tau}),
\qquad
\TrinityCapabilityTree=\operatorname{Trie}\bigl(\{\TrinityTaskPath{\tau}:\tau\in\TrinityTrainingTasks\}\bigr).
\label{eq:capability_tree_construction}
\end{equation}
The $\operatorname{Trie}$ merges \textbf{common prefixes} across domains. Each node $v$ supports an Explorer module configuration for its prefix; tasks retain endpoint associations, including at internal nodes. Node requirements define local objectives, with selective parent inheritance for Explorer; the order implies no causal dependencies. Requirements for unseen tasks $\tau\in\TrinityUnseenTasks$ are identified afterward using the fixed categories, solely for retrospective Router evaluation, without modifying the tree.

\begin{wrapfigure}[16]{r}{0.44\textwidth}
% Align the caption's last baseline with the adjacent body text.
\vspace{-17.08pt}
\centering
\captionsetup{font=normalsize,justification=raggedright,singlelinecheck=false,skip=4pt}
\includegraphics[width=\linewidth]{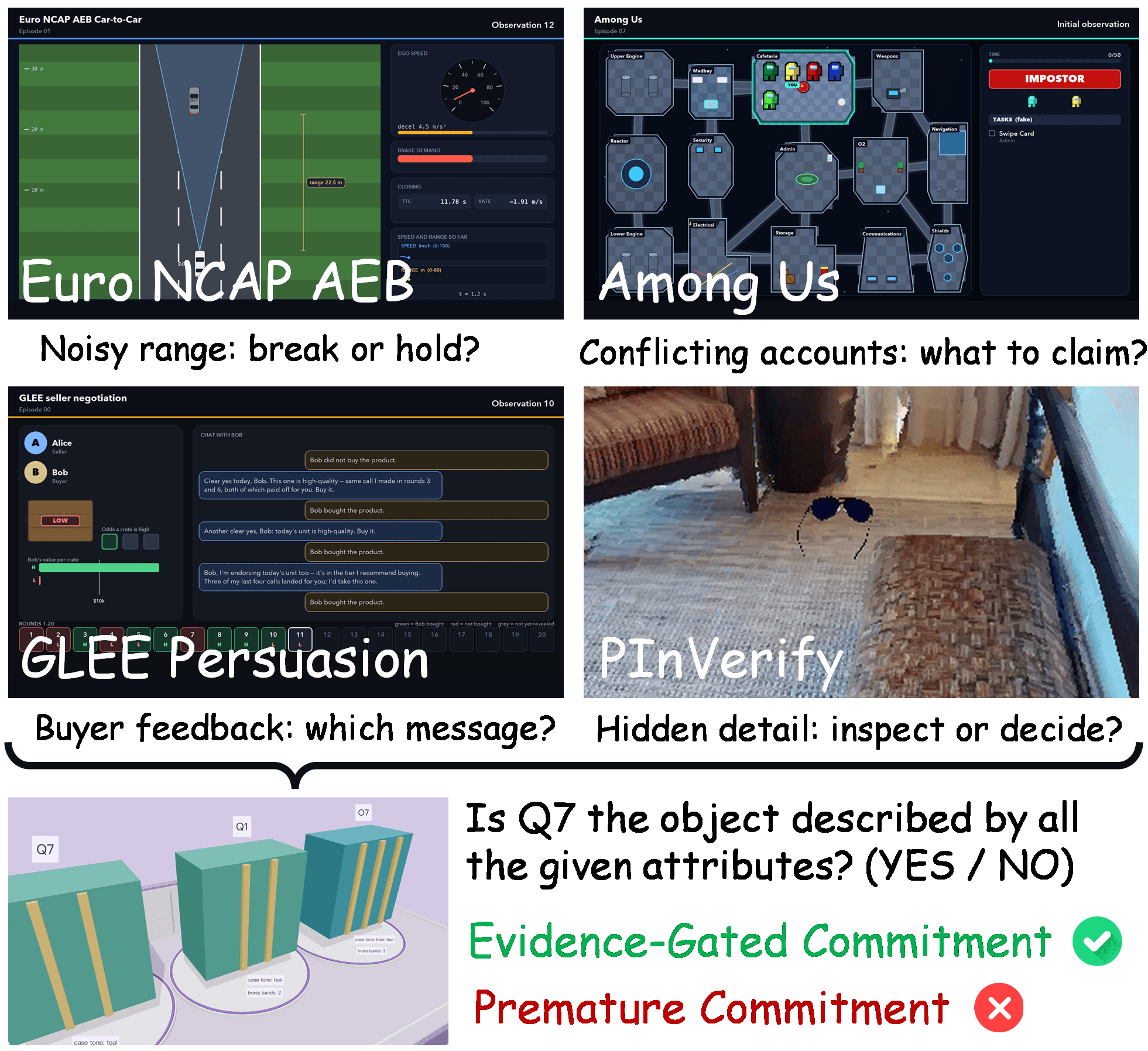}
\caption{\textbf{Capability-driven world creation} at node~d.}
\label{fig:world_creation}
\vspace{-6pt}
\end{wrapfigure}
\paragraph{World Creation.}
\label{sec:world_creation}
\looseness=-1
Once the capability tree is built, each node is linked to training benchmarks through shared requirements. Although the four tasks in Figure~\ref{fig:world_creation} differ, all require evidence verification before decisions. This motivates \textbf{Capability-Driven Real2Sim2Real}: Evolver designs simplified training levels from node requirements, descendants, and representative tasks' public specifications, excluding target-task layouts, private proofs, and solution traces while preserving core decision processes.
Creator implements these levels using a node Harness initialized from its Base Harness or inherited parent modules, omitting aesthetic optimization to focus on task mechanisms. Evolver's host-executed verifier pool checks node binding, execution, state transitions, rendering, and completion. Node identifiers link specifications, trajectories, and module updates, while private verifiers and public feedback guide revisions.
Explorer tests verified worlds with a frozen screening Harness, admitting only those within a predefined intermediate-performance band. Otherwise, Creator adjusts difficulty along predefined dimensions. Trial counts, performance thresholds, action budgets, and retry limits are fixed in advance; worlds and verifiers remain unchanged during local training. Curricula may be reused or regenerated, while original benchmarks assess capability transfer. Appendix~\ref{app:levels} details the curricula for each node.
\subsection{Stage 3: Harness Tree Construction}
\label{sec:harness_evolution}
\label{sec:local_harness_evolution}
For Explorer, capability-driven RSI builds a Harness tree aligned with the capability tree. At each node, Explorer executes admitted worlds with inherited and local modules. Evolver compares successful and failed trajectories using public observations, actions, and host-assigned outcomes to identify node-specific improvements. It develops \textbf{reusable modules} through host-registered bindings in the six-protocol library (Section~\ref{sec:harness_formalization}), refining interfaces, bounded configurations, or implementations. Updates record node, predecessor-version, episode, and executable-example references.
Before local updates, the host validates evidence, ownership, interfaces, lineage, and typed input--output assertions; ancestor modules remain preserved. Bindings use public context and permitted memory under a fixed runtime policy. Multiple bindings per protocol may coexist, with model calls determining execution order. Admission verifies grounding and executability; evaluation measures performance.
Creator evolves only upon verifier-detected failures. At each iteration's end, validated proposals are consolidated with existing Base Harness modules, merging complementary modules and removing near-duplicates. Section~\ref{sec:rsi_method} details persistence, selective inheritance, and consolidation.

\subsection{Recursive Self-Improvement}
\label{sec:rsi_method}
\label{sec:iteration_level_evolution}

Explorer evolves along the capability tree, with Creator following at corresponding nodes; their Harness trees share the same structure (Appendix~\ref{app:harness-tree}). Explorer selectively inherits node modules; at each iteration's end, Creator deduplicates and merges complementary updates into its next Base Harness. Let $V=V(\TrinityCapabilityTree)$ and $V^+$ denote all nodes and non-root nodes, respectively.

\paragraph{Node State.}
\label{sec:explorer_rsi}
For Explorer, let $\TrinityNodeHarness{v}{i}\equiv\TrinityHarness_{\phi_{\mathrm E,v}^{(i)}}$ denote the Harness at node $v\in V$ after iteration $i$. Nodes initialize with $\TrinityNodeHarness{v}{0}=\TrinityInitialHarness$ and available training feedback $\mathcal{F}_v^{(0)}$. Top-down iterations retain local modules and keep the synthetic root fixed.

\paragraph{Selective Inheritance.}
At non-root node $v\in V^+$, compare the previous inherited registry $P_v^{(i)}$ with the updated parent $\widehat{P}_v^{(i+1)}$, holding local modules $L_v^{(i)}$ fixed. Writing the Harness's module configuration as $\TrinityNodeHarness{v}{i}=P_v^{(i)}\oplus L_v^{(i)}$, registries use host-registered binding IDs and right-biased union:
\begin{equation}
(P\oplus L)[m]=
\begin{cases}
L[m], & m\in\operatorname{dom}(L),\\
P[m], & m\in\operatorname{dom}(P)\setminus\operatorname{dom}(L).
\end{cases}
\label{eq:harness_composition}
\end{equation}
Parent exports are bound to parent identifiers and state digests. Ancestor entries compose at recorded versions from root to parent, then local modules; overrides preserve ancestor sources and local ownership/version lineage. Under identical validation tasks, trial count, and budget, the updated parent is retained only with strictly higher host-verified performance, measured by task-completion counts; otherwise, the previous registry is retained.

\paragraph{Continual Refinement.}
Let $P_v^{(i+1)}$ be the retained inherited registry and $\mathcal{D}_v^{(i+1)}$ the admitted-world curriculum conditioned on the fixed requirement, previous feedback $\mathcal{F}_v^{(i)}$, and prior coverage. Curricula may be renewed or reused. Explorer's trajectories $\mathcal{Z}_v^{(i+1)}$ guide Evolver's local update:
\begin{equation}
\begin{aligned}
\mathcal{Z}_v^{(i+1)}
&=\operatorname{Rollout}\!\left(P_v^{(i+1)}\oplus L_v^{(i)},\mathcal{D}_v^{(i+1)}\right),\\
\widetilde{L}_v^{(i+1)}
&=\operatorname{Evolve}\!\left(L_v^{(i)},\mathcal{Z}_v^{(i+1)},\mathcal{F}_v^{(i)}\right).
\end{aligned}
\label{eq:local_refinement}
\end{equation}
Trajectories record public observations, actions, host-assigned outcomes, and executed configurations. The host checks update evidence, ownership, interfaces, predecessor versions, and executable contracts. Accepted updates modify local state; rejection preserves it. The retained $L_v^{(i+1)}$ yields $\TrinityNodeHarness{v}{i+1}=P_v^{(i+1)}\oplus L_v^{(i+1)}$; local admission requires no score increase.

Subsequent training-side evaluation of the retained Harness yields $\mathcal{E}_v^{(i+1)}$ and next-iteration feedback
\begin{equation}
\mathcal{F}_v^{(i+1)}
=\operatorname{Feedback}\!\left(\mathcal{Z}_v^{(i+1)},\mathcal{E}_v^{(i+1)}\right),
\label{eq:iteration_feedback}
\end{equation}
where $\operatorname{Feedback}$ retains verified observations, outcomes, failure evidence, and episode/module-version references for the next iteration. Completed nodes export retained state to children; sibling branches evolve independently. Explorer retains $\TrinityHarnessTree{i}=\{\TrinityNodeHarness{v}{i}:v\in V\}$ for routing.

\paragraph{Creator Base Harness Update.}
\label{sec:creator_rsi}
Creator's Base Harness $\TrinityCreatorHarness{i}$ has module registry $R_{\mathrm C}^{(i)}$, initialized with pre-RSI modules at $i=0$. Creator evolves nodes in top-down order within each iteration. Root children start from this base; other nodes inherit their parent's updated module configuration. Each node retains its own guidance. Verifier-detected failures trigger local evolution, yielding validated proposals $\Delta R_{\mathrm C,v}^{(i+1)}$. At iteration end, proposals and base modules are consolidated in learning order $\prec_{\mathrm C}$:
%Each node's coding Harness starts from this base and its own guidance, without parent--child module inheritance. Verifier-detected failures trigger local evolution, yielding validated proposals $\Delta R_{\mathrm C,v}^{(i+1)}$. At iteration end, proposals and base modules are consolidated in learning order $\prec_{\mathrm C}$:
\begin{equation}
R_{\mathrm C}^{(i+1)}
=\operatorname{Consolidate}_{\prec_{\mathrm C}}\!\left(
R_{\mathrm C}^{(i)}\cup\bigcup_{v\in V^+}\Delta R_{\mathrm C,v}^{(i+1)}
\right).
\label{eq:creator_consolidation}
\end{equation}
The resulting registry configures $\TrinityCreatorHarness{i+1}$, the Base Harness for the next iteration. Validated revisions replace predecessors; distinct capabilities are retained, partial overlaps are merged to preserve complementary functionality, and near-duplicates retain the earlier-learned module. Each binding retains one active definition and its provenance. Guidance, construction notes, and action budgets remain node-specific. Without accepted updates, the Base Harness remains unchanged. Explorer's evaluation-time skills are prepared separately (Section~\ref{sec:skill_preparation}).

\subsection{Capability-Aware Harness Routing}
\label{sec:router_method}
Routing follows a context--action--feedback loop~\citep{zhou2026agentrouter}. Let $x_n$ contain the public goal, initial observation, metadata, and images, and $\mathcal{C}\subseteq\TrinityHarnessTree{i}$ be the fixed, versioned Harness pool. Memory $\mathcal{M}_0$ is initialized by importing verified training-side categorization trajectories into empty memory once in file order (Section~\ref{sec:h0_exploration}). These $\TrinityInitialHarness$ executions record public context, the executed configuration, outcomes, and costs; unseen categorization records are excluded. The agent, prompt, candidate modules, retrieval settings, and verifier remain fixed.

For each task, candidate descriptions, retrieved experience $R$, and observed performance--cost summaries $S$ guide selection:
\begin{equation}
h_n=\pi\!\left(x_n,\mathcal{C},R(x_n,\mathcal{M}_n),S(\mathcal{M}_n)\right).
\label{eq:harness_routing}
\end{equation}
The agent commits $h_n$ with an evidence-based rationale before execution. Retrieval uses cosine similarity $\geq 0.5$, at most ten records, and a 20,000-record FIFO memory. Summaries include only matching executed configurations; untried outcomes remain unknown. Tasks are processed once in file order. After terminal receipts are verified against the committed decision and registry hash, outcome and cost $f_n$ update memory for later tasks:
\begin{equation}
\mathcal{M}_{n+1}=\operatorname{Append}\!\left(\mathcal{M}_n,(x_n,h_n,f_n)\right).
\label{eq:router_memory}
\end{equation}
Unseen task--capability assignments remain excluded from routing context, verification, and memory; they are accessed only for evaluation after the complete prediction ledger is hash-locked. Technical failures remain recorded and score zero; invalid selections receive no fallback.

%% file: sections/4_exp.tex
% Keep the section title with the setup heading and opening paragraph.
\clearpage
\section{Experiments}
\label{sec:experiments}

\subsection{Experimental Setup}
\label{sec:setup}
We use 28 training and 18 unseen benchmarks (Table~\ref{tab:benchmark_splits}; Appendix~\ref{app:task-tree}). Train evaluates a fixed set of representative instances per benchmark over multiple episodes; Unseen uses each benchmark's official evaluation set or a fixed official subset. Five experiments examine capability-driven organization, recursive self-improvement, capability-aware Harness routing, capability--module relationships, and embodied transfer of capability-evolved Harnesses.

\input{tab/benchmark_splits}

All variants share Base Harness $\TrinityInitialHarness$ (Prime Agent) and a frozen Claude Opus 5 backbone~\citep{anthropic2026opus5}. Table~\ref{tab:results} in Appendix~\ref{app:evaluation} reports metrics, instance and episode counts, and full results. Token costs are totals in millions (M); Tables~\ref{tab:exp_1.3}--\ref{tab:exp_4} report inference costs, and -- denotes data not required.

\paragraph{Evaluation-Time Skill Preparation.}
\label{sec:skill_preparation}
\label{sec:unseen_skill_adaptation}
To reflect real deployment and account for each Harness's effect on skill learning, we run five preparation trials per benchmark--Harness pair, summarize their trajectories into skills, and keep these skills fixed for subsequent evaluation on Train and Unseen.

\input{sections/4.2_exp}
\input{sections/4.3_exp}
\input{sections/4.4_exp}

%% file: tab/benchmark_splits.tex
\begin{table}[H]
\centering
\captionsetup{font=normalsize,skip=7pt}
\caption{\textbf{Training and unseen benchmarks.} \textsuperscript{*}Agent-built from web-searched sources.}
\label{tab:benchmark_splits}

\small
\setlength{\tabcolsep}{6pt}
\renewcommand{\arraystretch}{1.18}
\renewcommand{\tabularxcolumn}[1]{m{#1}}
\definecolor{SplitRule}{HTML}{FFFACC}
\definecolor{SplitConstraint}{HTML}{D8FFE5}
\definecolor{SplitControl}{HTML}{D8E5FF}
\definecolor{SplitInteraction}{HTML}{FCD4A3}
\definecolor{SplitHorizon}{HTML}{FFDEDF}
\newcommand{\benchname}[1]{\mbox{#1}}
\newcommand{\agentbench}[1]{\benchname{#1\textsuperscript{*}}}
% Match the current 46-benchmark inventory, including simulation and robot tasks.
% Task-domain assignments follow Appendix/scripts/tasks.py.
\begin{tabularx}{\linewidth}{>{\centering\arraybackslash}m{0.20\linewidth}
  >{\hsize=1.10\hsize\centering\arraybackslash}X
  >{\hsize=0.90\hsize\centering\arraybackslash}X}
Task domain & Train & Unseen \\
\hline
\rowcolor{SplitRule}
Rule discovery
  & \agentbench{Eleusis}~\citep{golden2020eleusisexpress}, \benchname{Mastermind}~\citep{golde2025mastermindeval}, \benchname{WILT}~\citep{banatt2024wilt}, \benchname{BoxingGym}~\citep{gandhi2025boxinggym}, \benchname{DiscoverPhysics}~\citep{wiemann2026discoverphysics}, \benchname{ARC-AGI-3}~\citep{arcprize2026arcagi3}, \benchname{PuzzleScript}~\citep{lavelle2026puzzlescript}
  & \benchname{Entity Deduction Arena}~\citep{zhang2024eda}, \benchname{SciGym}~\citep{duan2025scigym}, \benchname{DiG-bench}~\citep{battleday2026digbench} \\
\hline
\rowcolor{SplitConstraint}
Constrained problem solving
  & \agentbench{Chess Tactics}~\citep{lichess2026puzzles}, \benchname{puzz.link / Penpa+}~\citep{robx2026pzprjs,penpa2026}, \agentbench{WCA Cube Solving}~\citep{wca2025regulations}, \benchname{AdaPlanBench}~\citep{liu2026adaplanbench}, \agentbench{Master Keying}~\citep{pulford2007locks}
  & \benchname{Pencil Puzzle Bench}~\citep{waugh2026ppbench}, \benchname{MiniHack Boxoban}~\citep{samvelyan2021minihack}, \benchname{ALE-Bench}~\citep{imajuku2025alebench}, \benchname{BuildArena}~\citep{xia2026buildarena} \\
\hline
\rowcolor{SplitControl}
Perception and control
  & \agentbench{Pallet Rack Inspection}~\citep{en15635}, \benchname{PInVerify}~\citep{jiang2026pinverify}, \benchname{Memory Maze}~\citep{pasukonis2022memorymaze}, \agentbench{Micromouse}~\citep{ukmarsMicromouseRules}, \agentbench{Euro NCAP AEB}~\citep{euroncap2023aeb}, \benchname{Pooltool Billiards}~\citep{xu2025deepphy,kiefl2024pooltool}, \benchname{Video-Index}~\citep{song2026videoindex}, \benchname{SOON (simulation)}~\citep{zhu2021soon}, \benchname{CaP-X (simulation)}~\citep{fu2026capx}
  & \benchname{ActiView}~\citep{wang2025actiview}, \benchname{MiniHack Maze / Corridor}~\citep{samvelyan2021minihack}, \benchname{PHYRE}~\citep{bakhtin2019phyre}, \benchname{CaP-X (robot)}~\citep{fu2026capx}, \benchname{Video-MME-v2}~\citep{fu2026videommev2}, \benchname{SOON (robot)}~\citep{zhu2021soon} \\
\hline
\rowcolor{SplitInteraction}
Multi-agent interaction
  & \benchname{Among Us}~\citep{golechha2025amongus}, \agentbench{Saboteur}~\citep{moyersoen2023saboteur}, \benchname{Codenames}~\citep{stephenson2025codenames}, \benchname{Colonel Blotto}~\citep{guertler2025textarena}, \benchname{GLEE Persuasion}~\citep{shapira2024glee}
  & \benchname{Coup}~\citep{guertler2025textarena}, \benchname{Hanabi}~\citep{bard2020hanabi}, \benchname{Heads-Up NL Hold'em}~\citep{deepmind2026gamearena,lanctot2019openspiel} \\
\hline
\rowcolor{SplitHorizon}
Long-horizon environments
  & \benchname{TextQuests}~\citep{phan2025textquests}, \agentbench{The Long Dark}~\citep{hinterland2017longdark}
  & \benchname{NetHack}~\citep{kuttler2020nethack}, \benchname{Crafter}~\citep{hafner2022crafter} \\
\end{tabularx}
\end{table}

%% file: sections/4.2_exp.tex
\subsection{Capability-Driven Organization}
\label{sec:capability_evolution}
\label{sec:organization_composition}

\paragraph{Preliminary Comparison: Modules and Skills.}
\label{sec:component_ablation}

\input{tab/exp_1.2}
Table~\ref{tab:exp_1_2} compares four settings at \mbox{Iteration~1}. \textbf{Base} uses initial modules; \textbf{\mbox{Module Only}} evolves them; \textbf{\mbox{Skill Only}} adds skills to Base; \textbf{Ours} combines both and performs best on Train and Unseen. Skills follow Section~\ref{sec:skill_preparation}. \mbox{Organization} comparisons use \textbf{\mbox{Module Only}} to isolate the effect of Harness organization.

\par\WFclear

\Needspace{6\baselineskip}
\paragraph{Organization Comparison.}
Table~\ref{tab:exp_1_1} compares five organizations after one RSI iteration to assess task grouping, hierarchical inheritance, and training curricula. \textbf{Global} pools experience from all 28 original environments into one shared Harness. \textbf{Task-Domain} learns 15 task-specific Harnesses, whereas \textbf{Capability-Domain} uses six flat capability groups.

\textbf{Capability Tree (Real)} and \textbf{Capability Tree (Sim)} use the same prefix-based topology, selective parent inheritance, and node-specific local refinement described in Section~\ref{sec:rsi_method}. Real trains on original benchmark tasks; Sim uses verified synthetic worlds that preserve their decision processes. Domain/tree variants use one Harness and environment per domain/node. Training counts evolution-stage tokens; Inf.\ Train/Unseen count execution-only evaluation tokens.
\input{tab/exp_1.1}
\FloatBarrier

\paragraph{Evaluation.}
Flat grouping reduces training cost but underperforms Global: Capability-Domain uses the fewest training tokens at 136.94M, with Train/Unseen scores of 49.25/39.43. Capability Tree (Real) gains 4.19/2.88 points over this flat organization, supporting selective inheritance across shared capability prefixes. Sim adds another 3.51/4.95 points, reaching the best module-only scores of 56.95/47.26. Its inference costs are also lowest at 514.42M/549.07M. First-iteration training costs 405.70M, versus 211.56M for Real and 278.25M for Task-Domain.

\noindent
\begin{minipage}[t]{0.52\linewidth}
    \vspace{0pt}
    \paragraph{Training Cost and Environment Reuse.}
Figure~\ref{fig:training_cost_analysis} shows that Creator's environment construction dominates the initial training cost. Reusing worlds reduces this overhead in later iterations. Explorer collects the same number of trajectories across iterations with comparable inference-token usage. Gains at fixed trajectory volume suggest that cleaner evolution signals from capability-focused environments support more effective Harness updates. Reuse thus amortizes construction costs as Train performance improves. Full results and the evaluation protocol appear in Appendix~\ref{app:evaluation}, Table~\ref{tab:results}.
\end{minipage}\hfill
\begin{minipage}[t]{0.46\linewidth}
    \vspace{0pt}
    \centering
    \captionsetup{font=normalsize,justification=raggedright,singlelinecheck=false,skip=2pt}
    \includegraphics[width=\linewidth]{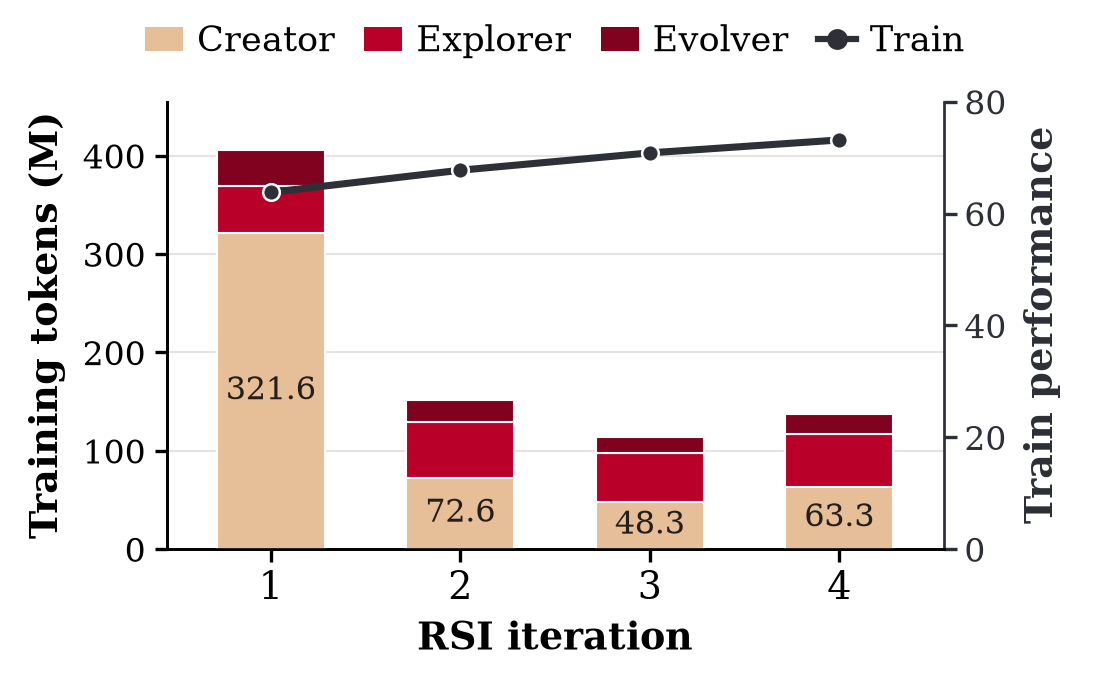}
    \captionof{figure}{\textbf{Training cost and Train performance across RSI iterations.}}
    \label{fig:training_cost_analysis}
\end{minipage}
\par

\TrinityFinding{\textcolor{DarkRed}{Finding 1.} Capability-driven organization outperforms task-driven specialization.}{A capability-centric perspective gives rise to a capability tree, where capability-driven Real2Sim2Real achieves stronger capability transfer and lower inference costs with fewer training environments, demonstrating that less is more.}

% ============================================================
% 4.3 Recursive Self-Improvement
% ============================================================
\newpage

\begin{figure}[t]
    \centering
    \includegraphics[width=\linewidth]{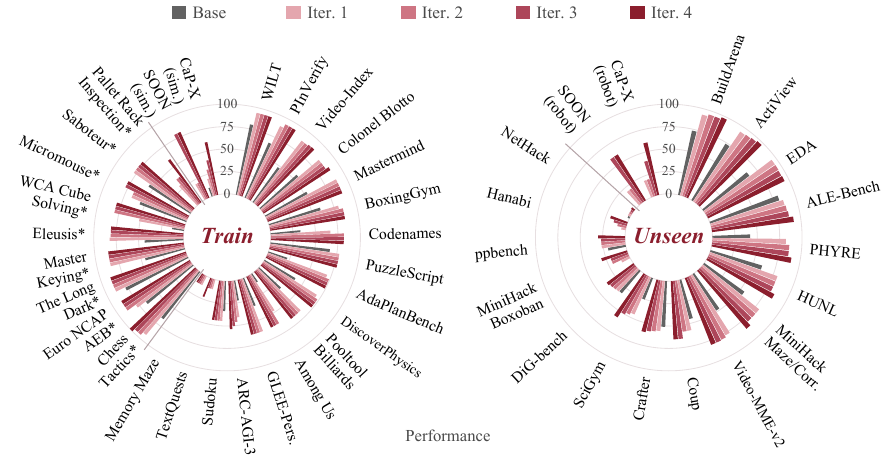}
    \caption{\textbf{Benchmark-level Harness evolution.}
    Left: Train; right: Unseen. Radial bars show Base and Iterations~1--4; \textsuperscript{*}Agent-built from web-searched sources.}
    \label{fig:exp_4.2.2}
    \vspace{-10pt}
\end{figure}
% Place the benchmark figure before the numbered wrapped trend figure.
\FloatBarrier

\Needspace{18\baselineskip}
\vspace{-1.5em}
\subsection{Recursive Self-Improvement}
\label{sec:exp_recursive_self_improvement}

% Keep the text close to the figure and stop wrapping at the caption's bottom.
\begingroup
\setlength{\columnsep}{9pt}
\begin{wrapfigure}[12]{r}{0.47\textwidth}
    \vspace{-1.5em}
    \centering
    \captionsetup{font=normalsize,justification=raggedright,singlelinecheck=false,skip=4pt}
    \includegraphics[width=\linewidth]{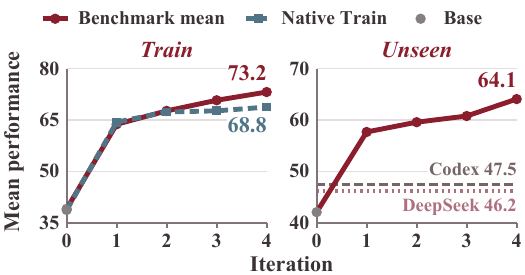}
    \caption{\textbf{RSI mean performance across benchmarks.} Horizontal lines show Codex and DeepSeek reference means.}
    \label{fig:rsi_mean_trends}
\end{wrapfigure}
We track the RSI loop (Section~\ref{sec:rsi_method}) from Base (Iteration~0) through Iteration~4 to test whether improvements continue after the first update. With the benchmark split, capability requirements, tree topology, and backbone fixed, Evolver refines modules from training feedback, Creator renews or reuses curricula, and Explorer collects new trajectories.
\par

\paragraph{Evaluation.}
Figure~\ref{fig:exp_4.2.2} reports scores for 28 Train and 18 Unseen benchmarks, with five radial bars per benchmark. Scores retain benchmark-specific metrics on a 0--100 scale; Appendix~\ref{app:evaluation} gives full results and protocols.
Figure~\ref{fig:rsi_mean_trends} averages benchmarks equally. Train covers all 28 benchmarks, with a dashed curve for the 17 native benchmarks; Unseen and its Codex~\citep{openai2025codex} and DeepSeek~\citep{deepseek2026harness} references cover the 16 non-robot benchmarks. From Base to Iteration~4, Train rises from 39.1 to 73.2 and Unseen from 42.1 to 64.1. The first iteration yields the largest gains, but Iterations~2--4 add another 9.4 Train and 6.4 Unseen points. Native Train also improves from 64.4 to 68.8 after Iteration~1, with smaller later gains than the full Train mean.
Progress varies across benchmarks: between Iterations~1 and~4, ARC-AGI-3 rises from 41.0 to 53.6, SciGym from 45.2 to 57.1, and BuildArena from 93.3 to 97.8. Mastermind stays at 88.9, while Eleusis declines from 82.2 to 81.1. The final Unseen mean exceeds the Codex (47.5) and DeepSeek (46.2) references, although Crafter and DiG-bench remain below both comparators (Appendix~\ref{app:evaluation}).

% Keep Evaluation in the wrap; restore full width before the finding box.
\par\WFclear
\endgroup

\TrinityFinding{\textcolor{DarkRed}{Finding 2.} Capability-driven RSI sustains iterative improvement.}{Continued gains on a fixed capability tree suggest that simulated environments offer further opportunities for capability refinement, with improvements extending to unseen tasks.}
% Keep the RSI figure and its finding ahead of the routing results.
\FloatBarrier

% ============================================================
% 4.4 Capability-Aware Harness Routing
% ============================================================

\subsection{Capability-Aware Harness Routing}
\label{sec:capability_routing}

We compare five configurations on the Iteration-4 Explorer Harness tree $\TrinityHarnessTree{4}$ (Section~\ref{sec:router_method}).
\textbf{Random} selects one candidate Harness per task at random.
\textbf{Agent Self-Routing} has three variants: public task information and interface descriptions alone; adding capability definitions, annotated branch cards, and aggregate observed performance--cost summaries; and further retrieving execution experience, yielding the full \textbf{Harness Router}.
\textbf{Oracle} selects the best compatible Harness retrospectively per instance as an upper reference.
In Table~\ref{tab:exp_1.3}, Cap.\ and Mem.\ denote explicit capability definitions and retrieved execution experience; these configuration columns apply only to self-routing variants.

\Needspace{10\baselineskip}
\paragraph{Evaluation.}
\input{tab/exp_1.3}
Table~\ref{tab:exp_1.3} pairs benchmark-equal performance with estimated total inference tokens (M) for each split. Full self-routing scores 71.27 on Train and 60.96 on Unseen, compared with Oracle's 73.16 and 63.61, respectively. The token estimates cover skill preparation, routing, and formal inference. Deployable variants share the frozen backbone, episode budgets (including routing calls), initial states, and seeds. Each Harness prepares and freezes skills under Section~\ref{sec:skill_preparation} using matched instances and budgets. Neither skills nor preparation records enter routing decisions or memory; evaluation capability labels are also excluded from memory. Unseen routing starts from training memory, and verified feedback updates it only for later instances. Technical failures score zero; invalid selections \mbox{receive no fallback.}
\par\WFclear

\TrinityFinding{\textcolor{DarkRed}{Finding 3.} Capability priors guide agent routing.}{An agent acting as a Router can use capability priors to select suitable Harnesses. The results suggest that the capability tree provides a semantic basis for matching inferred task requirements to Harness configurations.}

%% file: tab/exp_1.2.tex
\begin{wraptable}{r}{0.50\textwidth}
\vspace{-2.0 \baselineskip}
\vspace{-8pt}
\centering

\captionsetup{
    font=normalsize,
    justification=raggedright,
    singlelinecheck=false,
    skip=4pt
}
\vspace{-0.5em}
\caption{\textbf{Preliminary comparison of modules and skills.} Module denotes evolved modules beyond Base Harness. Combining both improves performance over either alone.}
\label{tab:exp_1_2}

\colorlet{HarnessTint}{DarkRed}
\colorlet{HarnessBase}{black!22}
\newcommand{\hfocus}[1]{\textcolor{white}{#1}}

\small
\setlength{\tabcolsep}{3pt}
\renewcommand{\arraystretch}{1.15}

\begin{adjustbox}{max width=\linewidth}
\begin{tabularx}{\linewidth}{>{\raggedright\arraybackslash}Xcccc}
Method & Module & Skill & Train & Unseen \\

\midrule

\rowcolor{HarnessBase}
Base ($\TrinityInitialHarness$)
& $\times$
& $\times$
& 39.06
& 37.45 \\

Module Only
& $\checkmark$
& $\times$
& 56.95
& 47.26 \\

Skill Only
& $\times$
& $\checkmark$
& 58.86
& 48.67 \\

\rowcolor{HarnessTint}
\hfocus{\textbf{Ours ($\TrinityHarnessTree{1}$)}}
& \hfocus{$\checkmark$}
& \hfocus{$\checkmark$}
& \hfocus{\textbf{63.80}}
& \hfocus{\textbf{53.53}} \\

\end{tabularx}
\end{adjustbox}

\vspace{-6pt}
\end{wraptable}

%% file: tab/exp_1.1.tex
\begin{table}[t]
\centering
\captionsetup{font=normalsize,skip=7pt}
\caption{\textbf{Harness organization and training curricula.} All methods use module-only evaluation without skills. The simulated capability tree achieves the highest Train and Unseen performance.}
\label{tab:exp_1_1}

% Use the report's existing red palette for the author's table style.
\colorlet{HarnessAccent}{DarkRed}
\colorlet{HarnessTint}{DarkRed}
\colorlet{HarnessBase}{black!22}
\definecolor{HarnessMuted}{HTML}{776C6C}

\small
\setlength{\tabcolsep}{2pt}
\renewcommand{\arraystretch}{1.15}

% Entries not required.
\newcommand{\hmiss}{%
  \multicolumn{1}{c}{\textcolor{HarnessMuted}{--}}}
\newcommand{\hfocus}[1]{\textcolor{white}{#1}}

% Each domain/node trains one Harness in one environment.
% Base and Capability Tree (Sim) match the Base and Module Only conditions
% in tab/exp_1.2.tex. Inference totals exclude skill preparation.
\begin{adjustbox}{max width=\linewidth}
\begin{tabularx}{\linewidth}{>{\raggedright\arraybackslash}X*{8}{c}}
Method & Curriculum & \#\,H & \#\,Env. & Train & Unseen
  & Training (M) & Inf. Train (M) & Inf. Unseen (M) \\
\midrule

\rowcolor{HarnessBase}
Base ($\TrinityInitialHarness$)
  & \hmiss & \hmiss & \hmiss
  & 39.06 & 37.45 & \hmiss & 582.54 & 613.07 \\

Global
  & Real & 1 & 28
  & 53.04 & 42.92 & 582.54 & 531.81 & 567.68 \\
  
\addlinespace[4pt]
\multicolumn{9}{l}{%
  \textbf{\groupheading{A.\quad Task vs.\ Capability Organization}}
} \\[2pt]

Task-Domain
  & Real & 15 & 15
  & 51.17 & 40.82 & 278.25 & 540.36 & 577.33 \\

Capability-Domain
  & Real & 6 & 6
  & 49.25 & 39.43 & 136.94 & 562.32 & 596.58 \\

\addlinespace[4pt]
\multicolumn{9}{l}{%
  \textbf{\groupheading{B.\quad Real Tasks vs.\ Capability Simulation}}
} \\[2pt]

Capability Tree
  & Real & 10 & 11
  & 53.44 & 42.31 & 211.56 & 526.53 & 563.90 \\

\rowcolor{HarnessTint}
\hfocus{\textbf{Capability Tree}}
  & \hfocus{\textbf{Sim}} & \hfocus{10} & \hfocus{\textbf{11}}
  & \hfocus{\textbf{56.95}} & \hfocus{\textbf{47.26}} & \hfocus{405.70}
  & \multicolumn{1}{c}{\hfocus{514.42}}
  & \multicolumn{1}{c}{\hfocus{549.07}} \\

\end{tabularx}
\end{adjustbox}

\end{table}

%% file: tab/exp_1.3.tex
\begin{wraptable}{r}{0.62\textwidth}
\vspace{-8pt}
\centering
\captionsetup{
    font=normalsize,
    justification=raggedright,
    singlelinecheck=false,
    skip=4pt
}
\vspace{-0.5em}
\caption{\textbf{Capability-aware Harness routing.} The full self-routing configuration approaches Oracle on both splits.}
\label{tab:exp_1.3}
\colorlet{HarnessTint}{DarkRed}
\colorlet{HarnessBase}{black!22}
\definecolor{HarnessMuted}{HTML}{776C6C}
\newcommand{\hmiss}{\multicolumn{1}{c}{\textcolor{HarnessMuted}{--}}}
\newcommand{\hfocus}[1]{\textcolor{white}{#1}}

\small
\setlength{\tabcolsep}{1.5pt}
\renewcommand{\arraystretch}{1.15}
\begin{adjustbox}{max width=\linewidth}
\begin{tabularx}{\linewidth}{>{\raggedright\arraybackslash}Xcccccc}
Method & Cap. & Mem. & Train & Tok. (M) & Unseen & Tok. (M) \\
\midrule
\rowcolor{HarnessBase}
Random & \hmiss & \hmiss & 52.74 & 567.08 & 46.23 & 594.25 \\
Agent Self-Routing & $\times$ & $\times$ & 65.79 & 535.53 & 52.36 & 582.29 \\
Agent Self-Routing & $\checkmark$ & $\times$ & 70.34 & 510.55 & 58.27 & 556.21 \\
\rowcolor{HarnessTint}
\hfocus{\textbf{Agent Self-Routing}} & \hfocus{$\checkmark$} & \hfocus{$\checkmark$}
  & \hfocus{\textbf{71.27}} & \hfocus{497.89}
  & \hfocus{\textbf{60.96}} & \hfocus{541.28} \\

Oracle & \hmiss & \hmiss & 73.16 & 486.53 & 63.61 & 528.34 \\
\end{tabularx}
\end{adjustbox}
\vspace{-6pt}
\end{wraptable}

%% file: sections/4.3_exp.tex
% Keep each visual inside its corresponding analysis paragraph.
\FloatBarrier

\subsection{Capability--Module Relationships}
\label{sec:capability_module}

\paragraph{Capability--Module Correlations.}
\label{sec:capability_module_heatmap}
\begin{wrapfigure}{r}{0.56\textwidth}
    \centering
    \vspace{-1.5em}
     \captionsetup{font=normalsize,justification=raggedright,singlelinecheck=false,skip=4pt}
    \includegraphics[width=\linewidth]{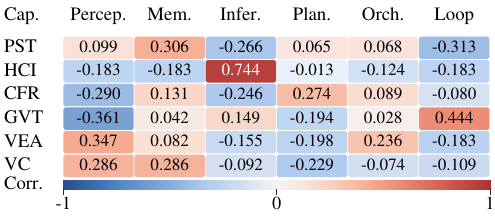}
    \caption{\textbf{Capability–protocol correlations across six protocols.}
    Red is positive; blue is negative.}
    \label{fig:capability_module_heatmap}
    \vspace{-6pt}
\end{wrapfigure}
Figure~\ref{fig:capability_module_heatmap} reports Iteration-4 correlations for 29 modules.
Rows abbreviate six shared capabilities: \textbf{PST}, perception and state tracking; \textbf{HCI}, hypothesis and counterpart inference; \textbf{CFR}, constraint checking and feedback repair; \textbf{GVT}, goal commitment and value trade-off; \textbf{VEA}, video evidence acquisition; and \textbf{VC}, video comprehension. Columns denote Perception, Memory, Inference, Planning, Orchestration, and Loop (Section~\ref{sec:harness_formalization}), with coefficients in $[-1,1]$. The strongest positive association in each row links PST to Memory ($0.306$), HCI to Inference ($0.744$), CFR to Planning ($0.274$), GVT to Loop ($0.444$), and VEA to Perception ($0.347$); VC correlates equally with Perception and Memory ($0.286$). The most negative coefficients link GVT to Perception ($-0.361$) and PST to Loop ($-0.313$). These descriptive associations motivate the \mbox{module interventions below.}

\par\WFclear

\Needspace{18\baselineskip}
\paragraph{Module Interventions.}
\label{sec:capability_module_ablation}
\input{tab/exp_2}
Table~\ref{tab:exp_4} evaluates module interventions on Train (the fixed instance set of Section~\ref{sec:setup}) with skills prepared under Section~\ref{sec:skill_preparation}. Scores are benchmark-equal means on a 0--100 scale; reported inference token totals are in millions (M). For addition, we randomly select one relevant or irrelevant module from the same library and add it to $\TrinityInitialHarness$. Candidates pass interface checks and exclude installed modules; additions match protocol class and have comparable description lengths. Relevance follows fixed task requirements; irrelevant modules target capabilities outside both these requirements.

Adding a relevant module raises the Base score from 39.06 to 58.45 (\textcolor{green!60!black}{$+19.39$}), versus 40.08 (\textcolor{green!60!black}{$+1.02$}) for an irrelevant module; token totals are 551.85M and 583.37M, respectively, against Base's 582.54M. Removing one relevant installed module from the intact Iteration-4 Harness lowers its score from 73.16 to 70.42 (\textcolor{red!75!black}{$-2.74$}) and raises tokens from 496.70M to 517.26M.
\par\WFclear
\TrinityFinding{\textcolor{DarkRed}{Finding 4.} Capability--module priors inform Harness design.}{Capability--protocol associations suggest specialization patterns, while controlled interventions show larger gains from relevant module additions and performance losses after their removal. Together, these results support using task capability requirements to guide module selection and composition.}
\FloatBarrier

%% file: tab/exp_2.tex
\begin{wraptable}{r}{0.56\textwidth}
\centering
\captionsetup{
    font=normalsize,
    justification=raggedright,
    singlelinecheck=false,
    skip=4pt
}
\caption{\textbf{Module interventions.} $\Delta$ is the difference in displayed scores from each bold baseline.}
\label{tab:exp_4}
\colorlet{HarnessTint}{DarkRed}
\definecolor{HarnessMuted}{HTML}{776C6C}
\newcommand{\hmiss}{\multicolumn{1}{c}{\textcolor{HarnessMuted}{--}}}

\small
\setlength{\tabcolsep}{2.5pt}
\renewcommand{\arraystretch}{1.28}
% This module-only experiment reports execution-only inference costs.
\begin{adjustbox}{max width=\linewidth}
\begin{tabularx}{\linewidth}{>{\raggedright\arraybackslash}Xccc}
Method & Train & $\Delta$ & Tokens (M) \\
\midrule
\rowcolor{black!22}
\textbf{Base} ($\TrinityInitialHarness$)
& 39.06 & 0.00 & 582.54 \\
\hspace{1em}$+$\quad irrelevant module
& 40.08 & \textcolor{green!60!black}{$+1.02$} & 583.37 \\
\hspace{1em}$+$\quad relevant module
& 58.45 & \textcolor{green!60!black}{$+19.39$} & 551.85 \\
\addlinespace[3pt]
\rowcolor{black!22}
\textbf{Ours} ($\TrinityHarnessTree{4}$) & \textbf{73.16} & 0.00 & 496.70 \\
\hspace{1em}$-$\quad relevant module & 70.42 & \textcolor{red!75!black}{$-2.74$} & 517.26 \\
\end{tabularx}
\end{adjustbox}
\vspace{-6pt}
\end{wraptable}

%% file: sections/4.4_exp.tex
\FloatBarrier
\subsection{Embodied Transfer of Capability-Evolved Harnesses}
\label{sec:embodied_real2sim2real}
\begingroup
\setlength{\parskip}{4pt}

\providecommand{\EmbodiedAppendixRef}{Appendix~\ref{app:embodied}}
\par\noindent
\begin{minipage}{\linewidth}
    \centering
    \includegraphics[width=\linewidth]{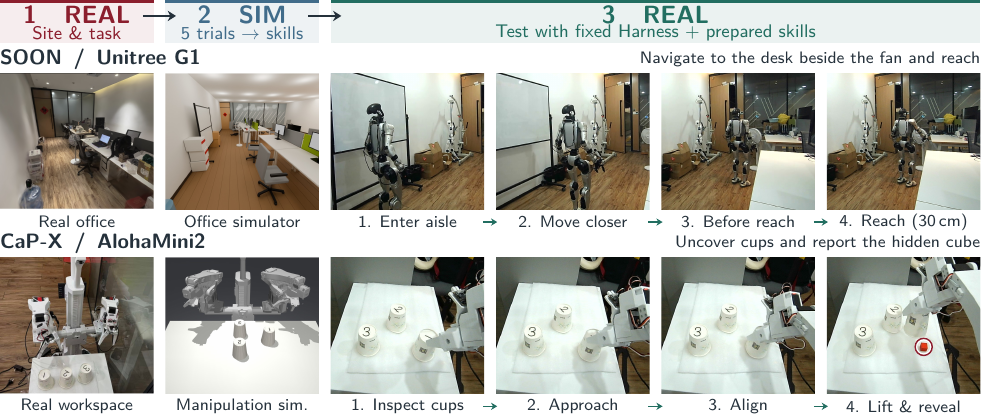}
    \captionsetup{font=normalsize,justification=raggedright,singlelinecheck=false,skip=4pt}
    \captionof{figure}{\textbf{Embodied Transfer of Capability-Evolved Harnesses.} Real scenes, simulators, and robot trials (SOON E01; CaP-X C-01). The circle marks the revealed cube.}
    \label{fig:embodied_real2sim2real}
\end{minipage}
\par
Based on their capability requirements, we map the embodied tasks to the $d\!\to\!a$ node of the capability tree and adopt the Harness configurations evolved there. We then use \textbf{Real2Sim2Real} to learn task-specific skills (Figure~\ref{fig:embodied_real2sim2real}). For each task--Harness pair, we hold the Harness fixed, run five trial-and-error episodes in the reconstructed simulator, and summarize the trajectories into skills. These skills are frozen for formal simulation and real-robot tests; the five preparation episodes are excluded from test counts and success rates.

\paragraph{Protocol.}
In \textbf{SOON}~\citep{zhu2021soon}, a Unitree G1 approaches the desk beside a white fan and reaches towards it; real-robot success requires the target desk first ahead of the extended arm, within 30\,cm. In \textbf{CaP-X}~\citep{fu2026capx}, an AlohaMini2 uncovers three inverted cups and reports a cube's presence, colour, and location, or its absence. All five Harness configurations are evaluated with their own prepared skills. \EmbodiedAppendixRef{} gives action budgets, observations, and available trajectory records.

\input{tab/embodied_transfer}

\paragraph{Evaluation.}
Table~\ref{tab:embodied_transfer} compares all five Harness configurations after skill preparation. SOON improves from 0 to 80\% in simulation and from 0 to 60\% on the robot; CaP-X improves from 10 to 60\% and from 0 to 60\%, respectively. Both robots reach 20\% with $\mathcal{H}^{(1)}$. SOON reaches its final real-robot score at $\mathcal{H}^{(3)}$, while CaP-X improves from 40 to 60\% between $\mathcal{H}^{(3)}$ and $\mathcal{H}^{(4)}$.

\par\WFclear
\paragraph{Scope.}
Recorded deployments differ in message handling, interfaces, guidance, and acceptance rules (including lifting versus pushing). These comparisons evaluate complete Harness--skill configurations and do not isolate the effects of module evolution and skill preparation.
\TrinityFinding{\textcolor{DarkRed}{Finding 5.} Capability-Driven RSI supports embodied transfer.}{The evaluated Harness--skill configurations improve on both simulated and physical tasks, linking node-level module evolution with task-specific skill preparation.}
\endgroup
\FloatBarrier

%% file: tab/embodied_transfer.tex
\begin{wraptable}{r}{0.58\textwidth}
% Align the caption's first baseline with the adjacent Evaluation paragraph.
\vspace{-14.15pt}
\centering
\captionsetup{font=normalsize,justification=raggedright,singlelinecheck=false,skip=4pt}
\caption{\textbf{Embodied performance across iterations.} Success rates (\%) with prepared skills.}
\label{tab:embodied_transfer}
\centering
\small
\setlength{\tabcolsep}{1pt}
\renewcommand{\arraystretch}{1.15}
% Simulation series follow appendix/tables/results.tex. Physical-robot
% percentages follow the author's updated Table 6 screenshot (2026-10-24).
\begin{tabularx}{\linewidth}{@{}l*{4}{>{\centering\arraybackslash}X}@{}}
Harness & \mbox{SOON (Sim)} & \mbox{CaP-X (Sim)} & \mbox{SOON (Real)} & \mbox{CaP-X (Real)} \\
\midrule
\rowcolor{black!22}
$\mathcal{H}_0$ & 0 & 10 & 0 & 0 \\
$\mathcal{H}^{(1)}$ & 20 & 20 & 20 & 20 \\
$\mathcal{H}^{(2)}$ & 20 & 30 & 20 & 20 \\
$\mathcal{H}^{(3)}$ & 80 & 40 & 60 & 40 \\
\rowcolor{DarkRed}
\textcolor{white}{$\boldsymbol{\mathcal{H}^{(4)}}$} & \textcolor{white}{\textbf{80}} & \textcolor{white}{\textbf{60}}
& \textcolor{white}{\textbf{60}} & \textcolor{white}{\textbf{60}} \\
\end{tabularx}
\vspace{-6pt}
\end{wraptable}

%% file: sections/5_related_work.tex
% =====================================================================
%  Related Work
% =====================================================================
\section{Related Work}
\label{sec:related}

\paragraph{Recursive Harness Self-Improvement.}
Self-improvement spans G\"odel machines~\citep{schmidhuber2007godel}, LLM systems that rewrite their own agent code~\citep{zelikman2023self,yin2024g,zhang2026darwin}, automated agent design~\citep{hu2025automated,zhang2025aflow,shang2025agentsquare}, prompt optimization~\citep{yang2024large,fernando2023promptbreeder,yuksekgonul2024textgrad}, and learning from self-feedback and reflection~\citep{madaan2023self,shinn2023reflexion,zhao2024expel}. With weights fixed, improvement shifts to the harness, which is synthesized from environment feedback~\citep{lou2026autoharness}, repaired from failed trajectories~\citep{chen2026failed,shao2026harness}, self-revised~\citep{zhang2026self}, optimized as code~\citep{lee2026meta,lin2026agentic}, adapted at test time~\citep{nie2026tthe} or per task~\citep{lee2026recursive}, improved at the meta level~\citep{wang2026metaskill,tan2026metarsi}, and shown to lift frozen MLLMs~\citep{lu2026muse}; benchmarks and surveys map this space~\citep{yao2026harness,wu2026harnessdev,ren2026self}. Closest to ours are JIT-Agent's harness generator~\citep{zhang2026jit}, HarnessX's typed primitives~\citep{chen2026harnessx}, HSI's per-task-family harnesses~\citep{zhou2026hierarchical}, ModularRSI's modules evolved on benchmark-disjoint tasks~\citep{wu2026modularrsi}, and the flat capability map of Harness Continual Learning~\citep{kang2026harness} and regime-branched harness tree of Adaptive Auto-Harness~\citep{liu2026adaptive}, both evolved on the task stream. Trinity instead induces a capability tree from training trajectories, shares prefixes across 44 heterogeneous benchmarks with selective module inheritance, evolves Harnesses in Creator-generated worlds that exclude target-task layouts and solutions, and evaluates on held-out benchmarks and real robots. Unlike LLM routers~\citep{zhou2026agentrouter}, ours routes among capability-specific Harnesses over one frozen backbone.

\paragraph{Co-evolution of Agents and Environments.}
Open-ended co-evolution~\citep{brant2017minimal,wang2019paired,wang2020enhanced}, unsupervised environment design~\citep{parker2022evolving,yuan2026pace}, curricula~\citep{bengio2009curriculum,graves2017automated}, and embodied co-design~\citep{wang2025embodied} adapt training to the learner; Voyager grows a skill library under an automatic curriculum~\citep{wang2023voyager}. Loops co-evolve language agents with environments~\citep{hu2026seal}, optimizers~\citep{liu2026escher}, or harnesses~\citep{chen2026evotrainer,hebbar2026sia}; environment generation spans embodied~\citep{zala2024envgen,kang2026simworld}, tool-use~\citep{dong2026agent}, and terminal settings~\citep{gandhi2026endless,cheng2026terminal}, language world models~\citep{zuo2026qwen}, and game verification~\citep{jia2026gamegen}. SPADE is closest~\citep{liu2026spade}: one LLM designs environments with verification code and learns roles by reinforcement learning. Trinity keeps the backbone frozen and couples world generation with Explorer and Creator Harness evolution along a capability hierarchy, execution outcomes refining Explorer and verifier-detected failures refining Creator.

%% file: sections/6_sum.tex
\section{Conclusion}
We advocate Capability-Driven Recursive Self-Improvement and introduce \textbf{The Trinity}, a framework that couples world generation with Harness evolution through Creator, Explorer, and Evolver.
By organizing heterogeneous tasks around shared capability requirements, the framework constructs simplified, verifiable worlds whose execution feedback guides the evolution of reusable modules.
These modules form specialized Harness configurations, with capability--module correlations and controlled interventions providing empirical support for capability requirements as priors for module design and composition.
To apply these evolved capabilities across tasks, capability-aware routing selects suitable Harness configurations using public task information and verified execution experience.
Experiments across diverse training and unseen benchmarks demonstrate improved aggregate performance and cross-task generalization, while deployment in two real-world embodied settings further illustrates the transfer of these capabilities to physical environments.

%% file: sections/7_diss.tex
\section{Discussion}
\label{sec:discussion}
\paragraph{Limitations.}
Our framework uses web search for benchmark discovery, but retrieval coverage and result quality may vary across queries.
In addition, our current study focuses primarily on self-improvement at the Harness level.

\paragraph{Outlook.}
Capability-driven learning offers a new paradigm for generalizable agent improvement, where shared capability requirements across tasks guide learning and optimization. Our framework and results at the Harness level demonstrate the promise of this perspective. Looking ahead, we envision dynamically expanding the capability tree to capture emerging capability requirements and extending this paradigm to model parameter training, where capability-aware routing and RSI form a closed loop that jointly drives capability discovery, acquisition, and refinement.

%% file: sections/8_authors.tex
\section{Authors}
\label{sec:authors}

% =========================
% Authors
% =========================
\textbf{Core Contributors:}
Weidong Tang$^{1,2}$\quad
Ruiheng Wang$^{3}$\quad
Jingxuan Zhang$^{4}$\quad
Hexin Wang$^{1}$\quad

\textbf{Contributors:}
Haoyuan Chen$^{1}$\quad
Zhengfeiyang Zhang$^{1,4}$\quad
Yixing Ma$^{1,5}$\quad
Zichen Jiang$^{4}$\quad
Ziyu Guo$^{6}$\quad
Haodong Li$^{6}$\quad
Zhengbo Jiao$^{7}$\quad
Zhiqing Cui$^{7}$\quad
Xinxiang Yin$^{8}$\quad
Shuo Liang$^{1,4}$\quad
Zhenglin Wan$^{4}$\quad
Hongzhan Lin$^{4}$\quad
Wangbo Zhao$^{7}$\quad
Kaipeng Zhang$^{9}$\quad

\textbf{Corresponding Authors:}
Pengfei Zhou$^{1,4}$\quad
Odin Zhang$^{6}$\quad
Yang You$^{4}$\quad

\textbf{Affiliations:}
$^{1}$ InfRec \quad
$^{2}$ Xidian University \quad
$^{3}$ Tongji University \quad
$^{4}$ National University of Singapore \quad
$^{5}$ University of California, Berkeley \quad
$^{6}$ The Chinese University of Hong Kong \quad
$^{7}$ The Hong Kong University of Science and Technology \quad
$^{8}$ The University of Hong Kong \quad
$^{9}$ Shanghai Innovation Institute \quad

\newpage

%% file: sections/appendix.tex
% Integrated appendix; the main document owns fonts, numbering and page style.
\clearpage
\begingroup
\colorlet{trinityred}{reportaccent}
\graphicspath{{appendix/figures/}}
\captionsetup{font=normalsize,justification=justified,singlelinecheck=false,hypcap=false}
\makeatletter
\setlength{\@fptop}{0pt}
\setlength{\@fpsep}{12pt}
\makeatother
\input{appendix/appendix}
\clearpage
\endgroup

%% file: appendix/appendix.tex
% Appendix (framework draft, 2026-10-20). Sections: A task tree, B capability tree, C Harness organization (C.1 Explorer, C.2 Creator),
% D evaluation details, E visual record (E.1 Create World simulations, E.2 interaction before/after RSI).
% Keep the prose short: every paragraph states facts the figures and tables carry; nothing is argued twice.
\definecolor{HarnessMuted}{HTML}{776C6C}
\newcommand{\hmiss}{%
  \multicolumn{1}{c}{\textcolor{HarnessMuted}{--}}}
\newcommand{\hfocus}[1]{\textcolor{white}{#1}}

\section{Task Tree}
\label{app:task-tree}
\begin{figure}[h]
\centering
\includegraphics[width=\linewidth]{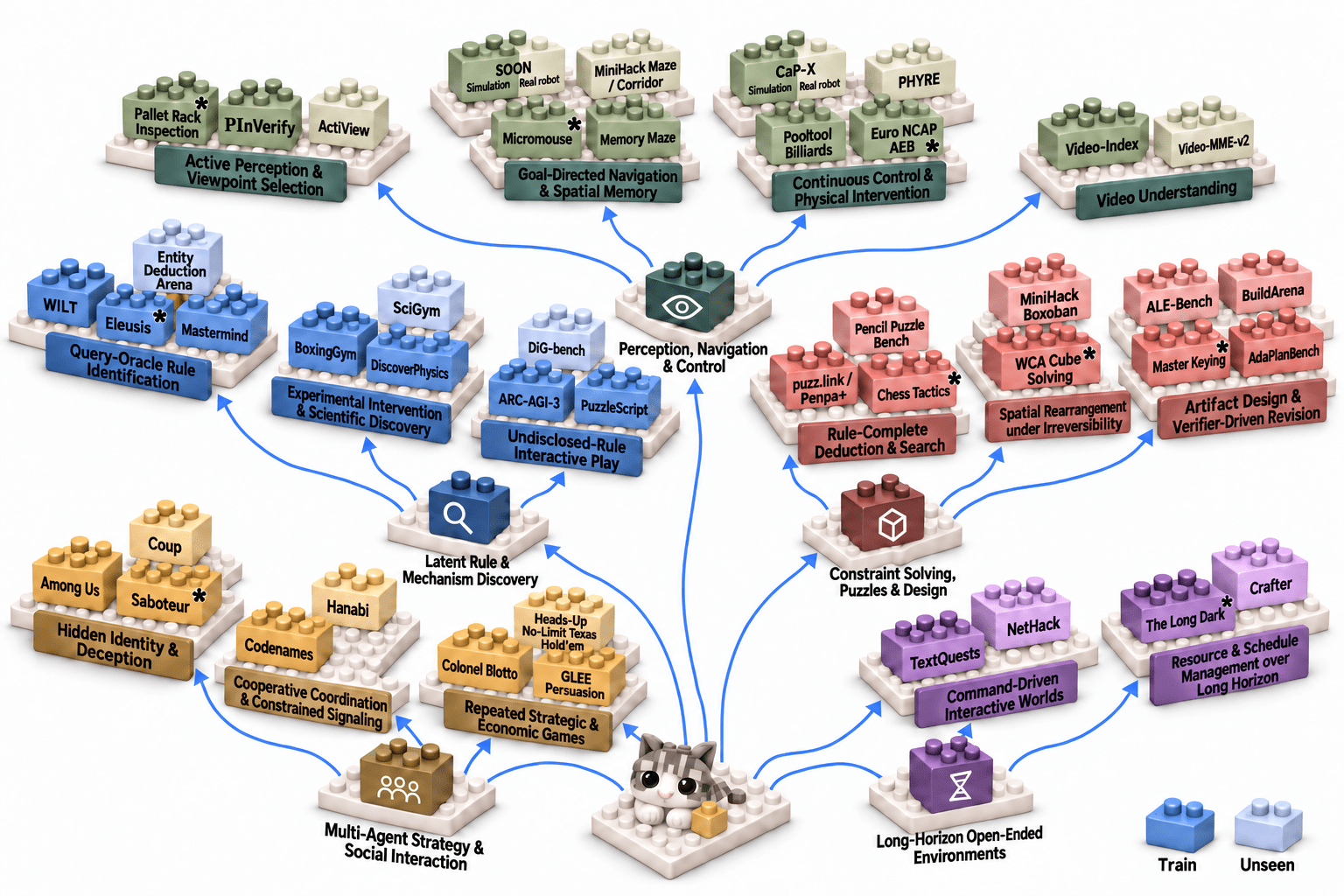}
\caption{\textbf{Task tree.} The 44 benchmarks are grouped into five domains and 15 subclasses. Training benchmarks take the colour of their domain and unseen benchmarks a lighter shade of it; SOON and CaP-X show their simulation and real-robot versions; * marks environments rebuilt from published rules.}
\label{fig:task-tree}
\end{figure}

The benchmark inventory comprises 44 benchmarks organized into five domains and 15 subclasses (Figure~\ref{fig:task-tree}): 26 training benchmarks, 16 unseen benchmarks, and two embodied benchmarks, SOON~\citep{zhu2021soon} and CaP-X~\citep{fu2026capx}, each evaluated both in simulation and on a real robot (Appendix~\ref{app:visual}). The five domains are Latent Rule \& Mechanism Discovery; Constraint Solving, Puzzles \& Design; Perception, Navigation \& Control; Multi-Agent Strategy \& Social Interaction; and Long-Horizon Open-Ended Environments. Counting the simulation version of each embodied benchmark as a training benchmark and its real-robot version as an unseen benchmark gives the 28 training and 18 unseen benchmarks of Section~\ref{sec:setup}. Every subclass contains at least one training and one unseen benchmark.

Candidate benchmarks were collected through agent web search over five types of source: research benchmarks released with code or data, public AI competitions, human contests and testing standards, published games and puzzle communities, and professional practice. Because a multimodal agent must perceive its world through whatever interface that world exposes, the inventory spans every input modality in which agentic tasks are posed, and each benchmark retains the observation protocol of its source. Each modality probes a different part of an agent's capability. Image input (photographs, rendered scenes, puzzle images, and robot camera views) tests perception from pixels, viewpoint selection, and the grounding of decisions in what is seen. Video input tests evidence acquisition over time: selecting frames within a budget, tracking events across them, and understanding what happened. Boards, grids, and maps rendered as text test spatial reasoning and state tracking over a symbolic visual state; with pixel-level perception removed, a failure can be attributed to reasoning rather than to seeing. Plain text tests the capabilities a multimodal agent relies on downstream of perception: rule discovery, experimental design, planning, negotiation, and strategic interaction. Keeping all four modalities in one inventory lets visual and text benchmarks share nodes in the capability tree (Appendix~\ref{app:capability-tree}).

Nine of the training sources publish their rules but provide no agent-facing environment: Chess Tactics~\citep{lichess2026puzzles}, Eleusis~\citep{golden2020eleusisexpress}, Euro NCAP AEB~\citep{euroncap2023aeb}, Master Keying~\citep{pulford2007locks}, Micromouse~\citep{ukmarsMicromouseRules}, Pallet Rack Inspection~\citep{en15635}, Saboteur~\citep{moyersoen2023saboteur}, The Long Dark~\citep{hinterland2017longdark}, and WCA Cube Solving~\citep{wca2025regulations} (marked * in Figure~\ref{fig:task-tree}). We rebuilt each of them through agent web search from the published rules and examples, preserving the mechanics, information access, and scoring of the source; the rebuilt environments present their state to the agent as images or symbolic text. Every unseen benchmark other than the two real-robot versions runs on its official environment or evaluation code, e.g., the BALROG harness~\citep{paglieri2025balrog}, the PHYRE simulator~\citep{bakhtin2019phyre}, the DiG-bench server~\citep{battleday2026digbench}, the ALE-Bench judge~\citep{imajuku2025alebench}, the Video-MME-v2 scorer~\citep{fu2026videommev2}, and the Besiege game of BuildArena~\citep{xia2026buildarena}.

\clearpage
\section{Capability Tree}
\label{app:capability-tree}
Independent agents propose capability categories from the task specifications and verified trajectories of the training benchmarks, a judge consolidates them, and reviewers independently check definitions, evidence, assignments, and coverage. The result is six capabilities and 30 defect types (Table~\ref{tab:defects}): a Perception \& State Tracking (PST), b Hypothesis \& Counterpart Inference (HCI), c Constraint Checking \& Feedback Repair (CFR), d Goal Commitment \& Value Trade-off (GVT), e Video Evidence Acquisition (VEA), and f Video Comprehension (VC). Each defect type is an observable failure pattern that evidences a gap in its capability. Every training record is used for capability induction, tree construction, level generation, module evolution, and the Router's training memory. Unseen benchmarks are recorded only after all Harness training has finished, from their runs with H$_0$; these records go through the same categorization procedure and are anchored to the fixed taxonomy without changing it.

\input{appendix/tables/defects.tex}

The six capabilities are parallel and non-overlapping, with no causal relation among them; ordered by the breadth of the domains each capability spans, the fixed order c, d, a, b, e, f defines the paths of the capability tree, and most benchmarks require more than one capability. The flat capability tree (Figure~\ref{fig:flat-tree}) shows this directly: each capability lists every benchmark that requires it, so a benchmark with several requirements appears under each of them.

\begin{center}
\includegraphics[width=\linewidth]{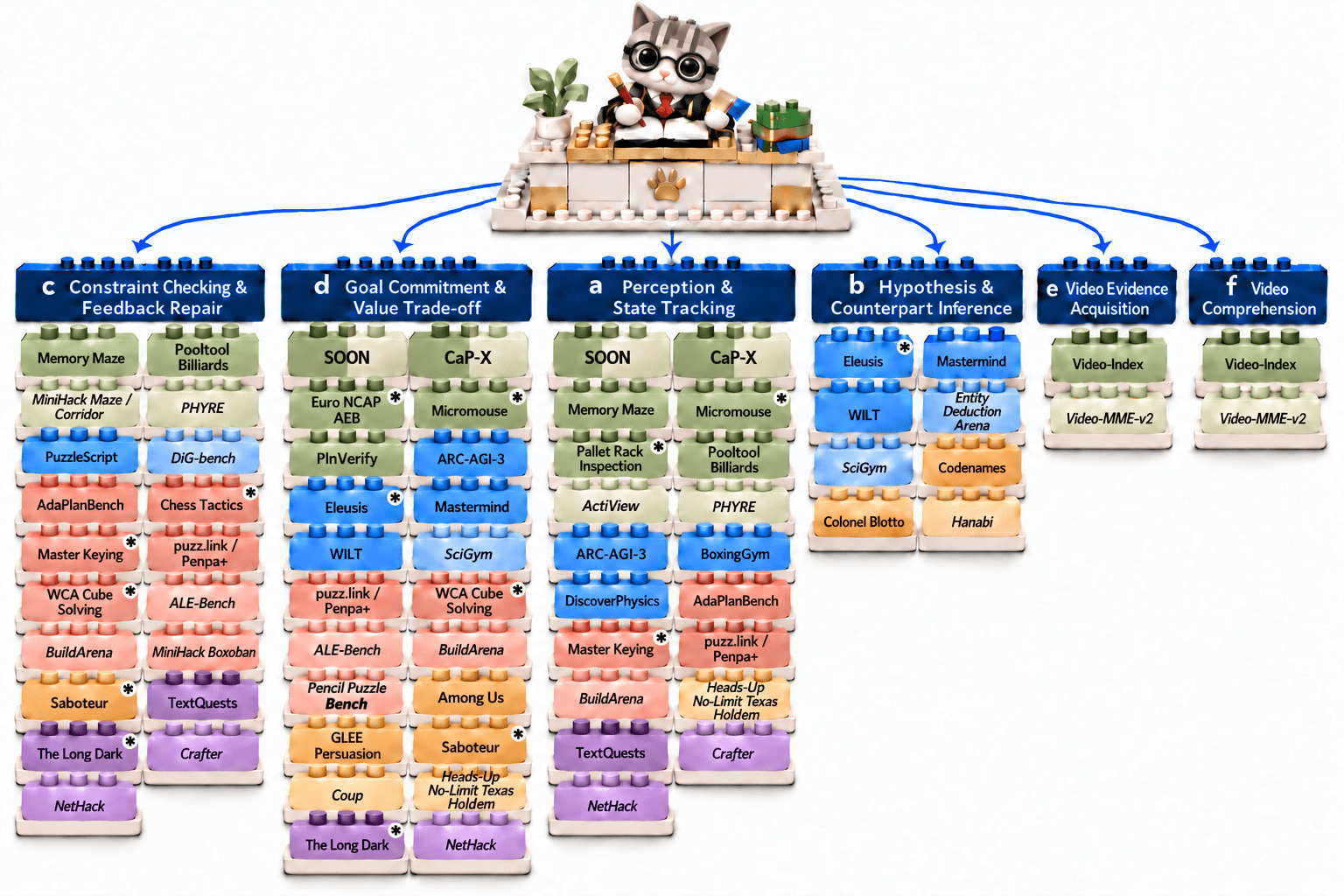}
\captionof{figure}{\textbf{Flat capability tree.} Bricks keep the colours of Figure~\ref{fig:task-tree}; unseen benchmarks are in italics; * marks rebuilt environments.}
\label{fig:flat-tree}
\end{center}

The capability tree (Figure~\ref{fig:capability-tree}) organizes the same assignment by combination: the requirement set of each training benchmark, sorted in this order, forms a path from the root, and these paths are merged along their common prefixes. Each node stands for one combination of capabilities, read along the branch from the root: for example, node c holds the benchmarks that require constraint checking alone, its child c$\to$d those that also require goal commitment, and c$\to$d$\to$a those that further require perception and state tracking. A benchmark sits at the node of its own combination, regardless of its domain. AdaPlanBench, Master Keying, Memory Maze, Pooltool Billiards, and TextQuests, for instance, come from three domains and differ widely in content, from adaptive planning and lock-system design to maze navigation, billiard shots, and text adventures; yet they share the same capability requirements and therefore all sit at c$\to$a. An edge adds a capability; it does not imply any dependency or causal relation between the two capabilities. The 11 non-root nodes define the shared structure of the Explorer and Creator Harness trees, and both evolve along the paths and structure of the capability tree (Appendix~\ref{app:harness-tree}). Unseen benchmarks are then placed at the nodes of their combinations without modifying the tree.

\begin{center}
\includegraphics[width=\linewidth]{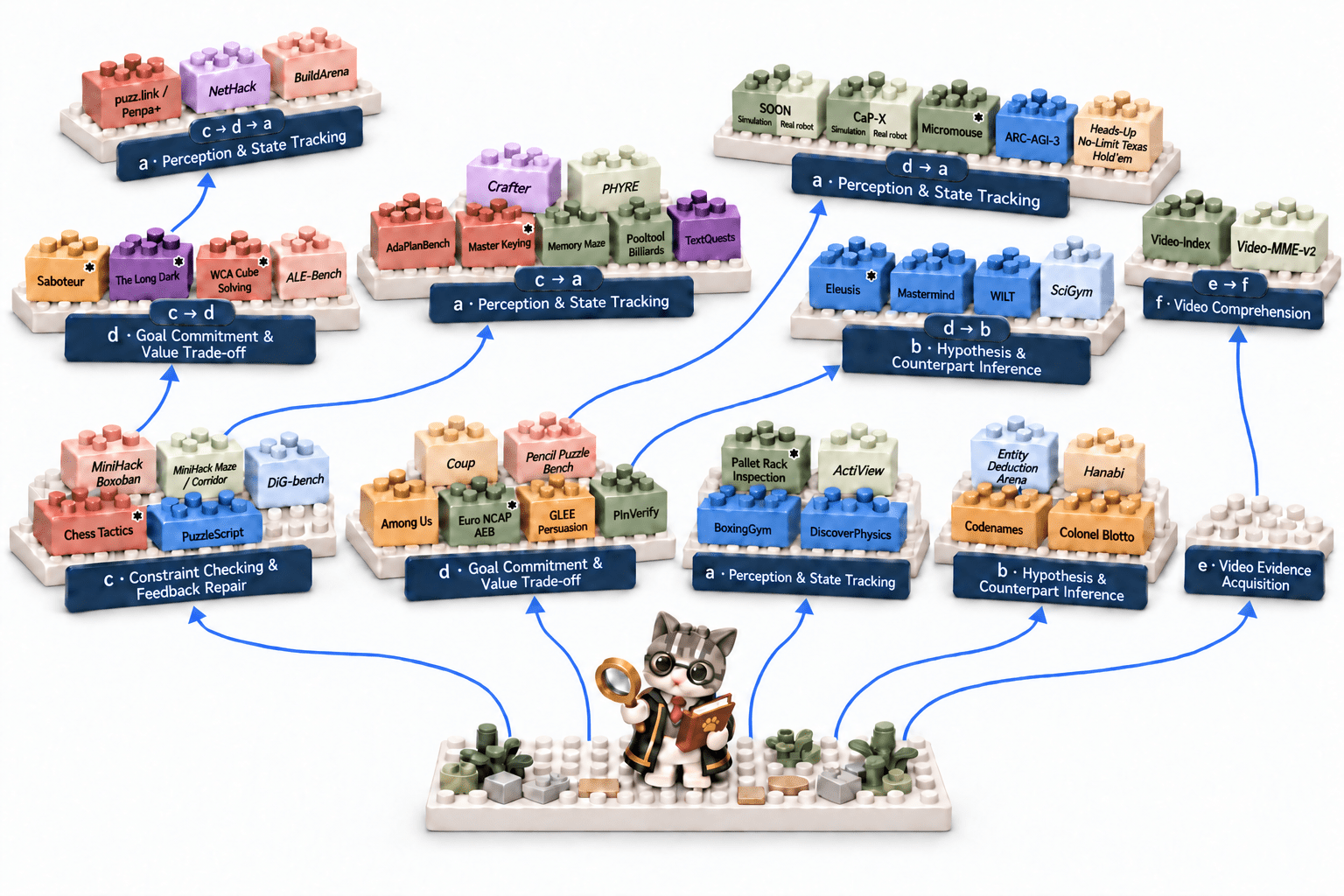}
\captionof{figure}{\textbf{Capability tree.} Each node is one combination of capabilities, read along the branch from the root, and each benchmark sits at the node of its combination. Bricks keep the colours of Figure~\ref{fig:task-tree}; unseen benchmarks are in italics; * marks rebuilt environments.}
\label{fig:capability-tree}
\end{center}

\clearpage
\section{Harness Tree}
\label{app:harness-tree}
Explorer and Creator both start from H$_0$ and evolve along the same capability tree, but they retain their learned modules differently. Explorer keeps a Harness at every node except e: each node selectively inherits its parent's modules and refines its own local modules, and the resulting Harness tree is retained for routing. Creator keeps a single shared library: within an iteration, its node Harnesses inherit updated parent modules, and at the end of the iteration the modules proposed at all nodes are consolidated with the current library into the Base Harness of the next iteration. Explorer's trajectories drive Evolver's local module revisions, whereas worlds rejected by the verifiers drive Creator's module proposals. Every module is a versioned function behind a prompt-facing interface and is installed only after the host checks its evidence, interface, and version lineage.

\subsection{Explorer Harness Tree}
\label{app:explorer}
Explorer plays every benchmark with the frozen backbone (Appendix~\ref{app:evaluation}), and each instance uses the best-scoring compatible node Harness under the Oracle setting. Every node Harness starts from H$_0$, which has no learned module, and combines the inherited modules at their retained versions with the node's local modules; a benchmark at c$\to$d$\to$a therefore uses the modules of c, c$\to$d, and c$\to$d$\to$a, subject to selective inheritance (Figure~\ref{fig:explorer}). At each node, Explorer plays the admitted training levels with these modules, and Evolver compares successful and failed trajectories to create or revise local modules; the host validates evidence, ownership, interfaces, and version lineage before every update, and ancestor modules are preserved. An updated parent is inherited only if it achieves strictly more host-verified task completions than the previously inherited modules under the same validation tasks, trial count, and budget; otherwise the previous modules are kept. A local module overrides an inherited module bound to the same identifier. Module calls take public inputs and return structured advice without advancing the environment, with at most two calls per step and none forced.

The catalog below lists the 29 modules M01--M29 of the Explorer Harness tree after the final iteration. Each was generated and revised during RSI; the modules are grouped by capability family: \textbf{a Perception \& State Tracking:} \begingroup\leavevmode\setlength{\arxiviconheight}{1.25em}\raisebox{-0.20\arxiviconheight}{\includegraphics[height=\arxiviconheight]{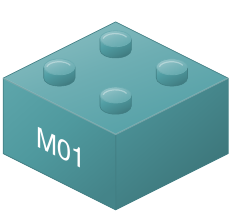}}\endgroup\,Object Evidence Ledger and Rule Classification; \begingroup\leavevmode\setlength{\arxiviconheight}{1.25em}\raisebox{-0.20\arxiviconheight}{\includegraphics[height=\arxiviconheight]{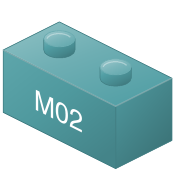}}\endgroup\,Complete Inspection Coverage and Report Reconciliation; \begingroup\leavevmode\setlength{\arxiviconheight}{1.25em}\raisebox{-0.20\arxiviconheight}{\includegraphics[height=\arxiviconheight]{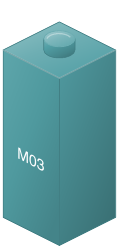}}\endgroup\,Identity Binding Across Reversed Views; \begingroup\leavevmode\setlength{\arxiviconheight}{1.25em}\raisebox{-0.20\arxiviconheight}{\includegraphics[height=\arxiviconheight]{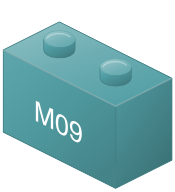}}\endgroup\,Relational Orders, Identity Binding and Position Transforms; \begingroup\leavevmode\setlength{\arxiviconheight}{1.25em}\raisebox{-0.20\arxiviconheight}{\includegraphics[height=\arxiviconheight]{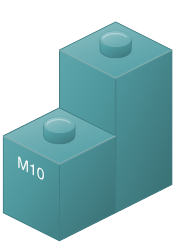}}\endgroup\,Sealed Observation Archives and Unordered Collection Tours; \begingroup\leavevmode\setlength{\arxiviconheight}{1.25em}\raisebox{-0.20\arxiviconheight}{\includegraphics[height=\arxiviconheight]{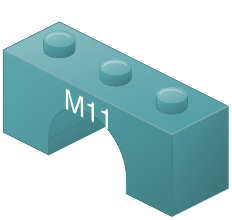}}\endgroup\,Ordered Prefix Validation and Capacity-Constrained Batch Planning; \begingroup\leavevmode\setlength{\arxiviconheight}{1.25em}\raisebox{-0.20\arxiviconheight}{\includegraphics[height=\arxiviconheight]{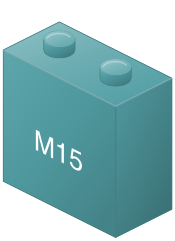}}\endgroup\,Execution Receipt Verification and State-Difference Memory; \begingroup\leavevmode\setlength{\arxiviconheight}{1.25em}\raisebox{-0.20\arxiviconheight}{\includegraphics[height=\arxiviconheight]{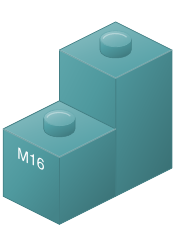}}\endgroup\,Stage Revision, Deduction Retraction and Execution Memory; \begingroup\leavevmode\setlength{\arxiviconheight}{1.25em}\raisebox{-0.20\arxiviconheight}{\includegraphics[height=\arxiviconheight]{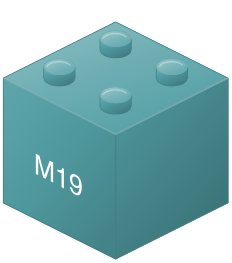}}\endgroup\,Target Collection, Current-Phase Routing and Frontier Exploration; \begingroup\leavevmode\setlength{\arxiviconheight}{1.25em}\raisebox{-0.20\arxiviconheight}{\includegraphics[height=\arxiviconheight]{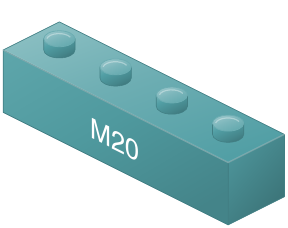}}\endgroup\,Product-State Routing for Ordered Collection; \begingroup\leavevmode\setlength{\arxiviconheight}{1.25em}\raisebox{-0.20\arxiviconheight}{\includegraphics[height=\arxiviconheight]{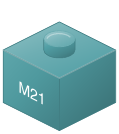}}\endgroup\,Phase-Dependent Edge Repair and Static Corridor Reuse. \textbf{b Hypothesis \& Counterpart Inference:} \begingroup\leavevmode\setlength{\arxiviconheight}{1.25em}\raisebox{-0.20\arxiviconheight}{\includegraphics[height=\arxiviconheight]{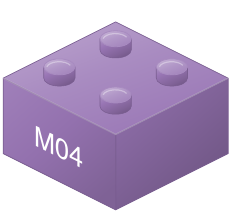}}\endgroup\,Parity Reversal and Two-Period Signal Selection; \begingroup\leavevmode\setlength{\arxiviconheight}{1.25em}\raisebox{-0.20\arxiviconheight}{\includegraphics[height=\arxiviconheight]{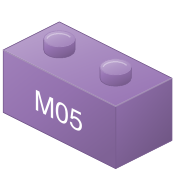}}\endgroup\,Contingent Policy Search with Endogenous Signals; \begingroup\leavevmode\setlength{\arxiviconheight}{1.25em}\raisebox{-0.20\arxiviconheight}{\includegraphics[height=\arxiviconheight]{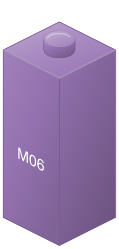}}\endgroup\,Opponent Inference, Allocation Scoring and Archive Value; \begingroup\leavevmode\setlength{\arxiviconheight}{1.25em}\raisebox{-0.20\arxiviconheight}{\includegraphics[height=\arxiviconheight]{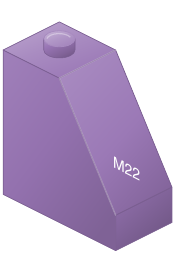}}\endgroup\,Candidate Ledger, Dynamic State Alignment and Dual-Objective Probing; \begingroup\leavevmode\setlength{\arxiviconheight}{1.25em}\raisebox{-0.20\arxiviconheight}{\includegraphics[height=\arxiviconheight]{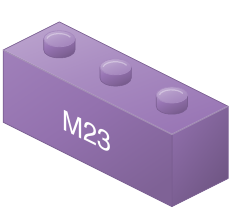}}\endgroup\,Decomposing Repeated Symbols into Multiplicity and Position. \textbf{c Constraint Checking \& Feedback Repair:} \begingroup\leavevmode\setlength{\arxiviconheight}{1.25em}\raisebox{-0.20\arxiviconheight}{\includegraphics[height=\arxiviconheight]{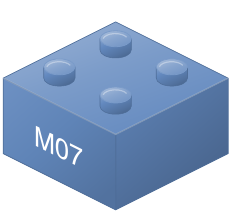}}\endgroup\,Joint Gate Planning, Proposal Validation and Milestone Scheduling; \begingroup\leavevmode\setlength{\arxiviconheight}{1.25em}\raisebox{-0.20\arxiviconheight}{\includegraphics[height=\arxiviconheight]{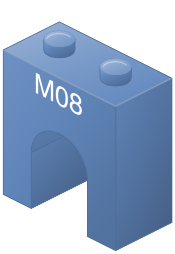}}\endgroup\,Return Reachability Under Asymmetric Gates. \textbf{d Goal Commitment \& Value Trade-off:} \begingroup\leavevmode\setlength{\arxiviconheight}{1.25em}\raisebox{-0.20\arxiviconheight}{\includegraphics[height=\arxiviconheight]{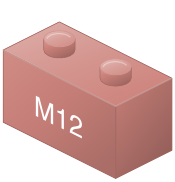}}\endgroup\,Investment Timing and Scored Value Trade-offs; \begingroup\leavevmode\setlength{\arxiviconheight}{1.25em}\raisebox{-0.20\arxiviconheight}{\includegraphics[height=\arxiviconheight]{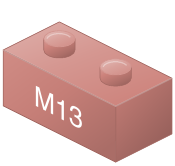}}\endgroup\,Resource Reserves and Delivery Opportunity Trade-offs; \begingroup\leavevmode\setlength{\arxiviconheight}{1.25em}\raisebox{-0.20\arxiviconheight}{\includegraphics[height=\arxiviconheight]{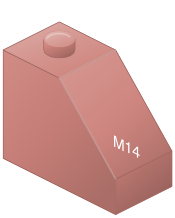}}\endgroup\,Goal Verification, Stopping Decisions and Plan Value; \begingroup\leavevmode\setlength{\arxiviconheight}{1.25em}\raisebox{-0.20\arxiviconheight}{\includegraphics[height=\arxiviconheight]{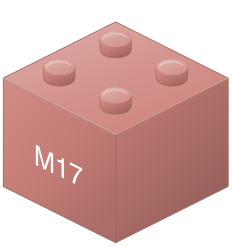}}\endgroup\,Evidence Verification, Counterexample Stopping and Submission Budget; \begingroup\leavevmode\setlength{\arxiviconheight}{1.25em}\raisebox{-0.20\arxiviconheight}{\includegraphics[height=\arxiviconheight]{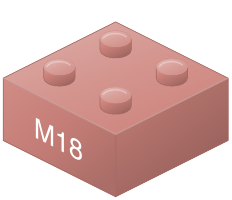}}\endgroup\,Relation Verification Within One Object and Coordinate Frame; \begingroup\leavevmode\setlength{\arxiviconheight}{1.25em}\raisebox{-0.20\arxiviconheight}{\includegraphics[height=\arxiviconheight]{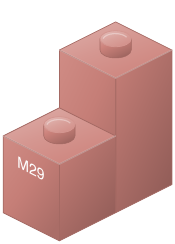}}\endgroup\,Fixed Goal Conditions, Evidence Ledger and Bounded Commit Review. \textbf{e Video Evidence Acquisition:} \begingroup\leavevmode\setlength{\arxiviconheight}{1.25em}\raisebox{-0.20\arxiviconheight}{\includegraphics[height=\arxiviconheight]{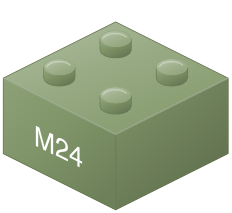}}\endgroup\,Adaptive Frame, Region and Audio Evidence Acquisition; \begingroup\leavevmode\setlength{\arxiviconheight}{1.25em}\raisebox{-0.20\arxiviconheight}{\includegraphics[height=\arxiviconheight]{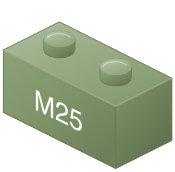}}\endgroup\,Delivered Evidence Coverage and Acquisition Budget Verification; \begingroup\leavevmode\setlength{\arxiviconheight}{1.25em}\raisebox{-0.20\arxiviconheight}{\includegraphics[height=\arxiviconheight]{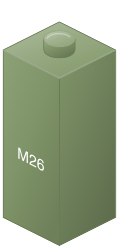}}\endgroup\,Evidence Gap Acquisition and Unproductive Retry Suppression. \textbf{f Video Comprehension:} \begingroup\leavevmode\setlength{\arxiviconheight}{1.25em}\raisebox{-0.20\arxiviconheight}{\includegraphics[height=\arxiviconheight]{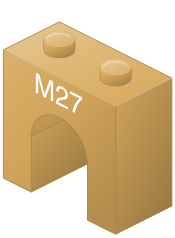}}\endgroup\,Handoff Episode Aggregation, Event Ledger and Endpoint Audit; \begingroup\leavevmode\setlength{\arxiviconheight}{1.25em}\raisebox{-0.20\arxiviconheight}{\includegraphics[height=\arxiviconheight]{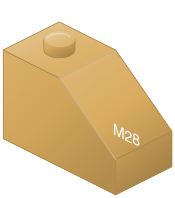}}\endgroup\,Candidate Identity Association Across Frames.

\begin{center}
\includegraphics[width=0.62\linewidth]{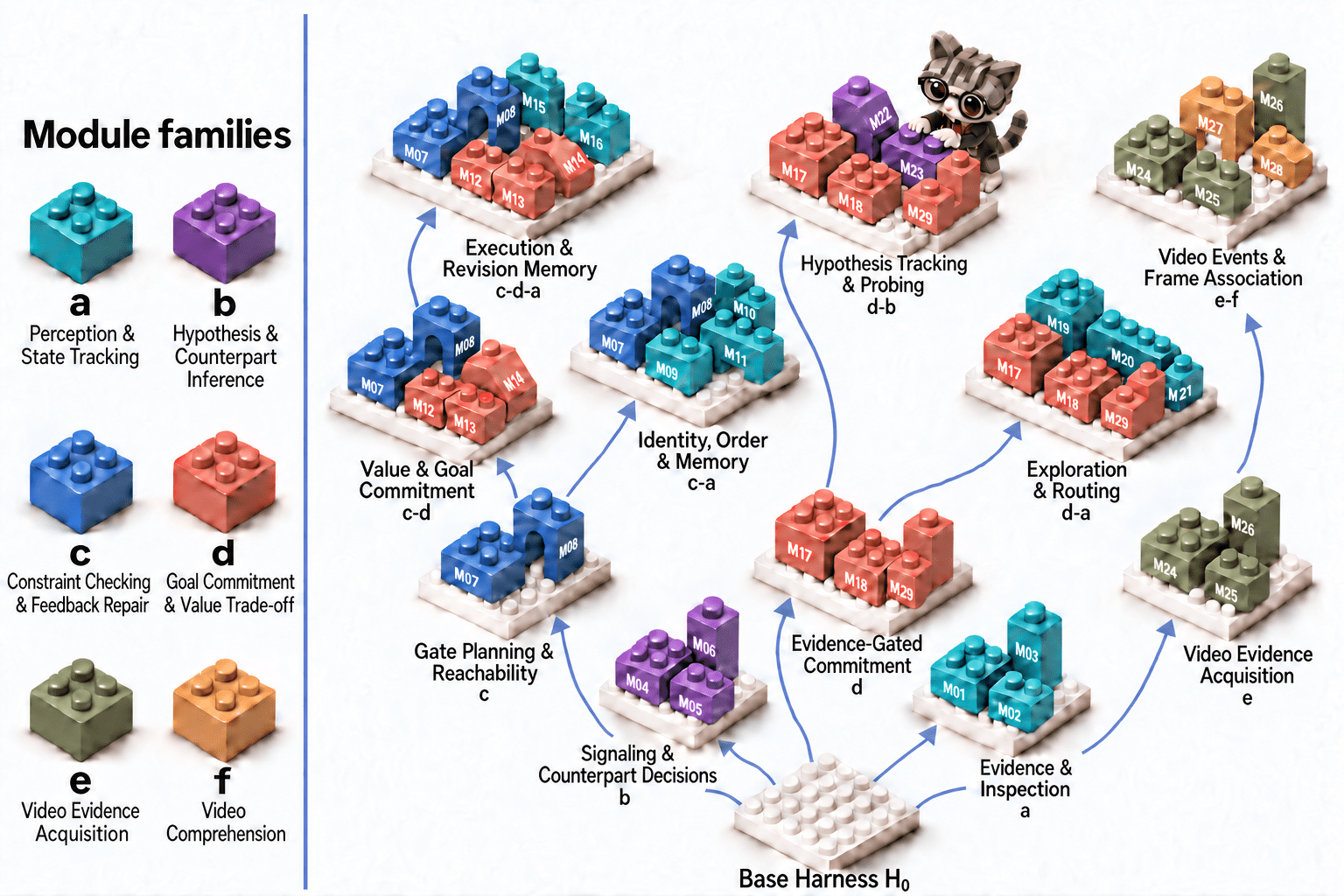}
\captionof{figure}{\textbf{Explorer Harness tree.} Every node Harness starts from H$_0$ and combines the modules it selectively inherits with its own local modules. Brick colours mark capability families, and labels give module IDs.}
\label{fig:explorer}
\end{center}

\subsection{Creator Shared Harness}
\label{app:creator}
Creator builds the simplified training worlds of each node with a coding harness, created per node and iteration, with a workspace for the world and an evidence directory outside it. Its Base Harness at the first iteration is H$_0$ extended with the pre-RSI modules listed below. Within each iteration, Creator visits the nodes top-down: the children of the root start from the Base Harness, and every other node inherits its parent's updated modules. Each node keeps its own guidance, one construction note, and one public action budget per iteration. Verifier-detected failures trigger local evolution, and at the end of the iteration the validated proposals of all nodes are consolidated with the current library in learning order (Equation~\ref{eq:creator_consolidation}): validated revisions replace their registered predecessors; among new modules, distinct capabilities are retained, partial overlaps are merged to preserve complementary functionality, and near-duplicates keep the earlier-learned module and discard the later one. The consolidated library becomes the Base Harness of the next iteration; it is shared by all nodes and exposed as a single tool that runs one module per call. The host, not the model, supplies the inputs of each call: the build digest, the submission allowance, and, for a rule module, a measurement of the build. Fields that the host does not supply remain unknown, and missing evidence never passes. Before submission, the host reruns the applicable audits and blocks any missing, stale, or conflicting result; Creator returns errors in the frozen task design or verifier contract to their owner instead of rewriting them.

In the module names, a witness is a recorded route run on the current build; the commitment cost is the number of extra actions that a wrong commitment costs over the direct route; the screening band is the performance range into which a world must fall; and a neutralized control is the previous version with the changed difficulty axis reset. The final Creator module library holds 25 modules (Figure~\ref{fig:creator}). Eleven are pre-RSI modules, audits that form the initial module library of the Creator Harness and apply the host's formulas to recorded routes, claims, edits, or submissions: \begingroup\leavevmode\setlength{\arxiviconheight}{1.25em}\raisebox{-0.20\arxiviconheight}{\includegraphics[height=\arxiviconheight]{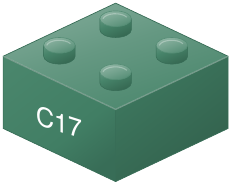}}\endgroup\,Commitment Cost Arithmetic Audit, \begingroup\leavevmode\setlength{\arxiviconheight}{1.25em}\raisebox{-0.20\arxiviconheight}{\includegraphics[height=\arxiviconheight]{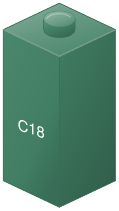}}\endgroup\,Witness and Action Ceiling Audit, \begingroup\leavevmode\setlength{\arxiviconheight}{1.25em}\raisebox{-0.20\arxiviconheight}{\includegraphics[height=\arxiviconheight]{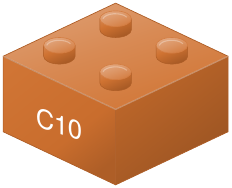}}\endgroup\,Difficulty Axis Range and Feasibility Audit, \begingroup\leavevmode\setlength{\arxiviconheight}{1.25em}\raisebox{-0.20\arxiviconheight}{\includegraphics[height=\arxiviconheight]{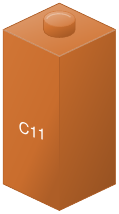}}\endgroup\,Screening Band Direction Diagnosis, \begingroup\leavevmode\setlength{\arxiviconheight}{1.25em}\raisebox{-0.20\arxiviconheight}{\includegraphics[height=\arxiviconheight]{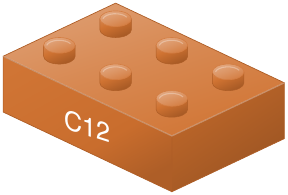}}\endgroup\,Layout Novelty Ledger, \begingroup\leavevmode\setlength{\arxiviconheight}{1.25em}\raisebox{-0.20\arxiviconheight}{\includegraphics[height=\arxiviconheight]{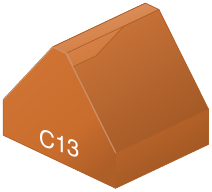}}\endgroup\,Effective Difficulty Change Audit, \begingroup\leavevmode\setlength{\arxiviconheight}{1.25em}\raisebox{-0.20\arxiviconheight}{\includegraphics[height=\arxiviconheight]{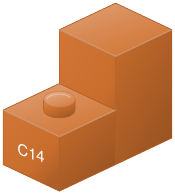}}\endgroup\,Single Change Attribution Audit, \begingroup\leavevmode\setlength{\arxiviconheight}{1.25em}\raisebox{-0.20\arxiviconheight}{\includegraphics[height=\arxiviconheight]{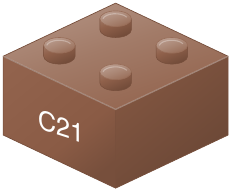}}\endgroup\,Repair Scope Gate, \begingroup\leavevmode\setlength{\arxiviconheight}{1.25em}\raisebox{-0.20\arxiviconheight}{\includegraphics[height=\arxiviconheight]{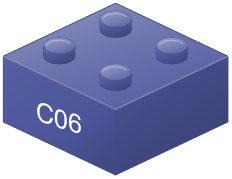}}\endgroup\,Claim and Execution Evidence Census, \begingroup\leavevmode\setlength{\arxiviconheight}{1.25em}\raisebox{-0.20\arxiviconheight}{\includegraphics[height=\arxiviconheight]{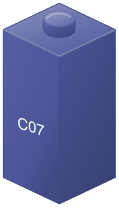}}\endgroup\,Bounded Search Declaration Audit, and \begingroup\leavevmode\setlength{\arxiviconheight}{1.25em}\raisebox{-0.20\arxiviconheight}{\includegraphics[height=\arxiviconheight]{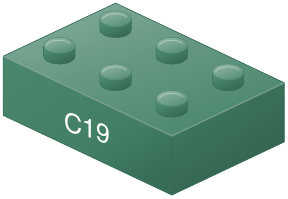}}\endgroup\,Submission Budget and Repeated Rejection Guard. The other 14 are rule modules, predicate programs over one kind of host measurement learned from rejected worlds, each listed with the nodes at which it was formed: \begingroup\leavevmode\setlength{\arxiviconheight}{1.25em}\raisebox{-0.20\arxiviconheight}{\includegraphics[height=\arxiviconheight]{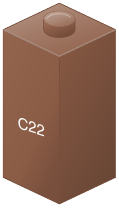}}\endgroup\,Production and Private Witness Isolation (d), \begingroup\leavevmode\setlength{\arxiviconheight}{1.25em}\raisebox{-0.20\arxiviconheight}{\includegraphics[height=\arxiviconheight]{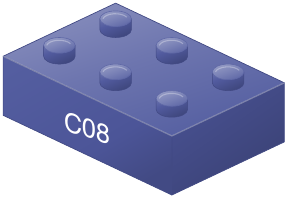}}\endgroup\,Accepted Entry and Terminal Consistency (d$\to$a), \begingroup\leavevmode\setlength{\arxiviconheight}{1.25em}\raisebox{-0.20\arxiviconheight}{\includegraphics[height=\arxiviconheight]{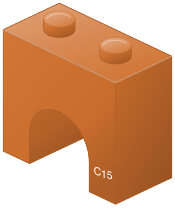}}\endgroup\,Reasoning Change and Direction Audit (a, c, d$\to$a, e), \begingroup\leavevmode\setlength{\arxiviconheight}{1.25em}\raisebox{-0.20\arxiviconheight}{\includegraphics[height=\arxiviconheight]{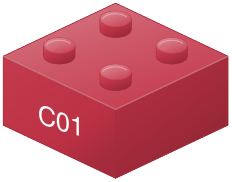}}\endgroup\,Public Visual State and Probe Coverage Audit (a, b, c), \begingroup\leavevmode\setlength{\arxiviconheight}{1.25em}\raisebox{-0.20\arxiviconheight}{\includegraphics[height=\arxiviconheight]{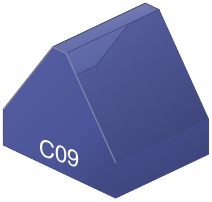}}\endgroup\,Required Composition and Negative Route Audit (b, d$\to$b), \begingroup\leavevmode\setlength{\arxiviconheight}{1.25em}\raisebox{-0.20\arxiviconheight}{\includegraphics[height=\arxiviconheight]{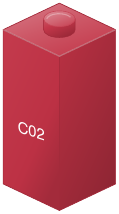}}\endgroup\,Target Attributes, Classes, and Acceptance Audit (a, b, c, d$\to$b), \begingroup\leavevmode\setlength{\arxiviconheight}{1.25em}\raisebox{-0.20\arxiviconheight}{\includegraphics[height=\arxiviconheight]{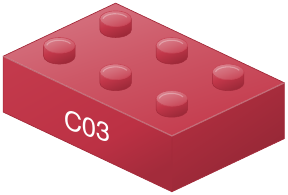}}\endgroup\,Controlled Identity and Frame-State Alignment Audit (b, c, d$\to$a), \begingroup\leavevmode\setlength{\arxiviconheight}{1.25em}\raisebox{-0.20\arxiviconheight}{\includegraphics[height=\arxiviconheight]{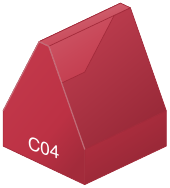}}\endgroup\,Action Polarity and Coordinate Basis Audit (a, b), \begingroup\leavevmode\setlength{\arxiviconheight}{1.25em}\raisebox{-0.20\arxiviconheight}{\includegraphics[height=\arxiviconheight]{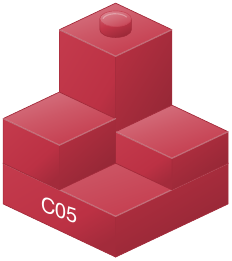}}\endgroup\,Interaction Coverage and Decision-Time Visibility Audit (b, d), \begingroup\leavevmode\setlength{\arxiviconheight}{1.25em}\raisebox{-0.20\arxiviconheight}{\includegraphics[height=\arxiviconheight]{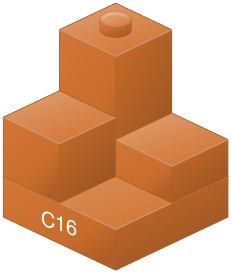}}\endgroup\,Neutralized Control Initialization and Completion Audit (b, d$\to$a), \begingroup\leavevmode\setlength{\arxiviconheight}{1.25em}\raisebox{-0.20\arxiviconheight}{\includegraphics[height=\arxiviconheight]{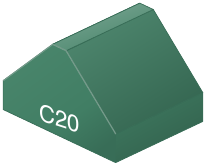}}\endgroup\,Meaningful Exploratory Witness Pricing (e), \begingroup\leavevmode\setlength{\arxiviconheight}{1.25em}\raisebox{-0.20\arxiviconheight}{\includegraphics[height=\arxiviconheight]{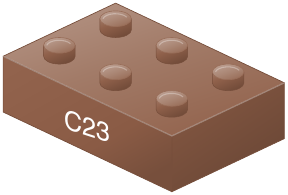}}\endgroup\,Submission Artifact Completeness (a), \begingroup\leavevmode\setlength{\arxiviconheight}{1.25em}\raisebox{-0.20\arxiviconheight}{\includegraphics[height=\arxiviconheight]{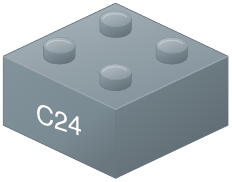}}\endgroup\,Task Design Schema and Axis Type Handoff (c, d$\to$a), and \begingroup\leavevmode\setlength{\arxiviconheight}{1.25em}\raisebox{-0.20\arxiviconheight}{\includegraphics[height=\arxiviconheight]{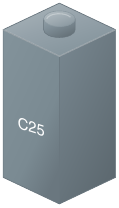}}\endgroup\,Authoritative Verifier Contract Handoff (all 11 nodes).

\noindent\begin{minipage}{\linewidth}\centering   % image and caption kept on one page
\includegraphics[width=0.68\linewidth]{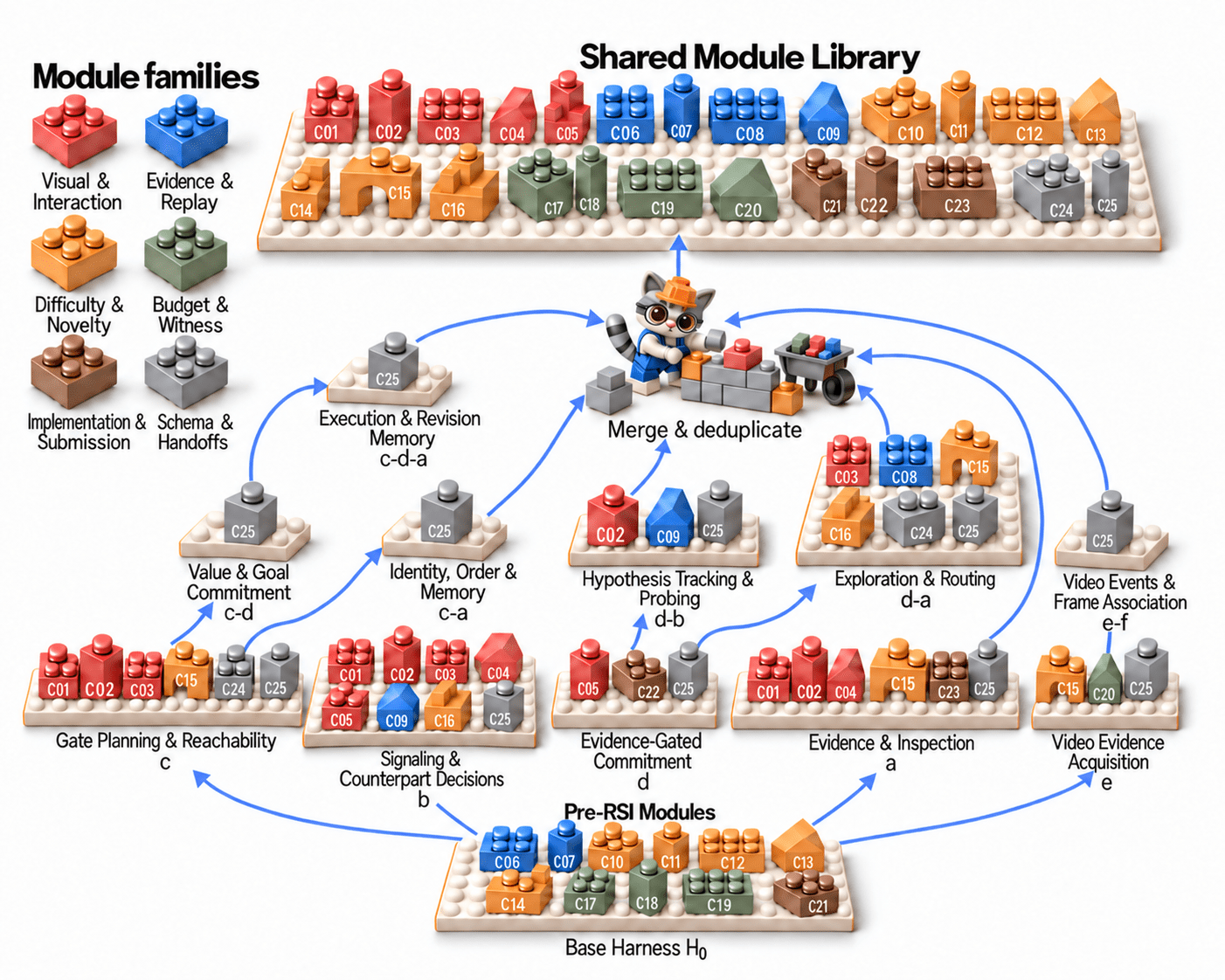}
\captionof{figure}{\textbf{Creator shared Harness.} The root is the Base Harness H$_0$ with the pre-RSI modules. Each node shows the modules proposed or revised at that node. The modules of all nodes are merged and deduplicated into one shared library of 25 modules. Brick colours group the library by function.}
\label{fig:creator}
\end{minipage}

\clearpage
\section{Evaluation Details}
\label{app:evaluation}
Every run uses the same frozen backbone, Claude Opus 5~\citep{anthropic2026opus5} with adaptive thinking at low effort, and the benchmark's native driver, which validates each action, returns the observation, and assigns the outcome. The Base Harness H$_0$ plays with no learned module and no skills; for H$^{(i)}$, each instance uses the best-scoring compatible node Harness under the Oracle setting, together with evaluation-time skills summarized from five preparation trials of that node Harness on the benchmark. Levels, action budgets, and scoring are fixed before the first run and shared by every Harness. Each benchmark reports performance on a 0--100 scale: task-completion, win, or solve rates where episodes pass or fail; the published bounded score where one exists; and, for ALE-Bench, the rank percentile among human entrants. The Codex Harness~\citep{openai2025codex} and the DeepSeek Harness~\citep{deepseek2026harness} ran the 16 unseen benchmarks other than the real-robot versions with the same backbone and settings through their native tool interfaces. Table~\ref{tab:results} reports every score together with each benchmark's measure, number of distinct instances, and number of episodes; * marks environments rebuilt from source rules through agent web search. Train benchmarks use their fixed instance sets (Section~\ref{sec:setup}), and ARC-AGI-3 covers the 25 games of the public set. Six unseen benchmarks run on a fixed subset of their official set: Pencil Puzzle Bench uses the official Golden 30 set, PHYRE the 50 test tasks of fold 0 in the within-template B tier, and Video-MME-v2 a stratified subset of 400 of the 800 videos (1,600 questions); DiG-bench uses 13 of the 21 public games, all games of tiers 1--3 and the first game of each of tiers 4--7; ALE-Bench uses 26 of the 40 problems and SciGym 126 of the 137 small systems, those on which the backbone's safety layer did not refuse the task inputs. NetHack is scored after a budget of 500 steps.

\input{appendix/tables/results.tex}

\clearpage
\section{Visual Record}
\label{app:visual}
This appendix records what the agent sees and does. Appendix~\ref{app:interaction} pairs H$_0$ and H$^{(4)}$ on the same configuration of each of the 44 benchmarks and shows where the two Harnesses part in reasoning and in outcome. Appendix~\ref{app:levels} follows the 11 capability nodes through four iterations of training levels, from the benchmarks that feed a node to the modules its levels produce. Appendix~\ref{app:embodied} presents SOON and CaP-X in their reconstructed simulators and on the real robots, with the runs of H$_0$ and H$^{(4)}$ in both.

\subsection{Interaction Process}
\label{app:interaction}
Figures~\ref{fig:pair-first}--\ref{fig:pair-last} show the actual decisions and outcomes of the agent on each of the 44 benchmarks (26 training, 16 unseen, and the two embodied benchmarks in simulation): on the same configuration, the different Harnesses lead H$_0$ and H$^{(4)}$ to different reasoning and judgements, and hence to different results.

{\captionsetup{font=normalsize,skip=2pt}%   pair figures flow inline; each image stays with its caption
\noindent\begin{minipage}{\linewidth}\centering
\includegraphics[width=\linewidth]{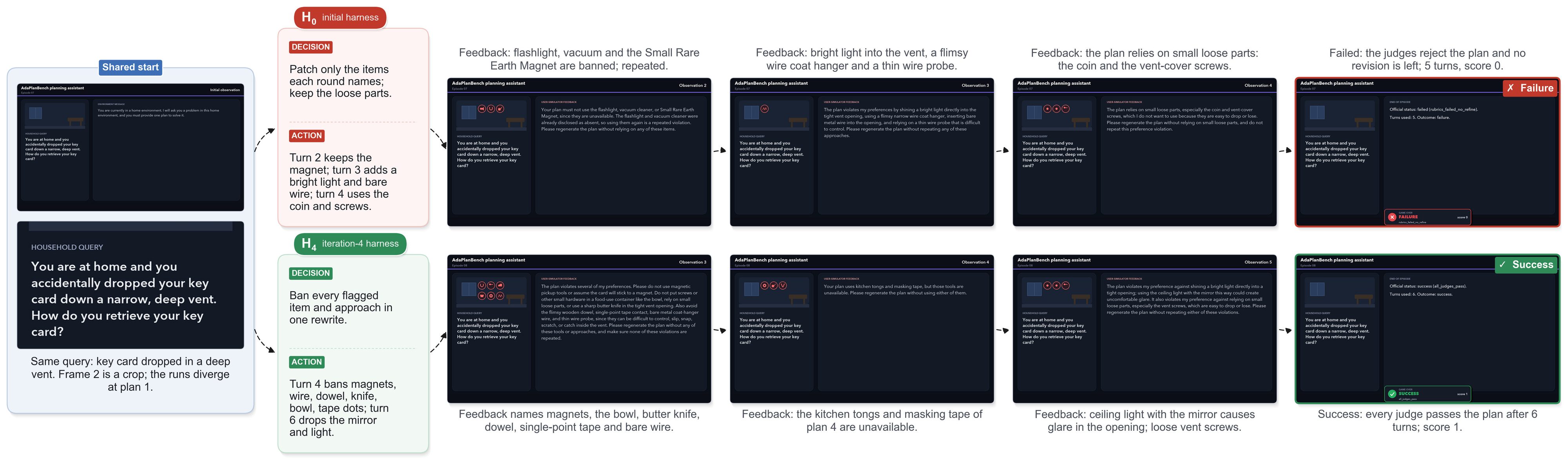}
\captionof{figure}{\textbf{AdaPlanBench (train)}}\label{fig:pair-first}
\end{minipage}\par\vspace{4pt}
\noindent\begin{minipage}{\linewidth}\centering
\includegraphics[width=\linewidth]{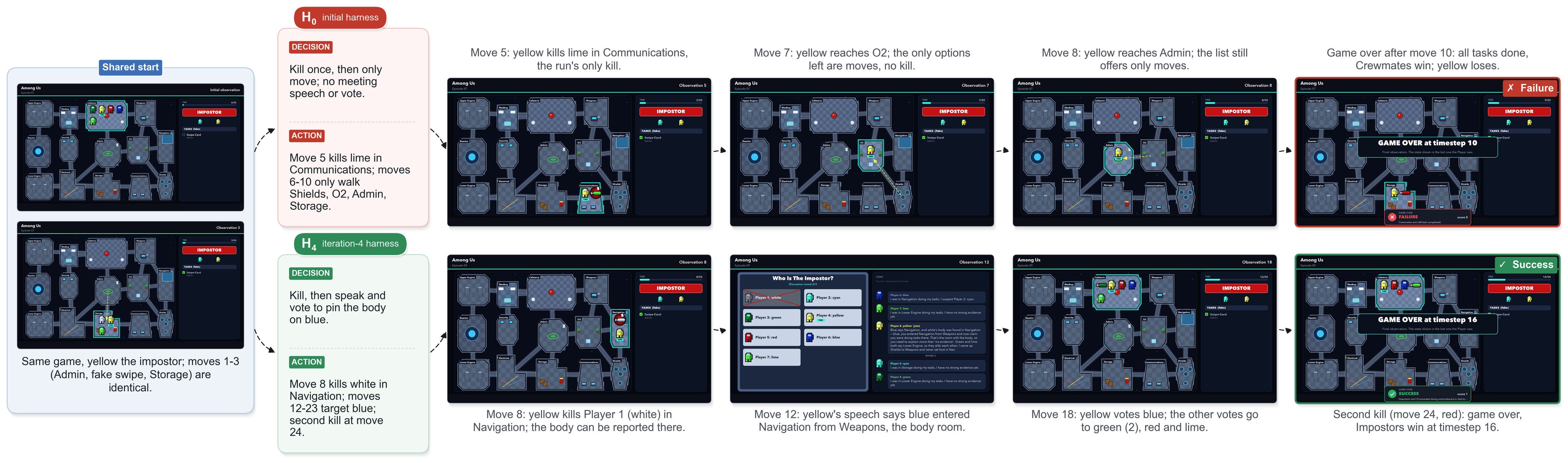}
\captionof{figure}{\textbf{Among Us (train)}}
\end{minipage}\par\vspace{4pt}
\noindent\begin{minipage}{\linewidth}\centering
\includegraphics[width=\linewidth]{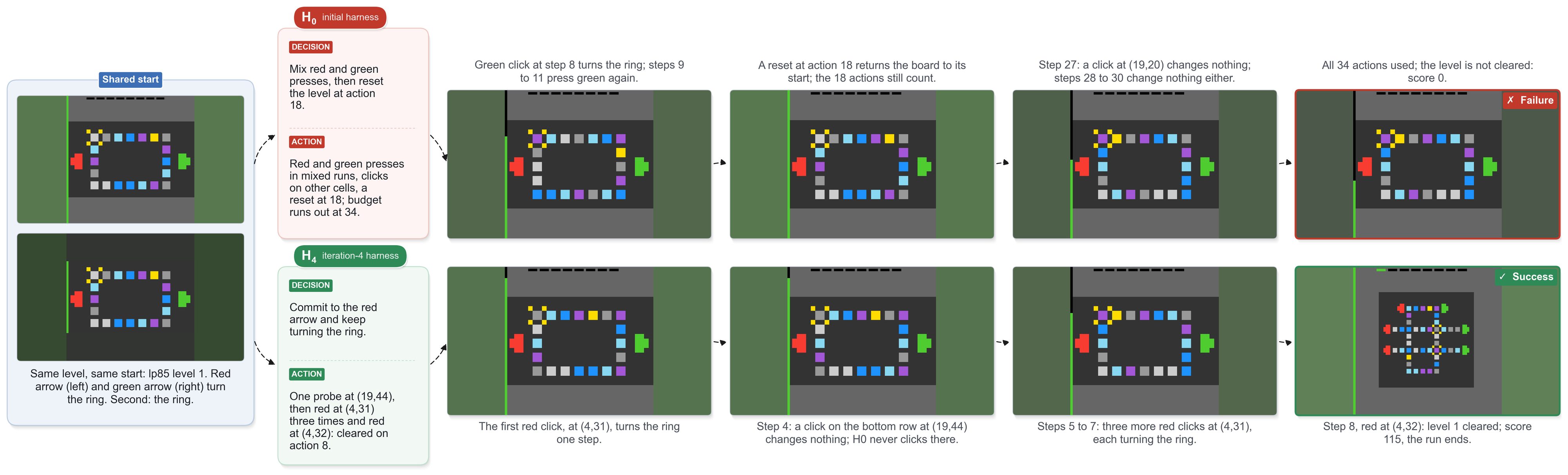}
\captionof{figure}{\textbf{ARC-AGI-3 (train)}}
\end{minipage}\par\vspace{4pt}
\noindent\begin{minipage}{\linewidth}\centering
\includegraphics[width=\linewidth]{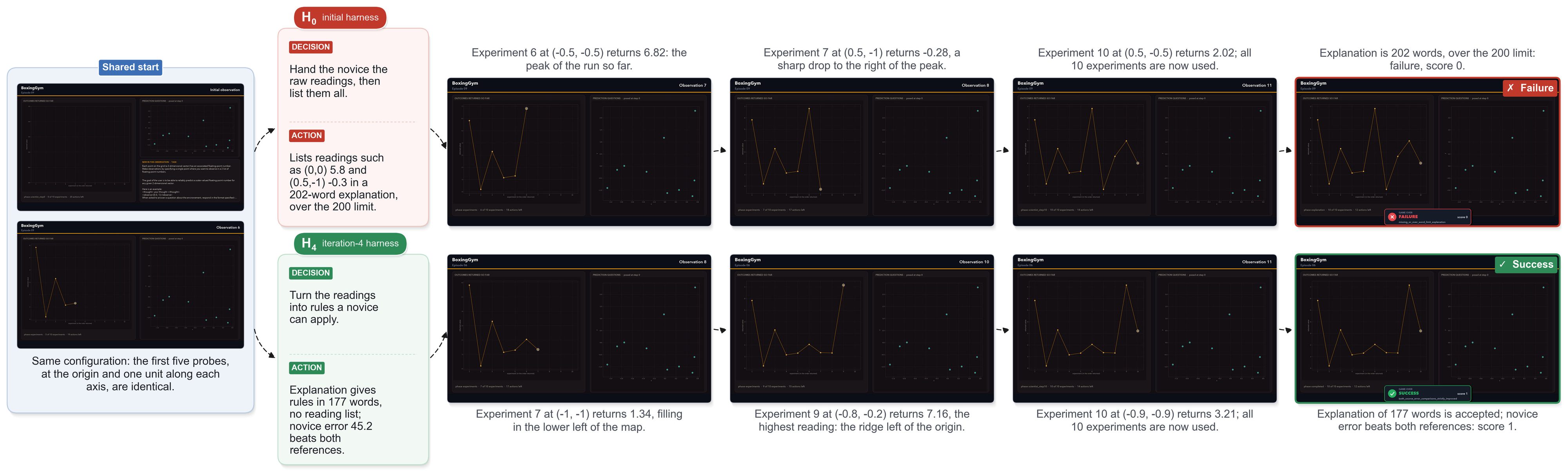}
\captionof{figure}{\textbf{BoxingGym (train)}}
\end{minipage}\par\vspace{4pt}
\noindent\begin{minipage}{\linewidth}\centering
\includegraphics[width=\linewidth]{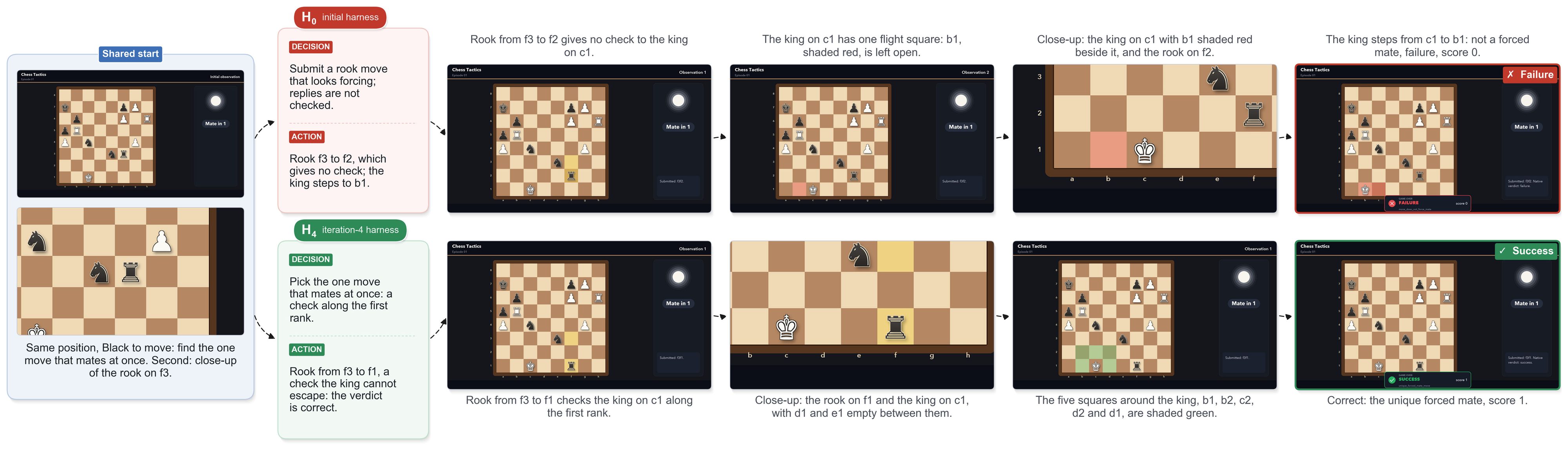}
\captionof{figure}{\textbf{Chess Tactics (train)}}
\end{minipage}\par\vspace{4pt}
\noindent\begin{minipage}{\linewidth}\centering
\includegraphics[width=\linewidth]{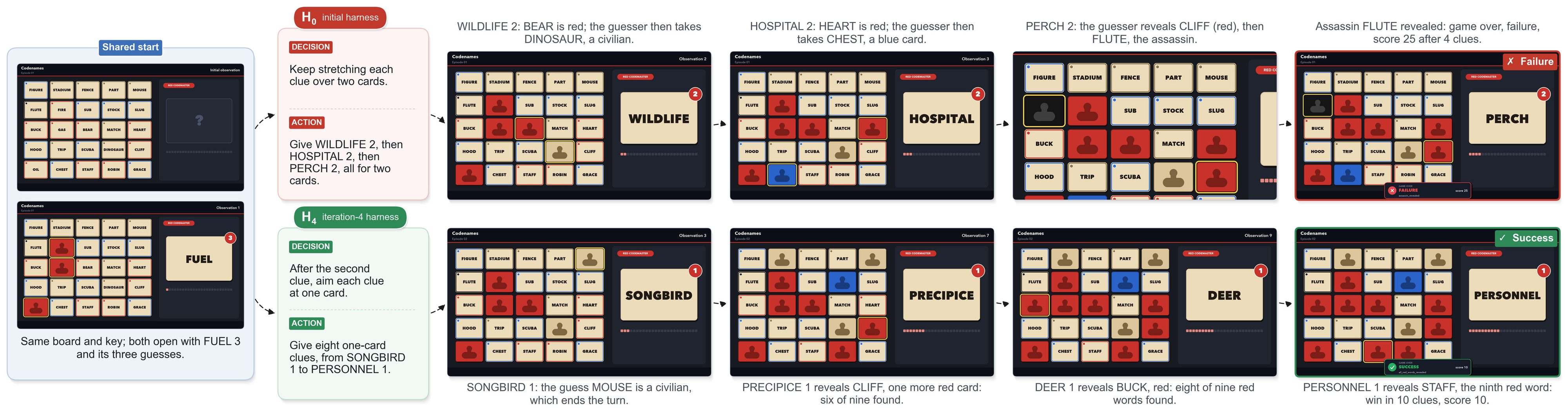}
\captionof{figure}{\textbf{Codenames (train)}}
\end{minipage}\par\vspace{4pt}
\noindent\begin{minipage}{\linewidth}\centering
\includegraphics[width=\linewidth]{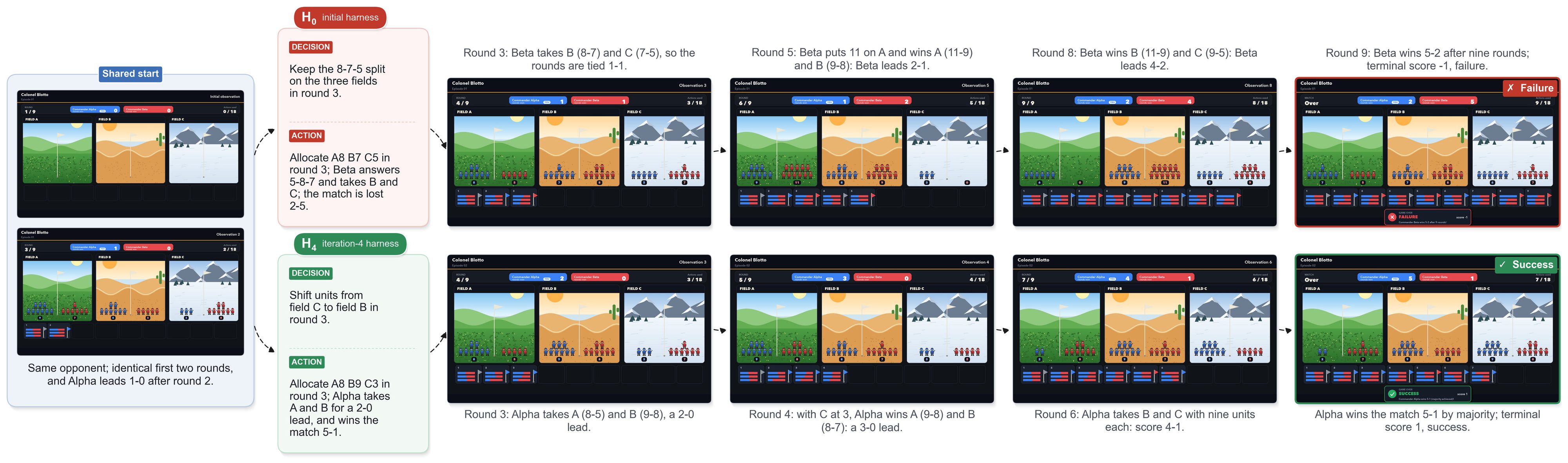}
\captionof{figure}{\textbf{Colonel Blotto (train)}}
\end{minipage}\par\vspace{4pt}
\noindent\begin{minipage}{\linewidth}\centering
\includegraphics[width=\linewidth]{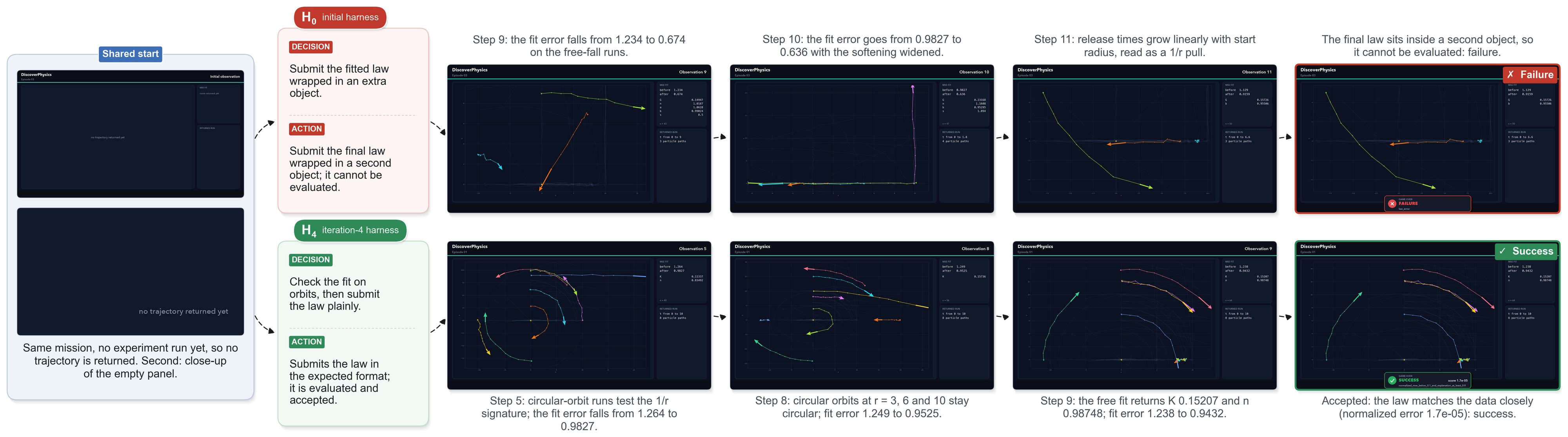}
\captionof{figure}{\textbf{DiscoverPhysics (train)}}
\end{minipage}\par\vspace{4pt}
\noindent\begin{minipage}{\linewidth}\centering
\includegraphics[width=\linewidth]{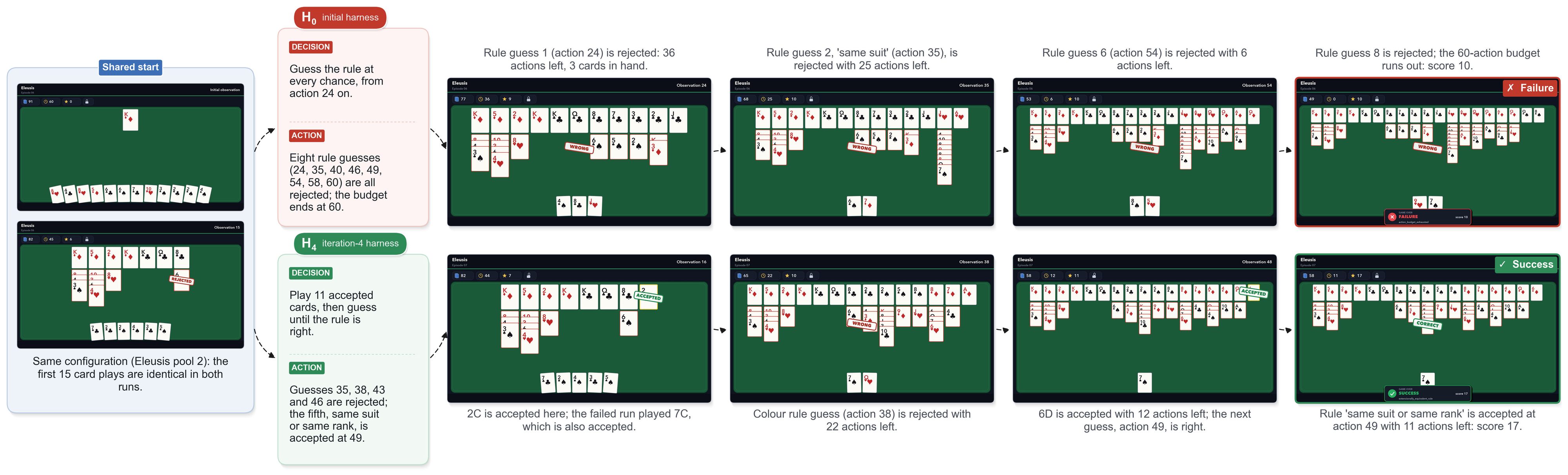}
\captionof{figure}{\textbf{Eleusis (train)}}
\end{minipage}\par\vspace{4pt}
\noindent\begin{minipage}{\linewidth}\centering
\includegraphics[width=\linewidth]{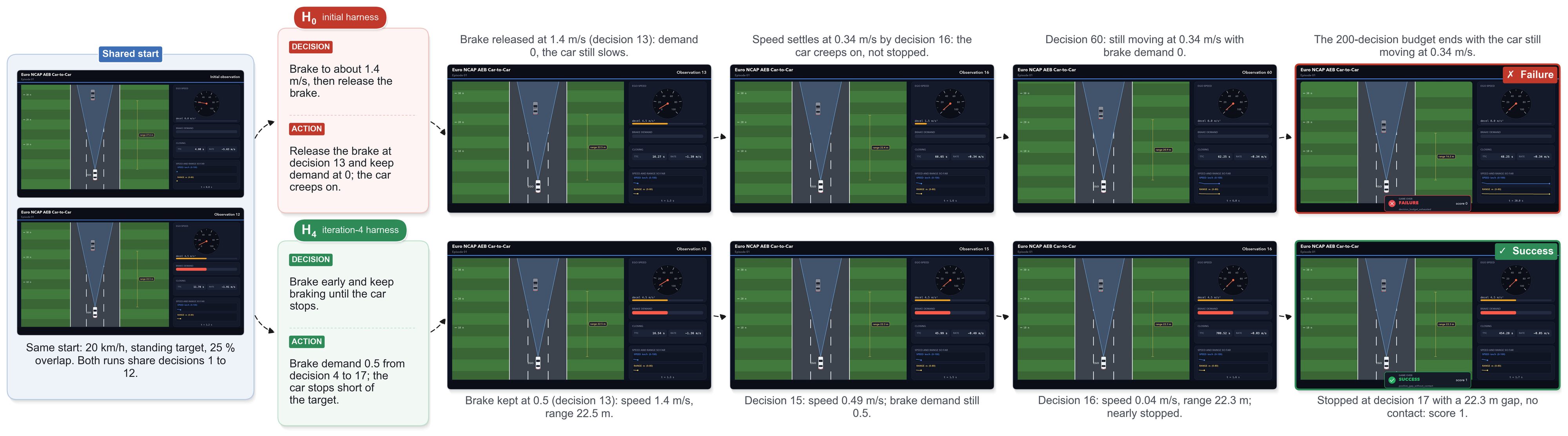}
\captionof{figure}{\textbf{Euro NCAP AEB (train)}}
\end{minipage}\par\vspace{4pt}
\noindent\begin{minipage}{\linewidth}\centering
\includegraphics[width=\linewidth]{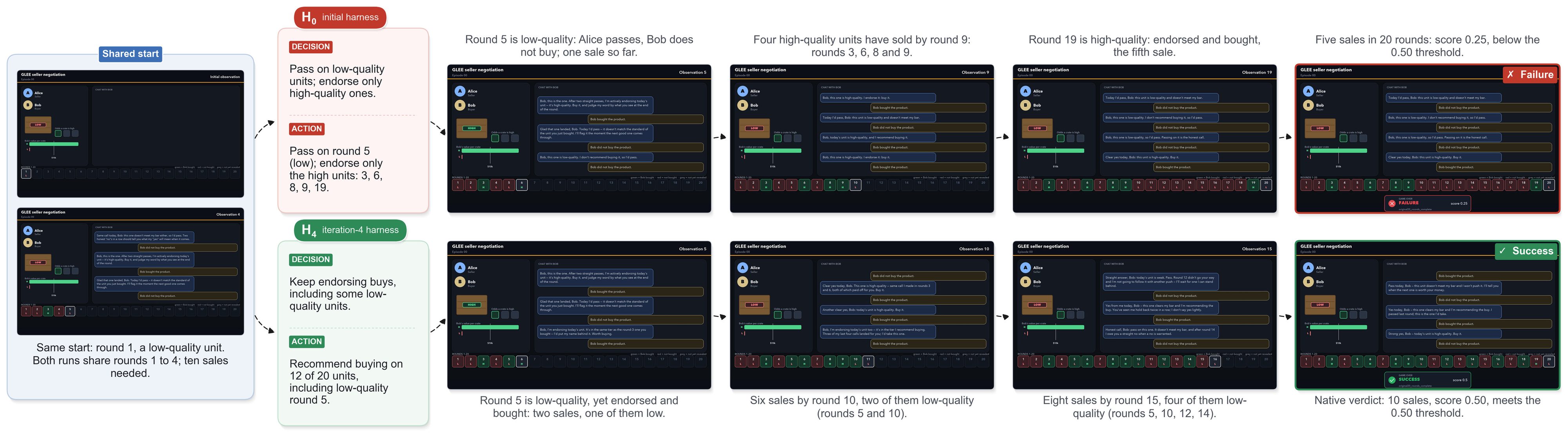}
\captionof{figure}{\textbf{GLEE Persuasion (train)}}
\end{minipage}\par\vspace{4pt}
\noindent\begin{minipage}{\linewidth}\centering
\includegraphics[width=\linewidth]{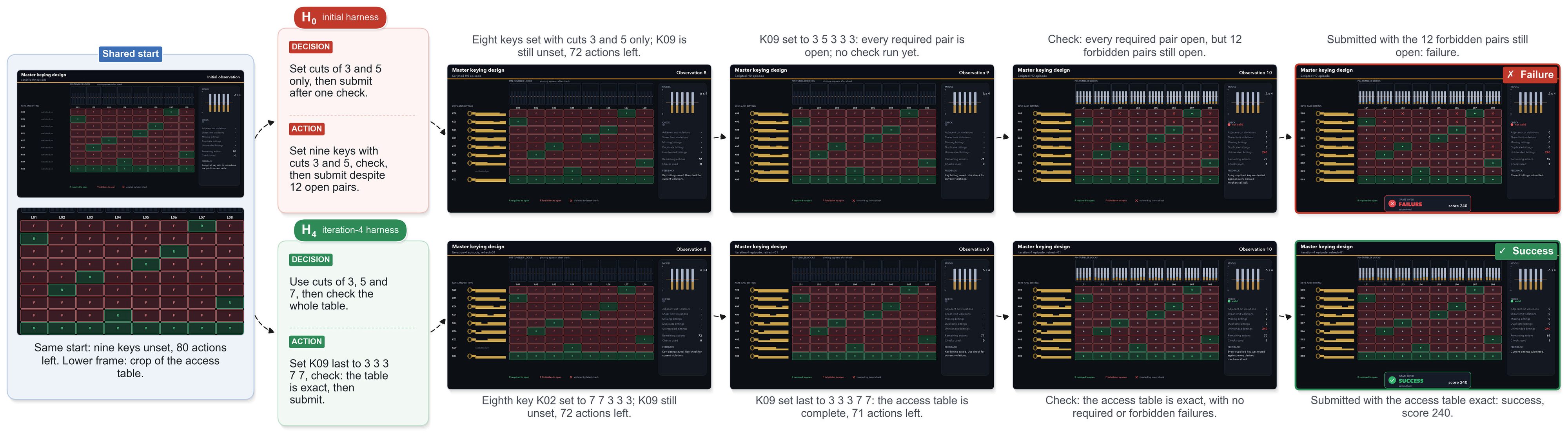}
\captionof{figure}{\textbf{Master Keying (train)}}
\end{minipage}\par\vspace{4pt}
\noindent\begin{minipage}{\linewidth}\centering
\includegraphics[width=\linewidth]{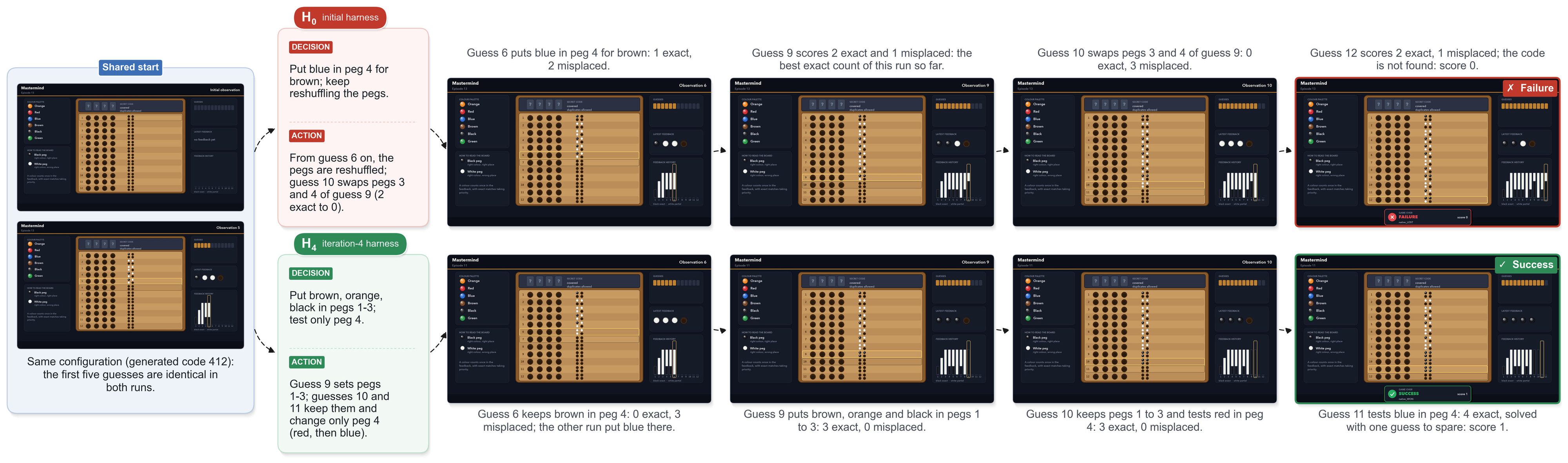}
\captionof{figure}{\textbf{Mastermind (train)}}
\end{minipage}\par\vspace{4pt}
\noindent\begin{minipage}{\linewidth}\centering
\includegraphics[width=\linewidth]{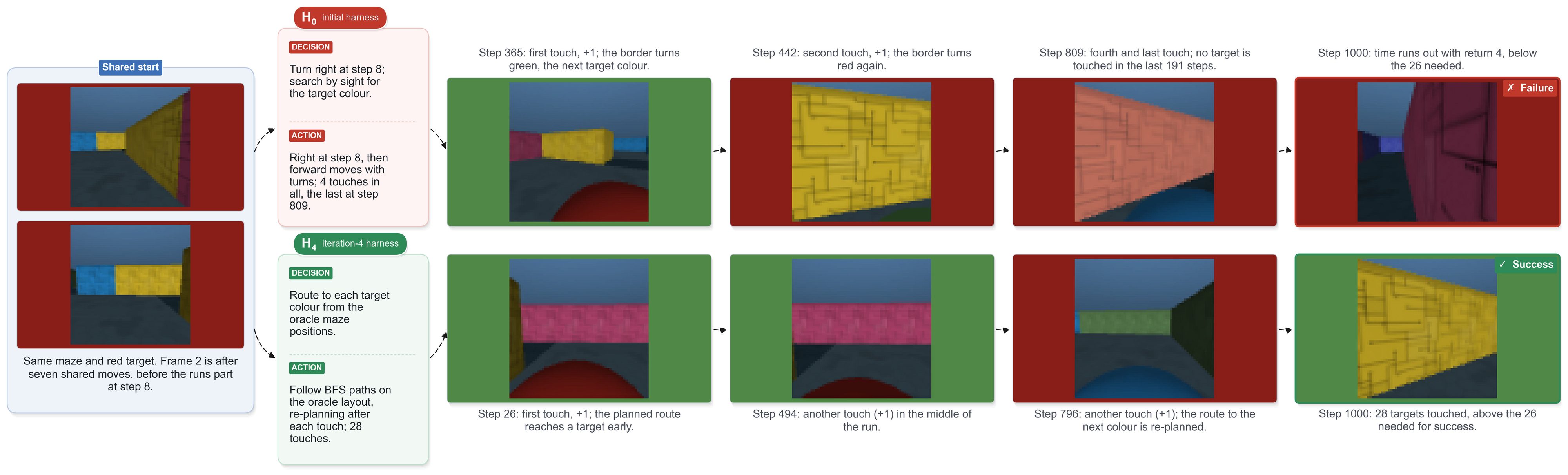}
\captionof{figure}{\textbf{Memory Maze (train)}}
\end{minipage}\par\vspace{4pt}
\noindent\begin{minipage}{\linewidth}\centering
\includegraphics[width=\linewidth]{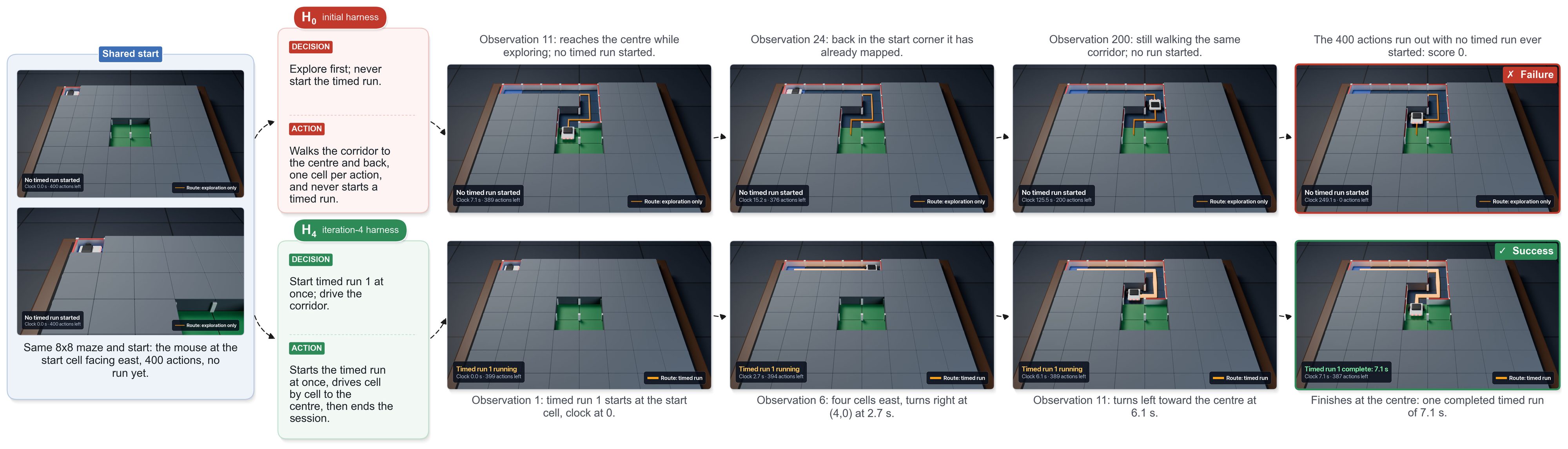}
\captionof{figure}{\textbf{Micromouse (train)}}
\end{minipage}\par\vspace{4pt}
\noindent\begin{minipage}{\linewidth}\centering
\includegraphics[width=\linewidth]{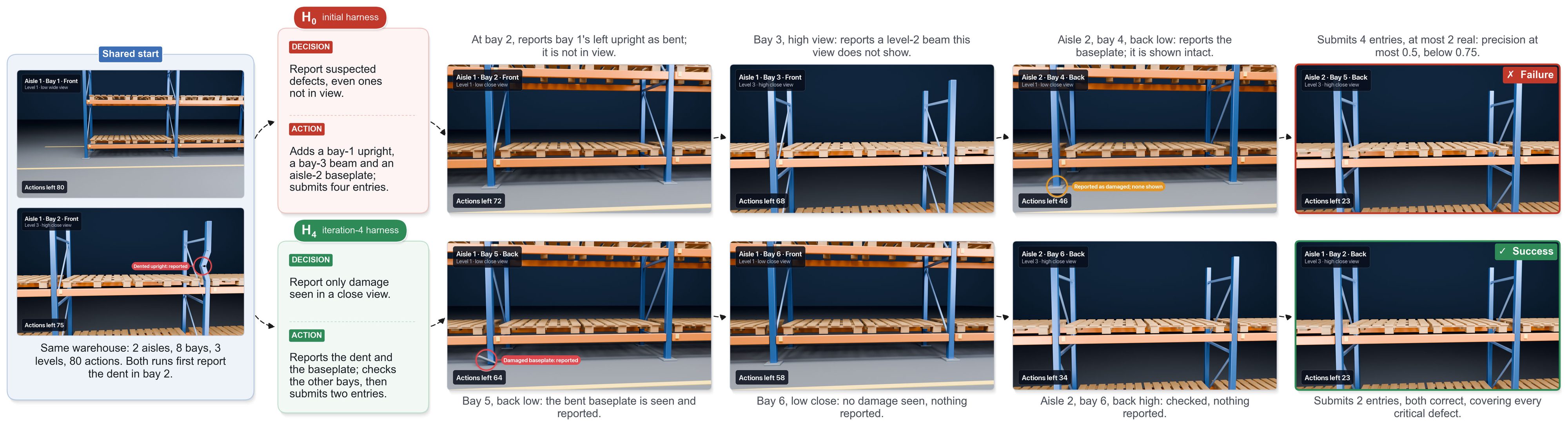}
\captionof{figure}{\textbf{Pallet Rack Inspection (train)}}
\end{minipage}\par\vspace{4pt}
\noindent\begin{minipage}{\linewidth}\centering
\includegraphics[width=\linewidth]{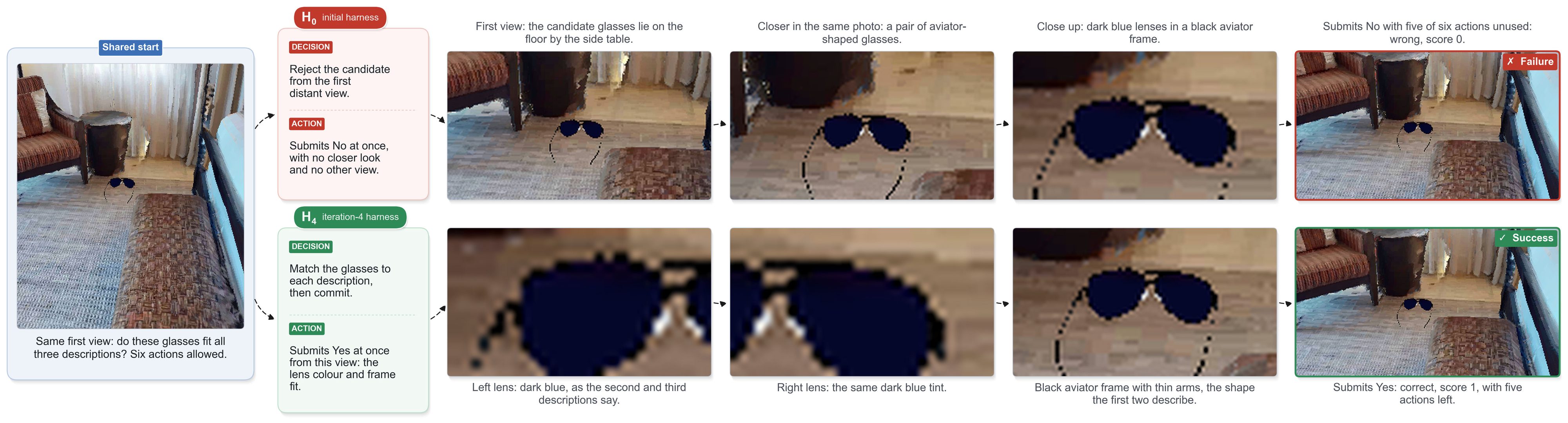}
\captionof{figure}{\textbf{PInVerify (train)}}
\end{minipage}\par\vspace{4pt}
\noindent\begin{minipage}{\linewidth}\centering
\includegraphics[width=\linewidth]{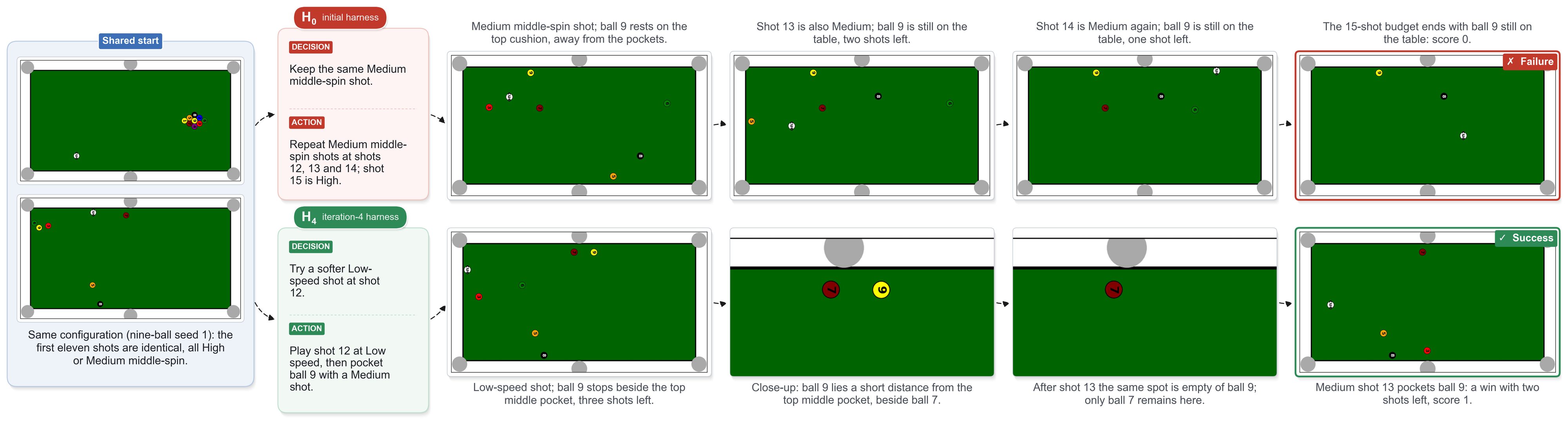}
\captionof{figure}{\textbf{Pooltool Billiards (train)}}
\end{minipage}\par\vspace{4pt}
\noindent\begin{minipage}{\linewidth}\centering
\includegraphics[width=\linewidth]{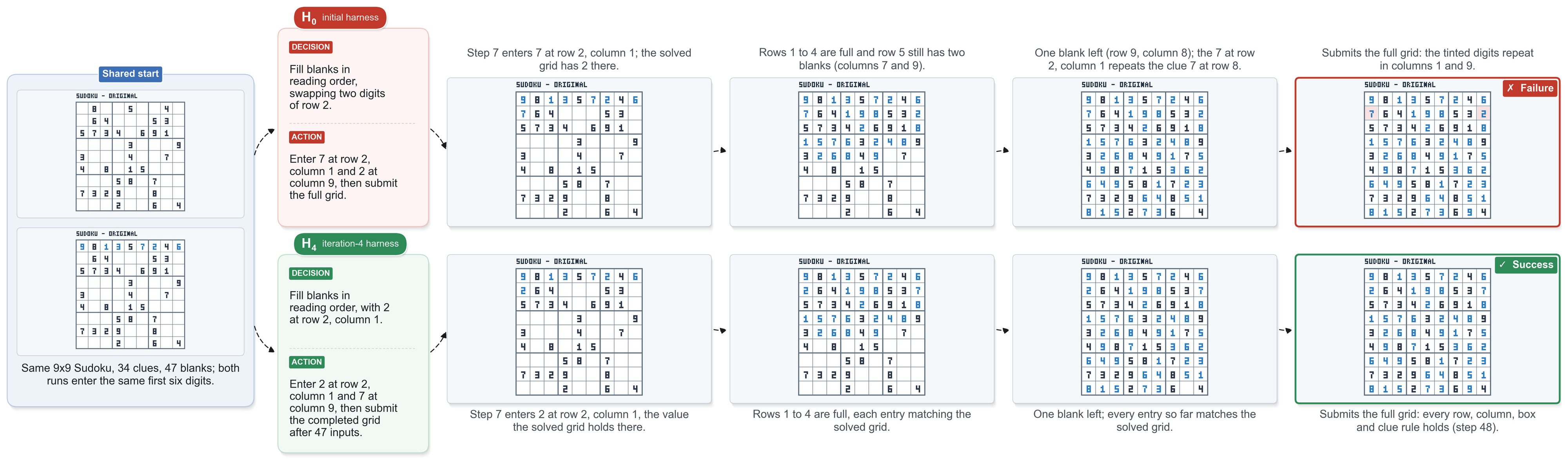}
\captionof{figure}{\textbf{puzz.link / Penpa+ (train)}}
\end{minipage}\par\vspace{4pt}
\noindent\begin{minipage}{\linewidth}\centering
\includegraphics[width=\linewidth]{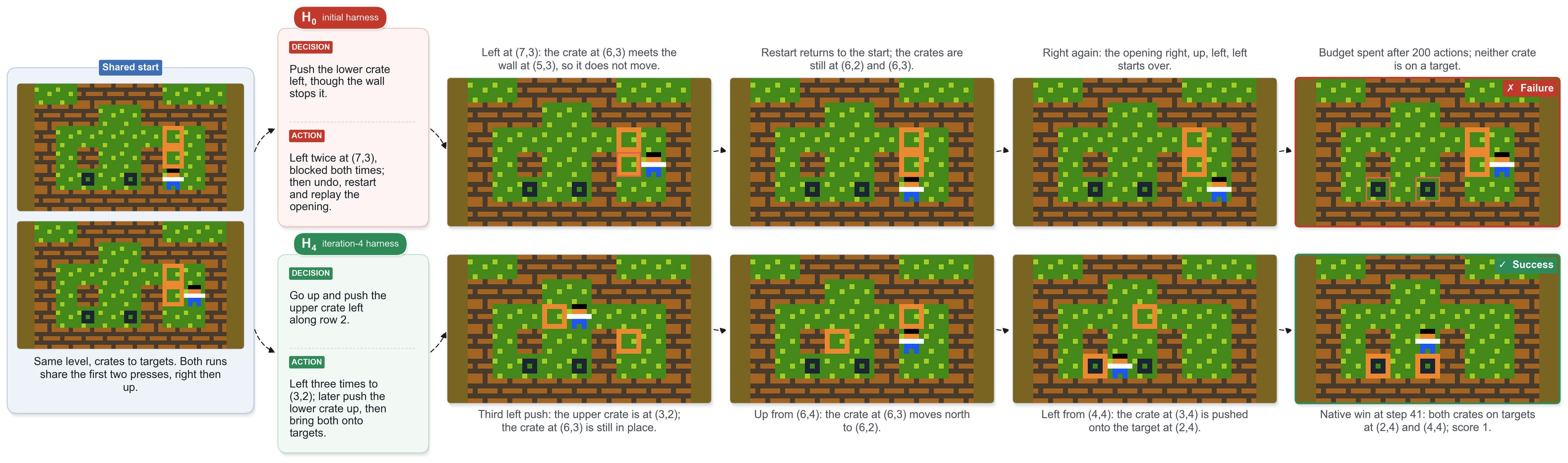}
\captionof{figure}{\textbf{PuzzleScript (train)}}
\end{minipage}\par\vspace{4pt}
\noindent\begin{minipage}{\linewidth}\centering
\includegraphics[width=\linewidth]{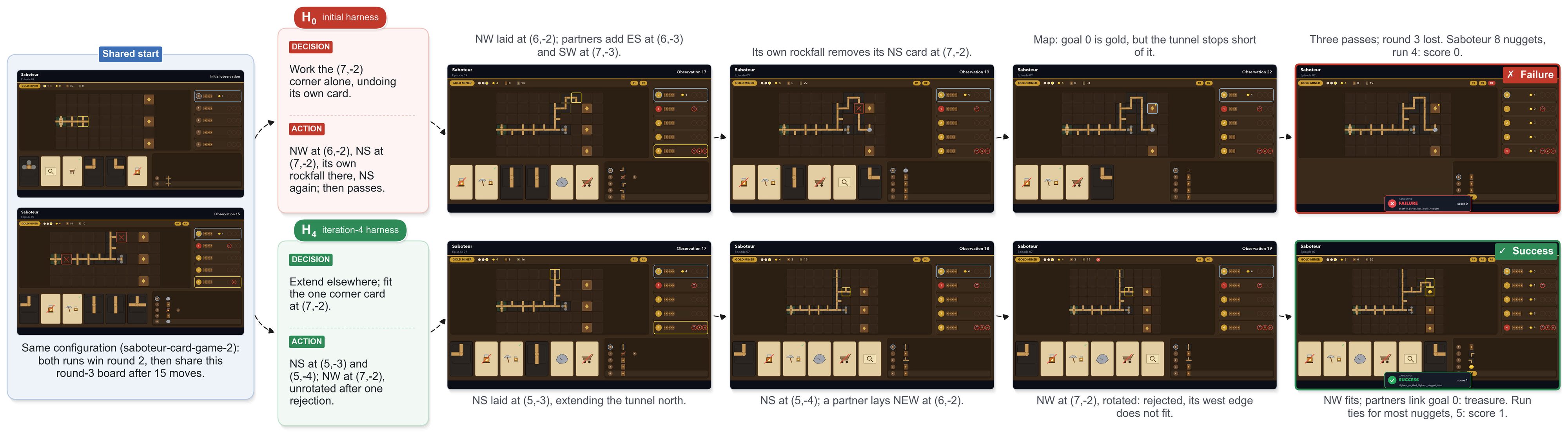}
\captionof{figure}{\textbf{Saboteur (train)}}
\end{minipage}\par\vspace{4pt}
\noindent\begin{minipage}{\linewidth}\centering
\includegraphics[width=\linewidth]{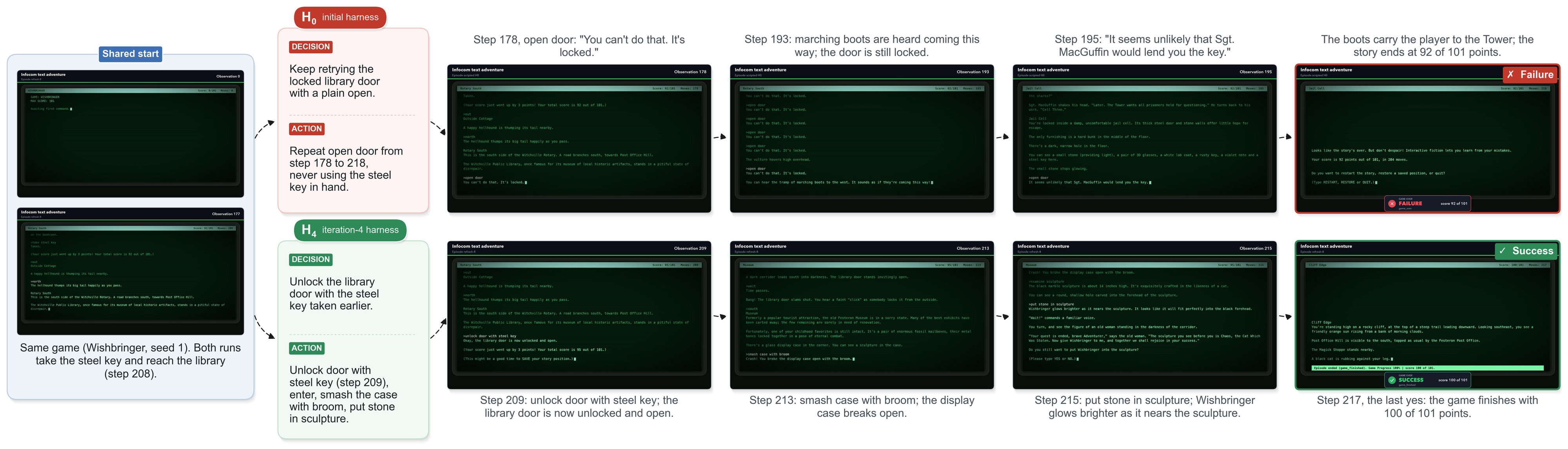}
\captionof{figure}{\textbf{TextQuests (train)}}
\end{minipage}\par\vspace{4pt}
\noindent\begin{minipage}{\linewidth}\centering
\includegraphics[width=\linewidth]{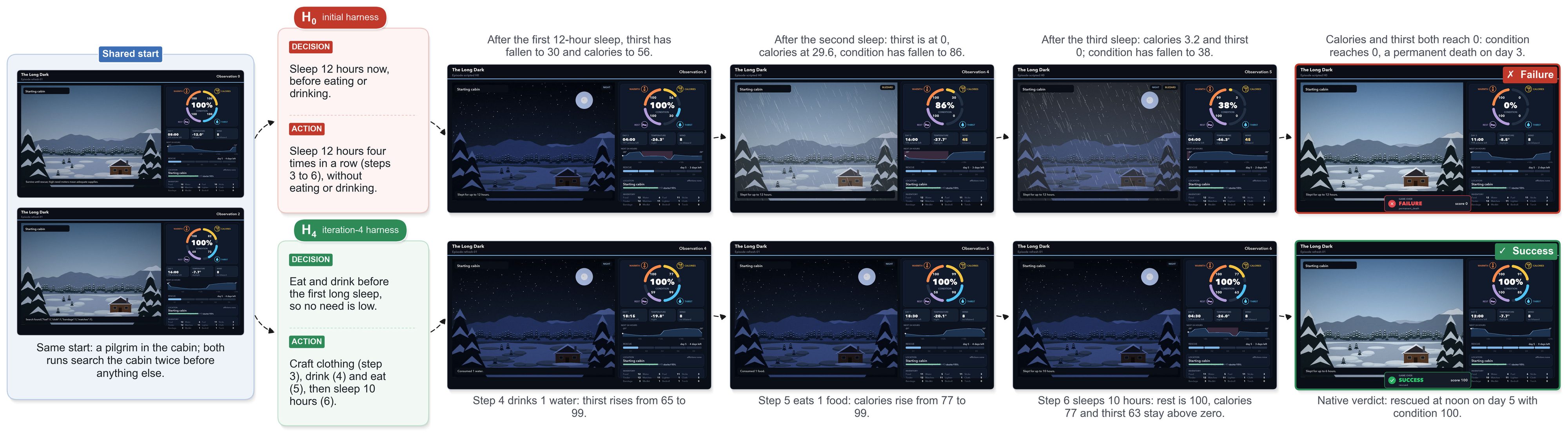}
\captionof{figure}{\textbf{The Long Dark (train)}}
\end{minipage}\par\vspace{4pt}
\noindent\begin{minipage}{\linewidth}\centering
\includegraphics[width=\linewidth]{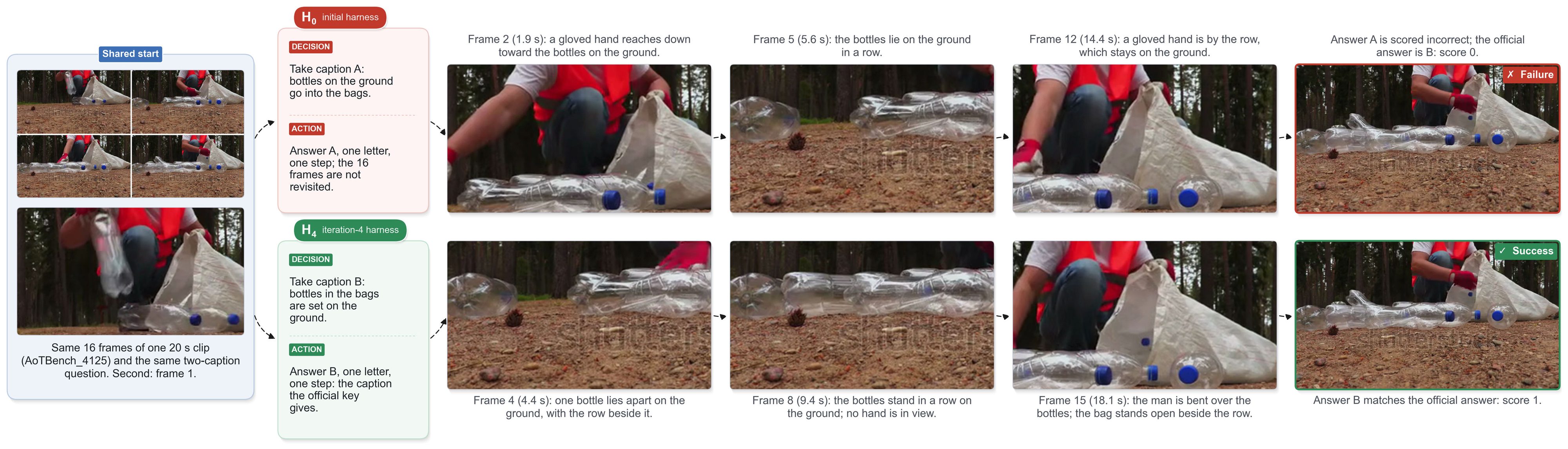}
\captionof{figure}{\textbf{Video-Index (train)}}
\end{minipage}\par\vspace{4pt}
\noindent\begin{minipage}{\linewidth}\centering
\includegraphics[width=\linewidth]{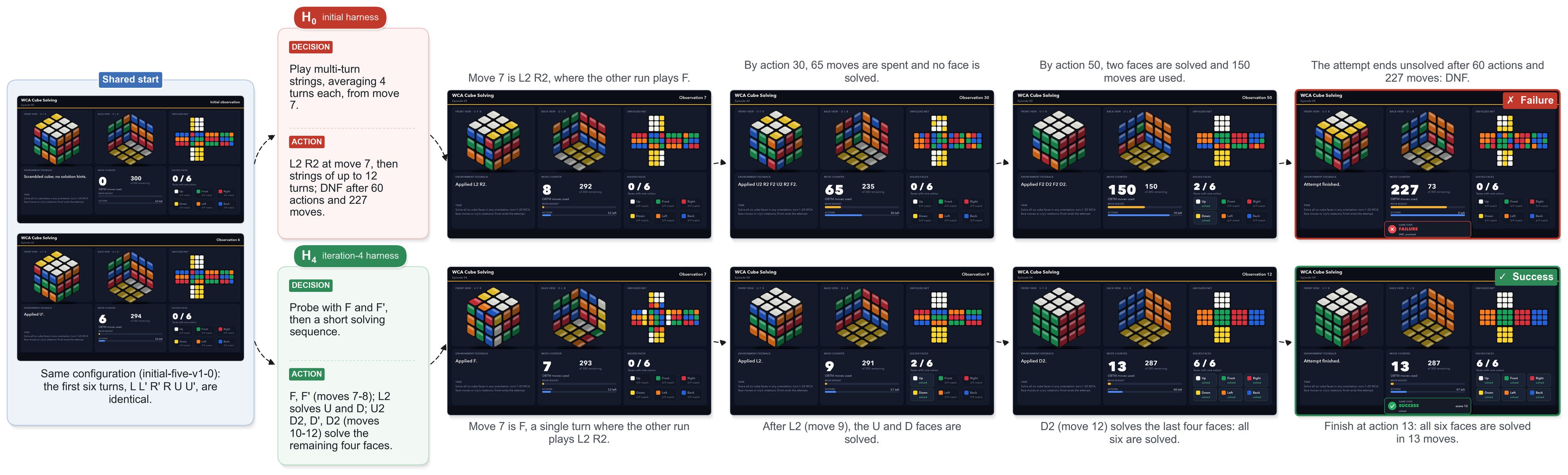}
\captionof{figure}{\textbf{WCA Cube Solving (train)}}
\end{minipage}\par\vspace{4pt}
\noindent\begin{minipage}{\linewidth}\centering
\includegraphics[width=\linewidth]{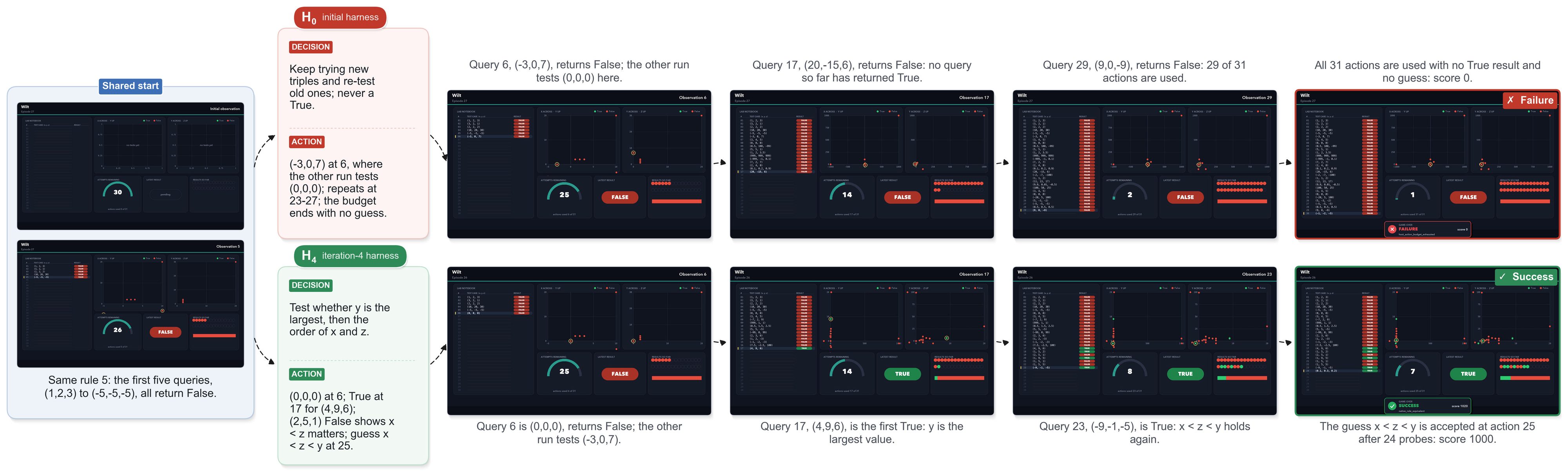}
\captionof{figure}{\textbf{WILT (train)}}
\end{minipage}\par\vspace{4pt}
\noindent\begin{minipage}{\linewidth}\centering
\includegraphics[width=\linewidth]{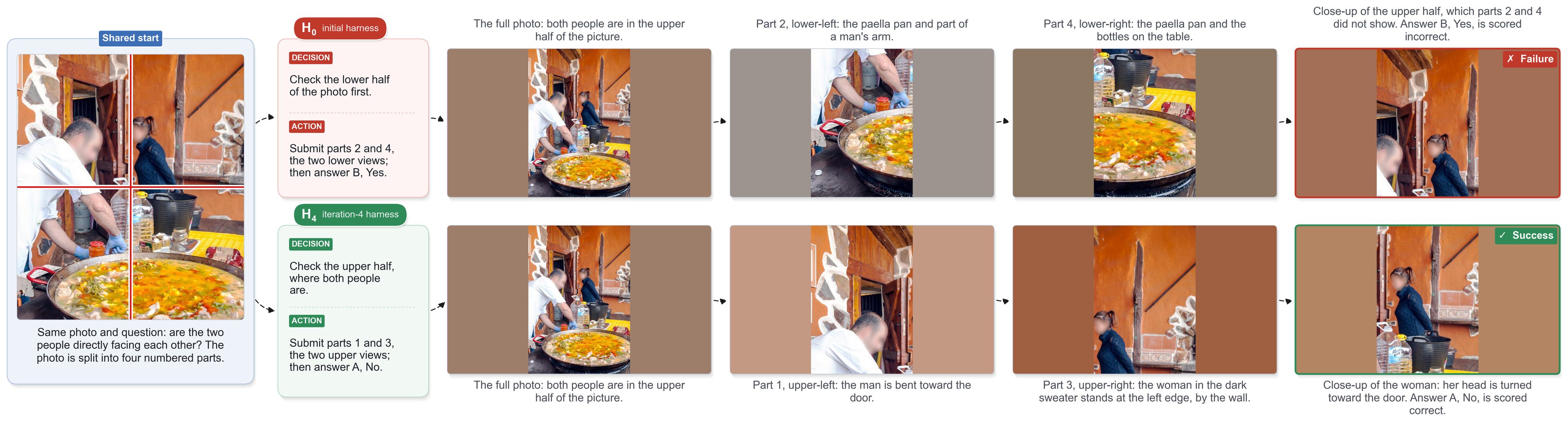}
\captionof{figure}{\textbf{ActiView (unseen)}}
\end{minipage}\par\vspace{4pt}
\noindent\begin{minipage}{\linewidth}\centering
\includegraphics[width=\linewidth]{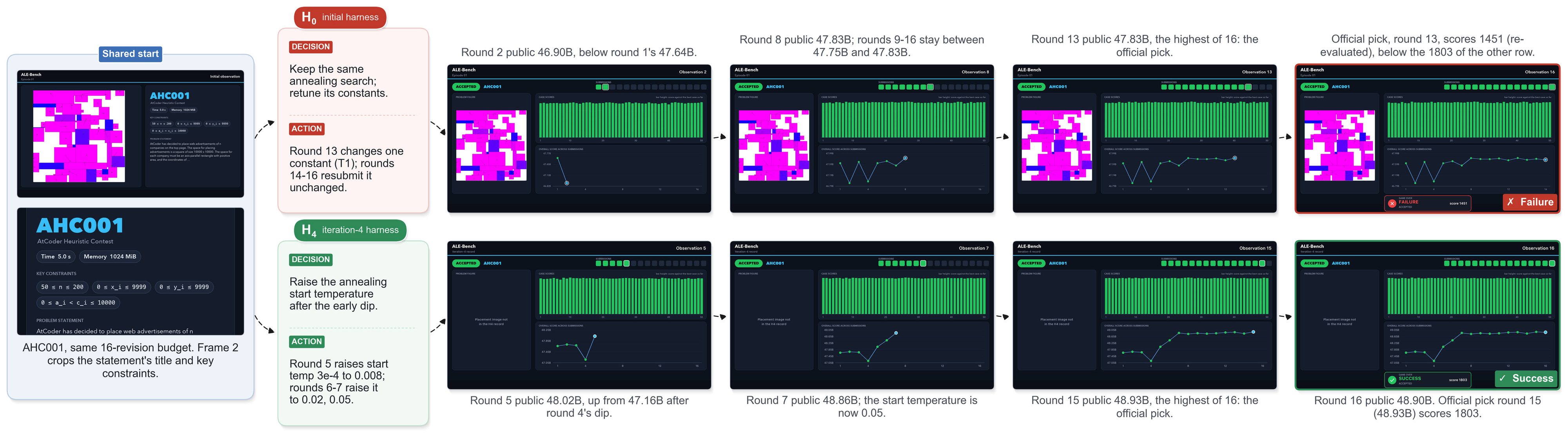}
\captionof{figure}{\textbf{ALE-Bench (unseen)}}
\end{minipage}\par\vspace{4pt}
\noindent\begin{minipage}{\linewidth}\centering
\includegraphics[width=\linewidth]{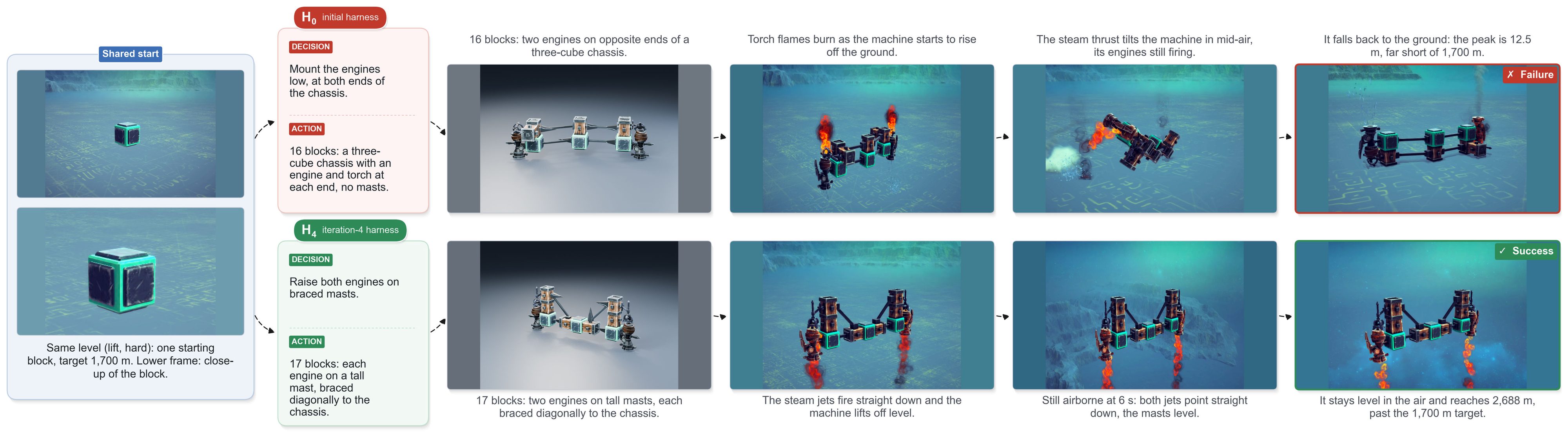}
\captionof{figure}{\textbf{BuildArena (unseen)}}
\end{minipage}\par\vspace{4pt}
\noindent\begin{minipage}{\linewidth}\centering
\includegraphics[width=\linewidth]{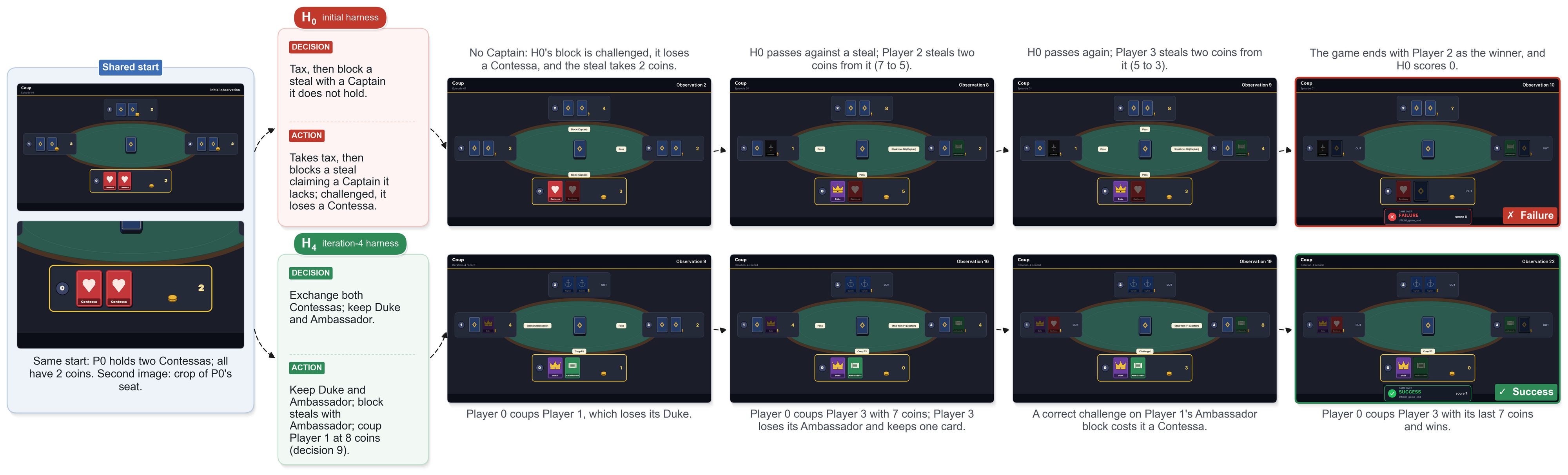}
\captionof{figure}{\textbf{Coup (unseen)}}
\end{minipage}\par\vspace{4pt}
\noindent\begin{minipage}{\linewidth}\centering
\includegraphics[width=\linewidth]{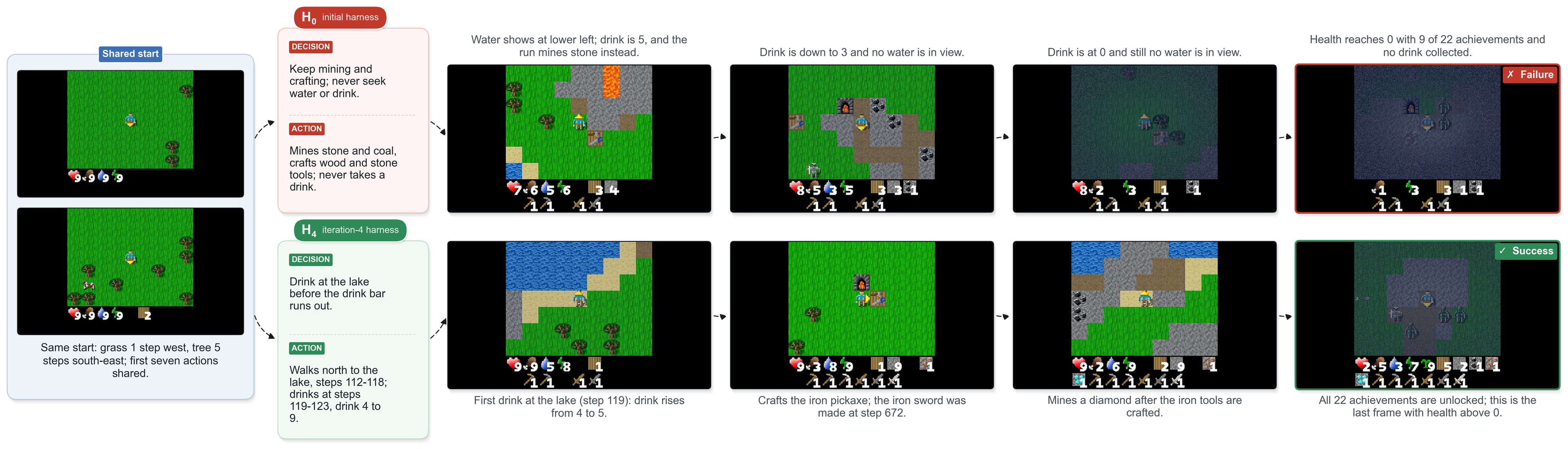}
\captionof{figure}{\textbf{Crafter (unseen)}}
\end{minipage}\par\vspace{4pt}
\noindent\begin{minipage}{\linewidth}\centering
\includegraphics[width=\linewidth]{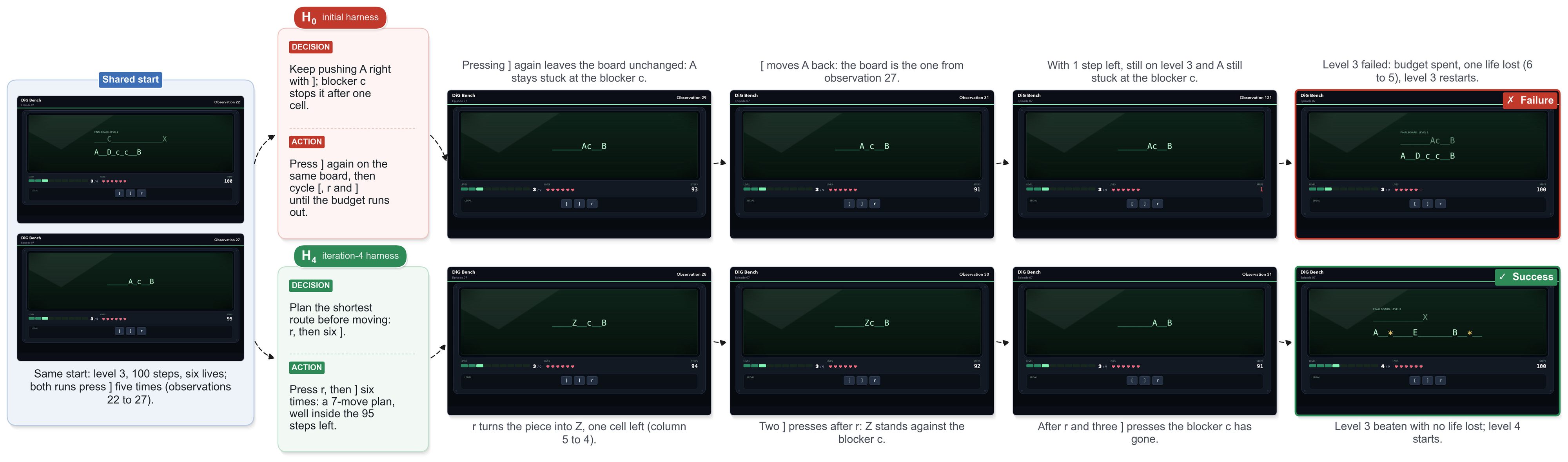}
\captionof{figure}{\textbf{DiG-bench (unseen)}}
\end{minipage}\par\vspace{4pt}
\noindent\begin{minipage}{\linewidth}\centering
\includegraphics[width=\linewidth]{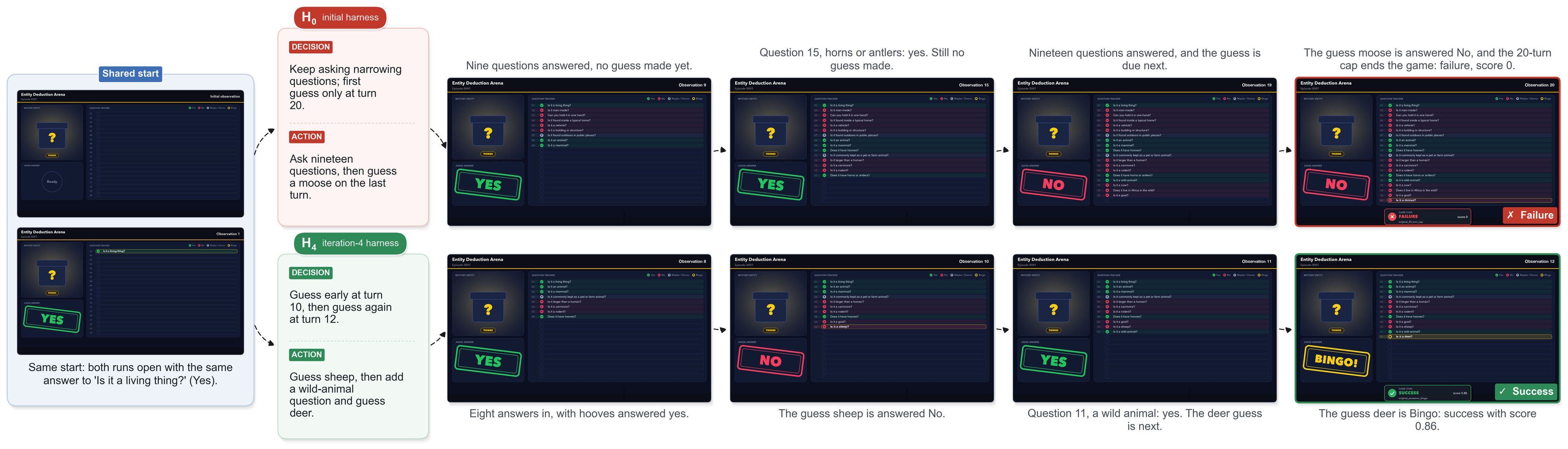}
\captionof{figure}{\textbf{Entity Deduction Arena (unseen)}}
\end{minipage}\par\vspace{4pt}
\noindent\begin{minipage}{\linewidth}\centering
\includegraphics[width=\linewidth]{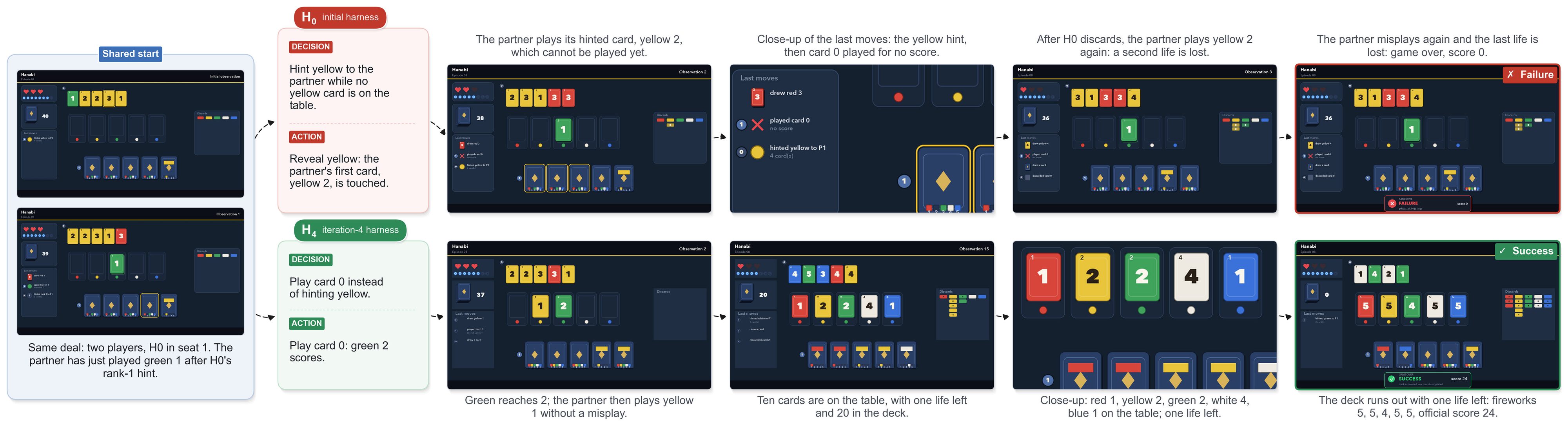}
\captionof{figure}{\textbf{Hanabi (unseen)}}
\end{minipage}\par\vspace{4pt}
\noindent\begin{minipage}{\linewidth}\centering
\includegraphics[width=\linewidth]{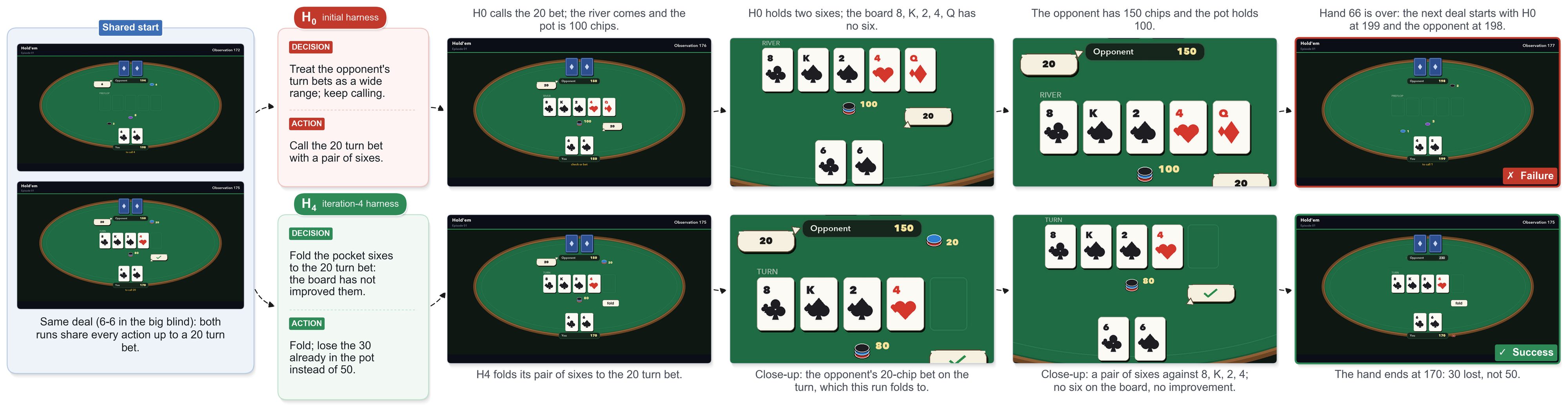}
\captionof{figure}{\textbf{Heads-Up NL Hold'em (unseen)}}
\end{minipage}\par\vspace{4pt}
\noindent\begin{minipage}{\linewidth}\centering
\includegraphics[width=\linewidth]{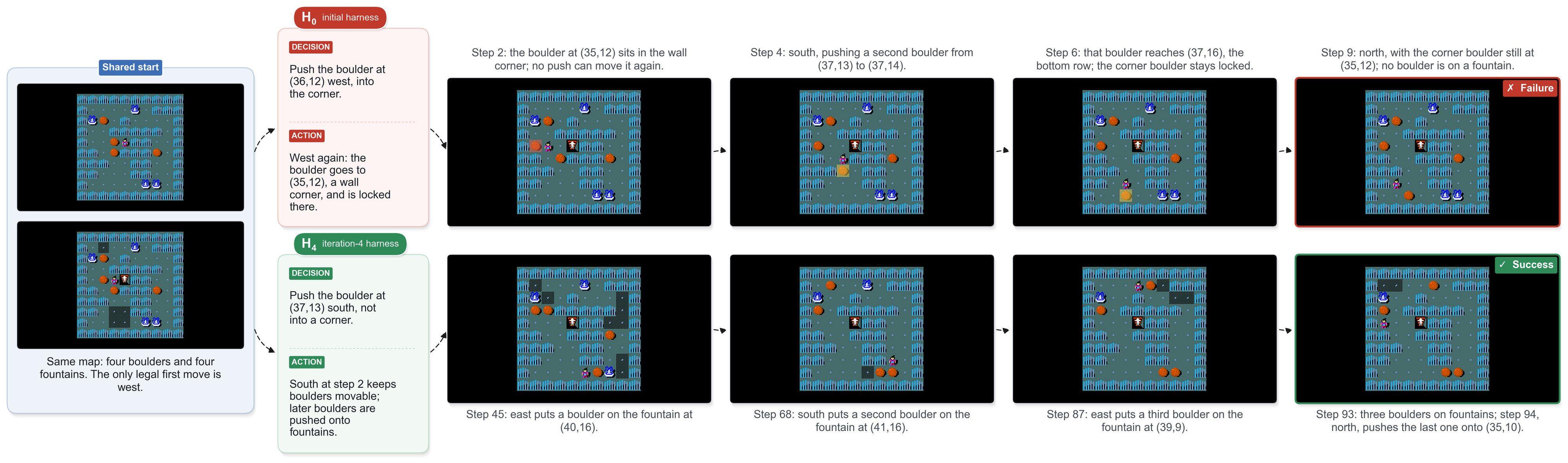}
\captionof{figure}{\textbf{MiniHack Boxoban (unseen)}}
\end{minipage}\par\vspace{4pt}
\noindent\begin{minipage}{\linewidth}\centering
\includegraphics[width=\linewidth]{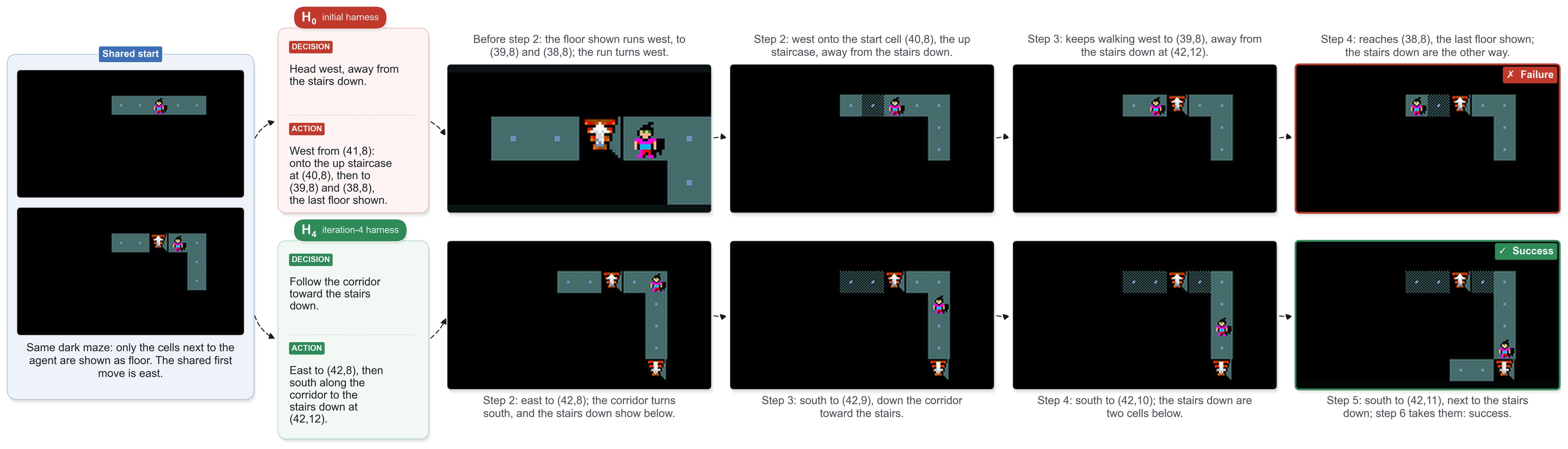}
\captionof{figure}{\textbf{MiniHack Maze / Corridor (unseen)}}
\end{minipage}\par\vspace{4pt}
\noindent\begin{minipage}{\linewidth}\centering
\includegraphics[width=\linewidth]{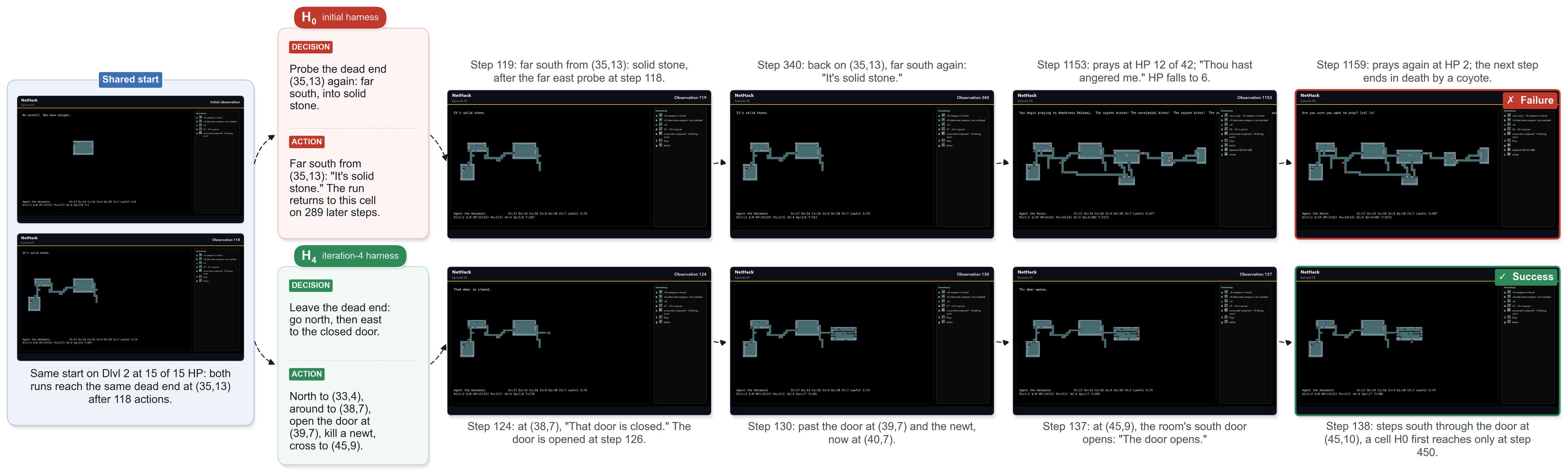}
\captionof{figure}{\textbf{NetHack (unseen)}}
\end{minipage}\par\vspace{4pt}
\noindent\begin{minipage}{\linewidth}\centering
\includegraphics[width=\linewidth]{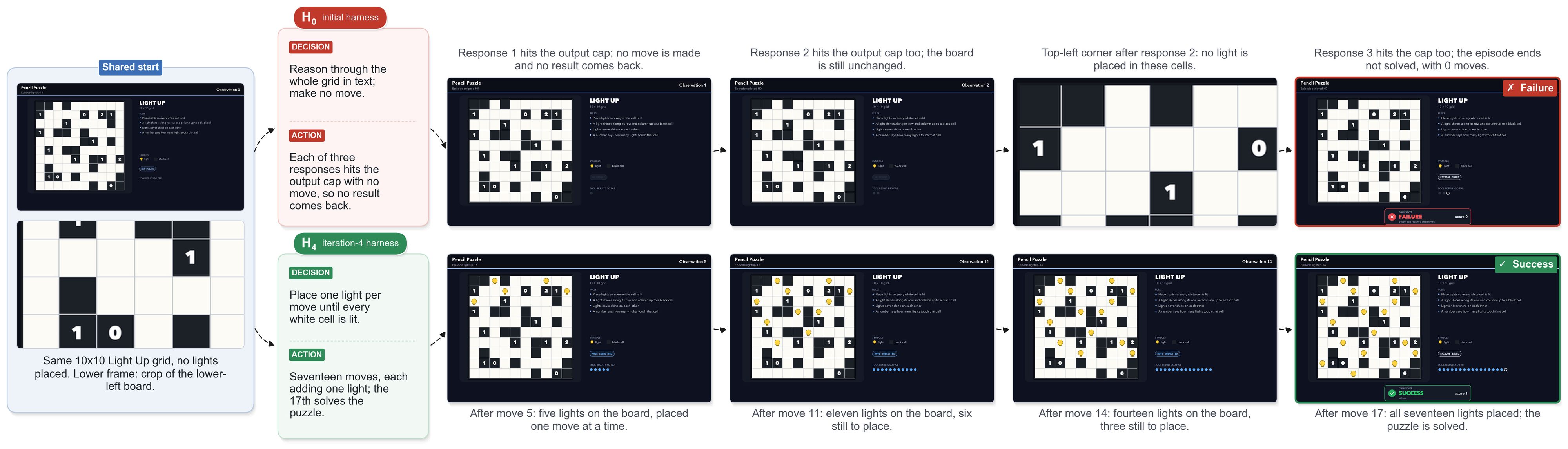}
\captionof{figure}{\textbf{Pencil Puzzle Bench (unseen)}}
\end{minipage}\par\vspace{4pt}
\noindent\begin{minipage}{\linewidth}\centering
\includegraphics[width=\linewidth]{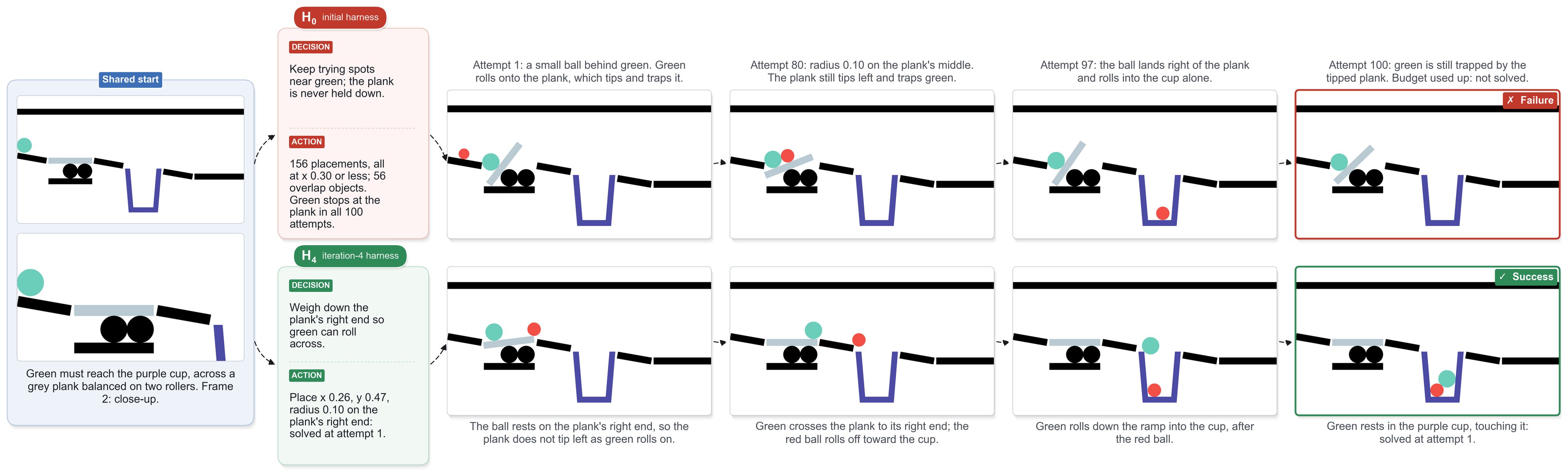}
\captionof{figure}{\textbf{PHYRE (unseen)}}
\end{minipage}\par\vspace{4pt}
\noindent\begin{minipage}{\linewidth}\centering
\includegraphics[width=\linewidth]{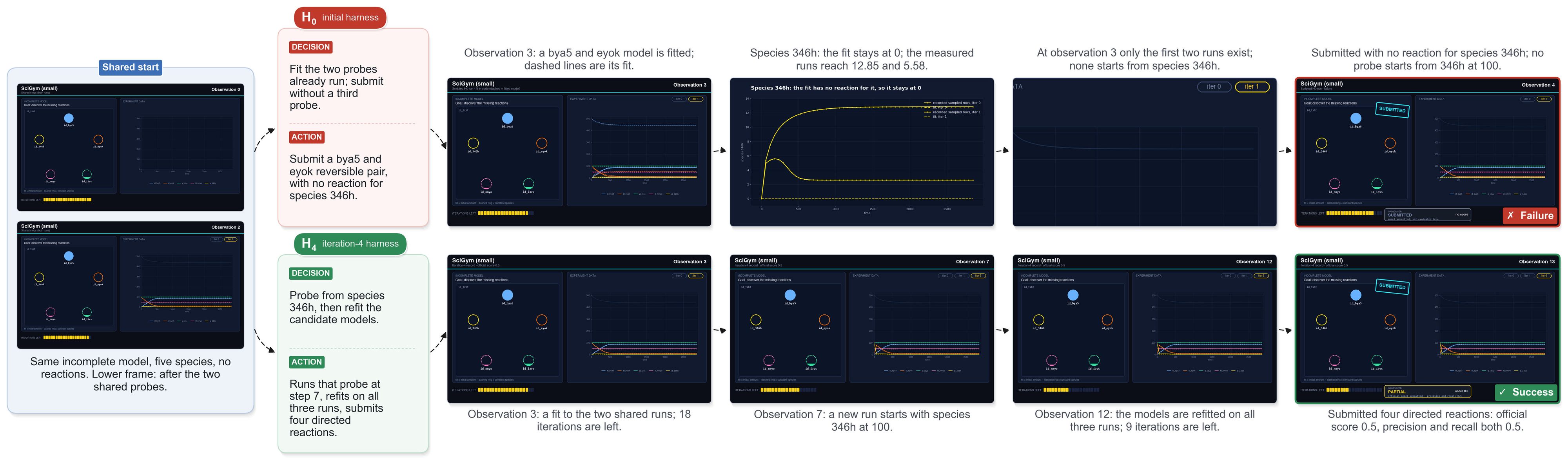}
\captionof{figure}{\textbf{SciGym (unseen)}}
\end{minipage}\par\vspace{4pt}
\noindent\begin{minipage}{\linewidth}\centering
\includegraphics[width=\linewidth]{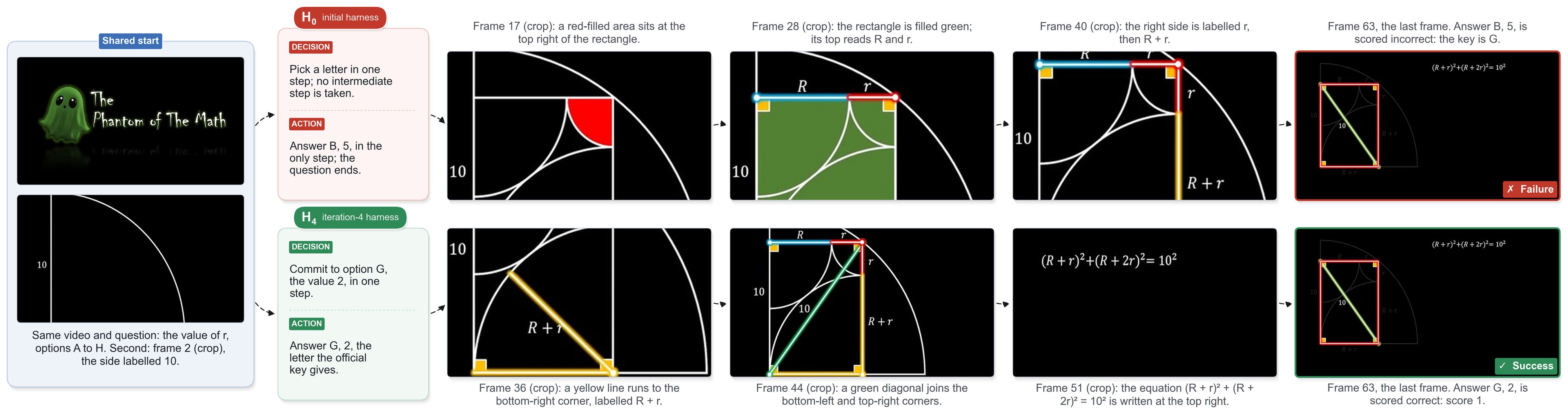}
\captionof{figure}{\textbf{Video-MME-v2 (unseen)}}
\end{minipage}\par\vspace{4pt}
\noindent\begin{minipage}{\linewidth}\centering
\includegraphics[width=\linewidth]{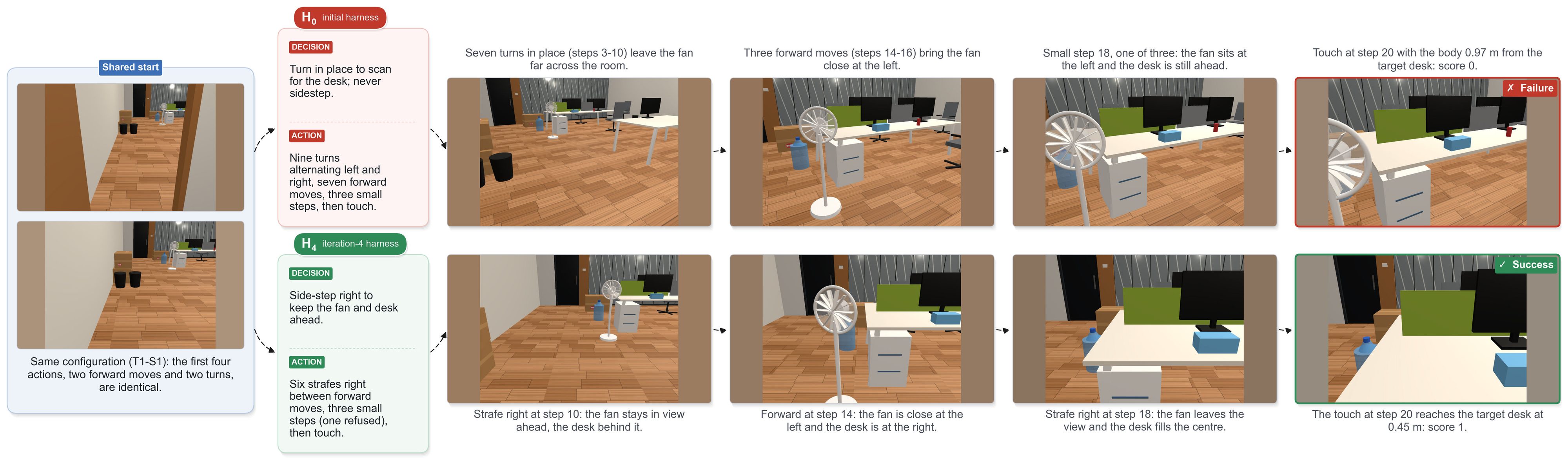}
\captionof{figure}{\textbf{SOON (embodied, simulation)}}
\end{minipage}\par\vspace{4pt}
\noindent\begin{minipage}{\linewidth}\centering
\includegraphics[width=\linewidth]{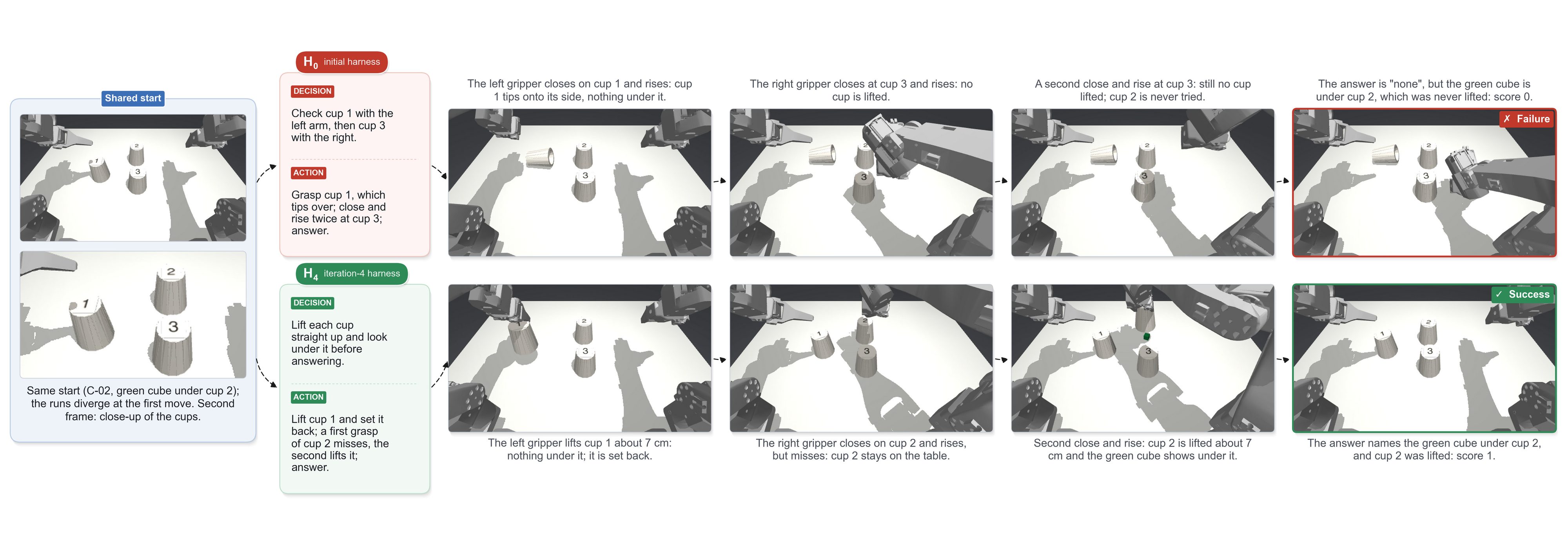}
\captionof{figure}{\textbf{CaP-X (embodied, simulation)}}\label{fig:pair-last}
\end{minipage}\par\vspace{4pt}
}

\FloatBarrier
\subsection{Node Training Levels}
\label{app:levels}
At every capability node, Evolver designs short training levels from the node's requirement, and Creator implements and revises them under the Creator Harness of Appendix~\ref{app:creator}; each level is a simulation world with a 3D scene, public rules, a fixed action budget, and a scorer, and one level per node and iteration gives 44 levels. Because e and e$\to$f form a separate path for video understanding, Video-Index, the only training benchmark with video input, feeds both nodes. Figures~\ref{fig:node-first}--\ref{fig:node-last} show how each node evolves: on the left are the training benchmarks that form the node's input, and on the right its four levels are played in turn, each tracing how the failure on that level is distilled into a new or revised module and how the next run with that module completes the level. The levels of a node keep one task family and raise the difficulty from iteration to iteration. At c, two robots deliver crates through gates that open only while the partner stands on a linked plate (\begingroup\leavevmode\setlength{\arxiviconheight}{1.25em}\raisebox{-0.20\arxiviconheight}{\includegraphics[height=\arxiviconheight]{appendix/figures/harness-icons/pdf-numbered/M07.pdf}}\endgroup\,M07, \begingroup\leavevmode\setlength{\arxiviconheight}{1.25em}\raisebox{-0.20\arxiviconheight}{\includegraphics[height=\arxiviconheight]{appendix/figures/harness-icons/pdf-numbered/M08.pdf}}\endgroup\,M08). At c$\to$d, a factory must meet orders due on fixed shifts under staged shortages (\begingroup\leavevmode\setlength{\arxiviconheight}{1.25em}\raisebox{-0.20\arxiviconheight}{\includegraphics[height=\arxiviconheight]{appendix/figures/harness-icons/pdf-numbered/M12.pdf}}\endgroup\,M12, \begingroup\leavevmode\setlength{\arxiviconheight}{1.25em}\raisebox{-0.20\arxiviconheight}{\includegraphics[height=\arxiviconheight]{appendix/figures/harness-icons/pdf-numbered/M13.pdf}}\endgroup\,M13, \begingroup\leavevmode\setlength{\arxiviconheight}{1.25em}\raisebox{-0.20\arxiviconheight}{\includegraphics[height=\arxiviconheight]{appendix/figures/harness-icons/pdf-numbered/M14.pdf}}\endgroup\,M14). At c$\to$d$\to$a, ring-marked cells must satisfy relations set in stages (\begingroup\leavevmode\setlength{\arxiviconheight}{1.25em}\raisebox{-0.20\arxiviconheight}{\includegraphics[height=\arxiviconheight]{appendix/figures/harness-icons/pdf-numbered/M15.pdf}}\endgroup\,M15, \begingroup\leavevmode\setlength{\arxiviconheight}{1.25em}\raisebox{-0.20\arxiviconheight}{\includegraphics[height=\arxiviconheight]{appendix/figures/harness-icons/pdf-numbered/M16.pdf}}\endgroup\,M16). At c$\to$a, similar jars must be collected in a revealed order (\begingroup\leavevmode\setlength{\arxiviconheight}{1.25em}\raisebox{-0.20\arxiviconheight}{\includegraphics[height=\arxiviconheight]{appendix/figures/harness-icons/pdf-numbered/M10.pdf}}\endgroup\,M10, \begingroup\leavevmode\setlength{\arxiviconheight}{1.25em}\raisebox{-0.20\arxiviconheight}{\includegraphics[height=\arxiviconheight]{appendix/figures/harness-icons/pdf-numbered/M09.pdf}}\endgroup\,M09, \begingroup\leavevmode\setlength{\arxiviconheight}{1.25em}\raisebox{-0.20\arxiviconheight}{\includegraphics[height=\arxiviconheight]{appendix/figures/harness-icons/pdf-numbered/M11.pdf}}\endgroup\,M11). At d, a claim may be submitted only once an object's hidden faces have been seen (\begingroup\leavevmode\setlength{\arxiviconheight}{1.25em}\raisebox{-0.20\arxiviconheight}{\includegraphics[height=\arxiviconheight]{appendix/figures/harness-icons/pdf-numbered/M17.pdf}}\endgroup\,M17, \begingroup\leavevmode\setlength{\arxiviconheight}{1.25em}\raisebox{-0.20\arxiviconheight}{\includegraphics[height=\arxiviconheight]{appendix/figures/harness-icons/pdf-numbered/M18.pdf}}\endgroup\,M18). At d$\to$a, stamps and a parcel must be brought back while gates close on the return (\begingroup\leavevmode\setlength{\arxiviconheight}{1.25em}\raisebox{-0.20\arxiviconheight}{\includegraphics[height=\arxiviconheight]{appendix/figures/harness-icons/pdf-numbered/M19.pdf}}\endgroup\,M19, \begingroup\leavevmode\setlength{\arxiviconheight}{1.25em}\raisebox{-0.20\arxiviconheight}{\includegraphics[height=\arxiviconheight]{appendix/figures/harness-icons/pdf-numbered/M20.pdf}}\endgroup\,M20, \begingroup\leavevmode\setlength{\arxiviconheight}{1.25em}\raisebox{-0.20\arxiviconheight}{\includegraphics[height=\arxiviconheight]{appendix/figures/harness-icons/pdf-numbered/M21.pdf}}\endgroup\,M21). At d$\to$b, a hidden pattern with repeated symbols must be identified by probes (\begingroup\leavevmode\setlength{\arxiviconheight}{1.25em}\raisebox{-0.20\arxiviconheight}{\includegraphics[height=\arxiviconheight]{appendix/figures/harness-icons/pdf-numbered/M22.pdf}}\endgroup\,M22, \begingroup\leavevmode\setlength{\arxiviconheight}{1.25em}\raisebox{-0.20\arxiviconheight}{\includegraphics[height=\arxiviconheight]{appendix/figures/harness-icons/pdf-numbered/M23.pdf}}\endgroup\,M23). At a, structural members are inspected for defects beside look-alikes (\begingroup\leavevmode\setlength{\arxiviconheight}{1.25em}\raisebox{-0.20\arxiviconheight}{\includegraphics[height=\arxiviconheight]{appendix/figures/harness-icons/pdf-numbered/M01.pdf}}\endgroup\,M01, \begingroup\leavevmode\setlength{\arxiviconheight}{1.25em}\raisebox{-0.20\arxiviconheight}{\includegraphics[height=\arxiviconheight]{appendix/figures/harness-icons/pdf-numbered/M02.pdf}}\endgroup\,M02, \begingroup\leavevmode\setlength{\arxiviconheight}{1.25em}\raisebox{-0.20\arxiviconheight}{\includegraphics[height=\arxiviconheight]{appendix/figures/harness-icons/pdf-numbered/M03.pdf}}\endgroup\,M03). At b, tokens are allocated over three fields against a rotating formation (\begingroup\leavevmode\setlength{\arxiviconheight}{1.25em}\raisebox{-0.20\arxiviconheight}{\includegraphics[height=\arxiviconheight]{appendix/figures/harness-icons/pdf-numbered/M06.pdf}}\endgroup\,M06, \begingroup\leavevmode\setlength{\arxiviconheight}{1.25em}\raisebox{-0.20\arxiviconheight}{\includegraphics[height=\arxiviconheight]{appendix/figures/harness-icons/pdf-numbered/M04.pdf}}\endgroup\,M04, \begingroup\leavevmode\setlength{\arxiviconheight}{1.25em}\raisebox{-0.20\arxiviconheight}{\includegraphics[height=\arxiviconheight]{appendix/figures/harness-icons/pdf-numbered/M05.pdf}}\endgroup\,M05). At e, the segments holding a parcel's hand-offs must be bought within a budget (\begingroup\leavevmode\setlength{\arxiviconheight}{1.25em}\raisebox{-0.20\arxiviconheight}{\includegraphics[height=\arxiviconheight]{appendix/figures/harness-icons/pdf-numbered/M25.pdf}}\endgroup\,M25, \begingroup\leavevmode\setlength{\arxiviconheight}{1.25em}\raisebox{-0.20\arxiviconheight}{\includegraphics[height=\arxiviconheight]{appendix/figures/harness-icons/pdf-numbered/M24.pdf}}\endgroup\,M24, \begingroup\leavevmode\setlength{\arxiviconheight}{1.25em}\raisebox{-0.20\arxiviconheight}{\includegraphics[height=\arxiviconheight]{appendix/figures/harness-icons/pdf-numbered/M26.pdf}}\endgroup\,M26). At e$\to$f, the parcel must be followed across hand-offs to its endpoint (\begingroup\leavevmode\setlength{\arxiviconheight}{1.25em}\raisebox{-0.20\arxiviconheight}{\includegraphics[height=\arxiviconheight]{appendix/figures/harness-icons/pdf-numbered/M28.pdf}}\endgroup\,M28, \begingroup\leavevmode\setlength{\arxiviconheight}{1.25em}\raisebox{-0.20\arxiviconheight}{\includegraphics[height=\arxiviconheight]{appendix/figures/harness-icons/pdf-numbered/M27.pdf}}\endgroup\,M27).

{\captionsetup{font=normalsize,skip=4pt}%   node figures flow inline, one per page, slightly wider than the text block
\noindent\begin{minipage}{\linewidth}\centering
\makebox[\linewidth][c]{\includegraphics[width=1.18\linewidth,height=0.9\textheight,keepaspectratio]{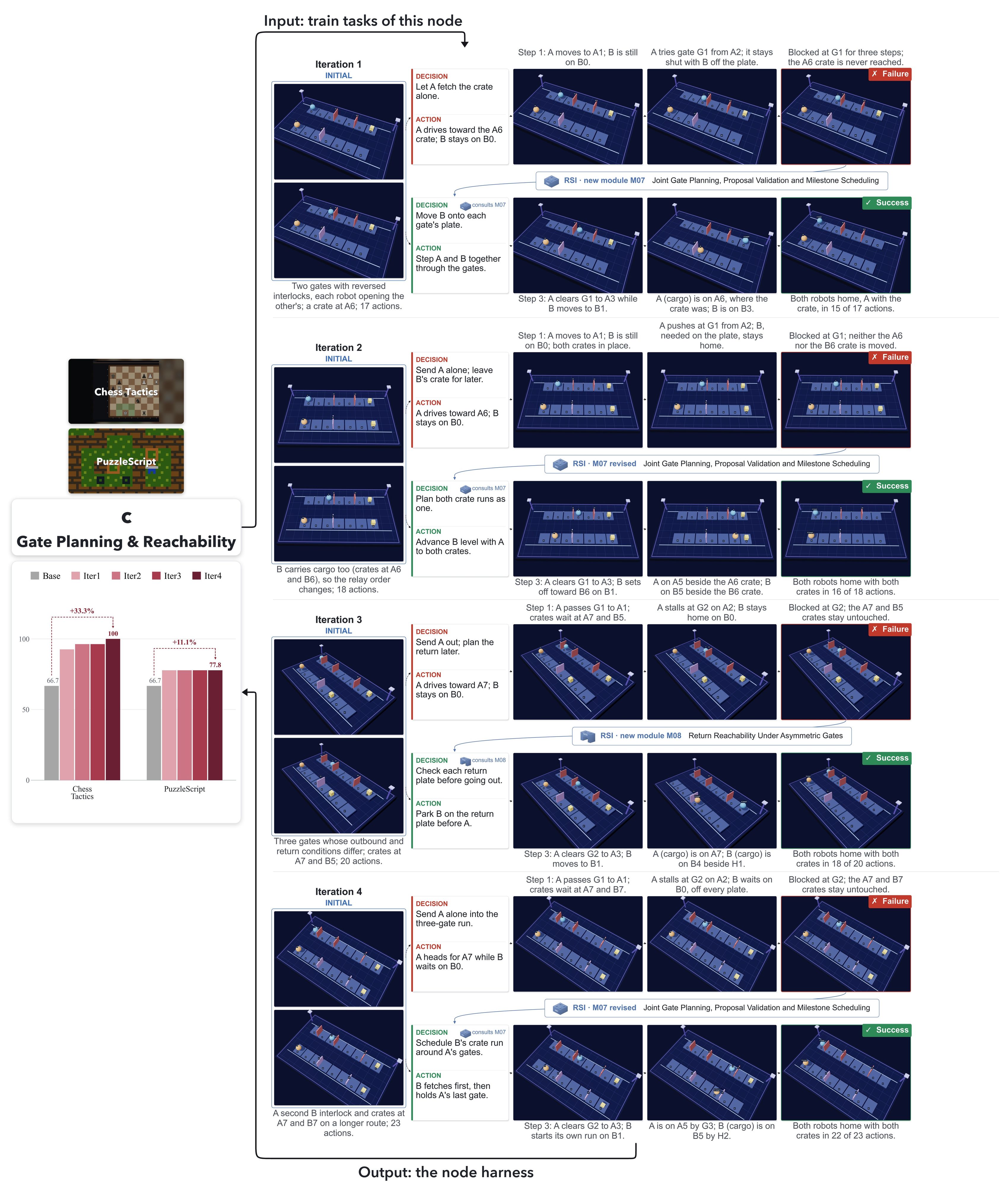}}
\captionof{figure}{\textbf{c: Gate Planning \& Reachability}}\label{fig:node-first}
\end{minipage}\par\vspace{6pt}
\noindent\begin{minipage}{\linewidth}\centering
\makebox[\linewidth][c]{\includegraphics[width=1.18\linewidth,height=0.9\textheight,keepaspectratio]{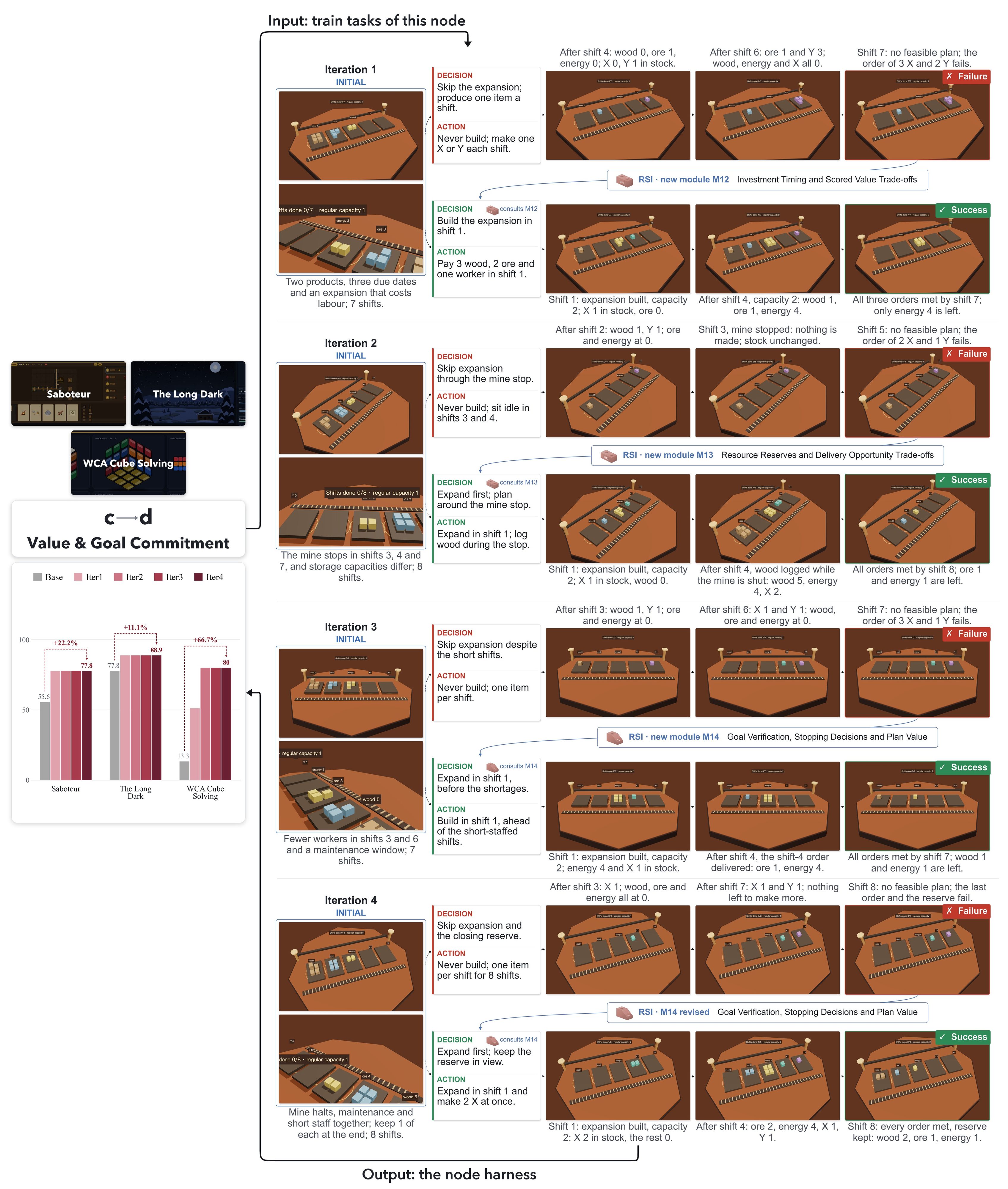}}
\captionof{figure}{\textbf{c$\to$d: Value \& Goal Commitment}}
\end{minipage}\par\vspace{6pt}
\noindent\begin{minipage}{\linewidth}\centering
\makebox[\linewidth][c]{\includegraphics[width=1.18\linewidth,height=0.9\textheight,keepaspectratio]{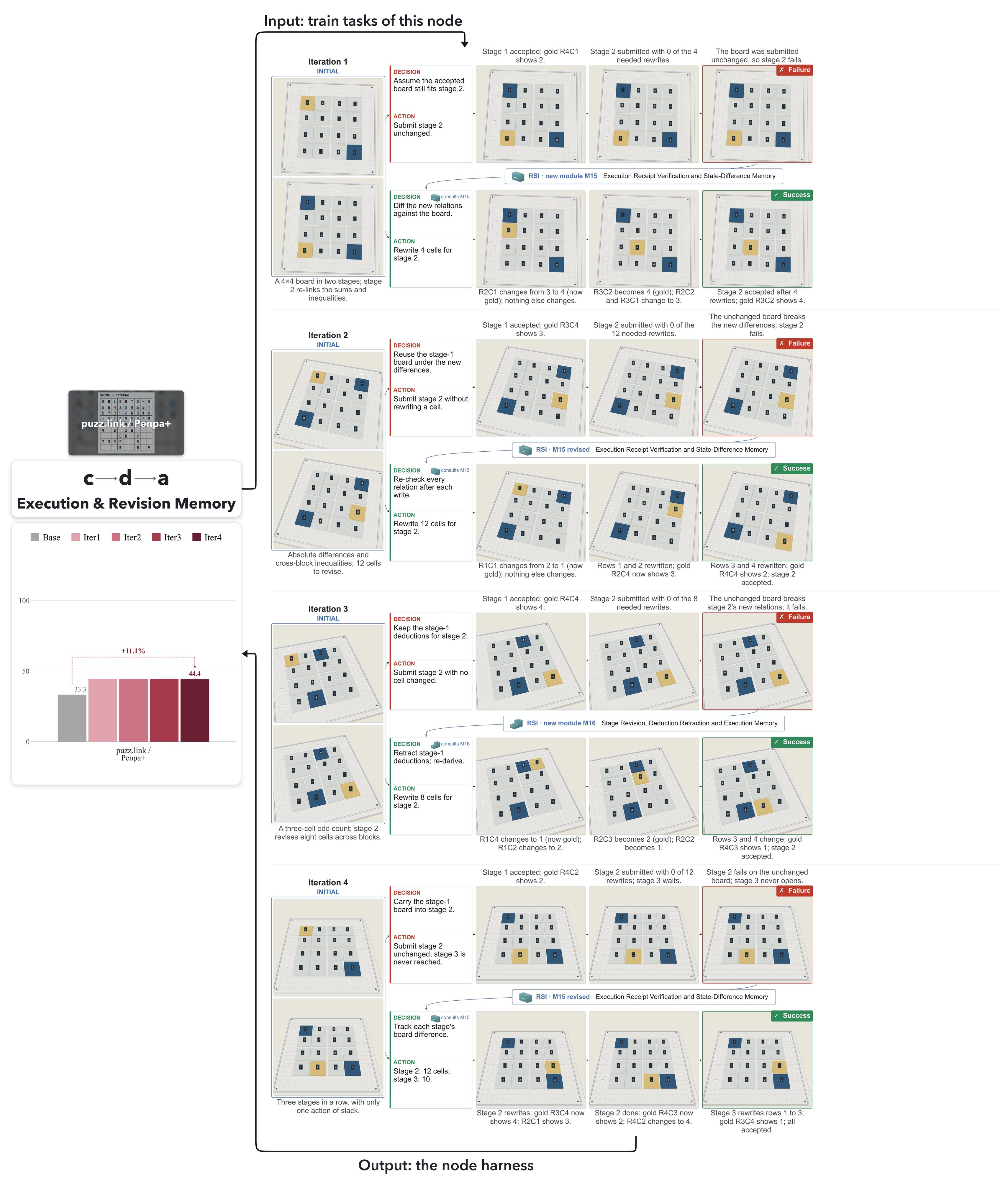}}
\captionof{figure}{\textbf{c$\to$d$\to$a: Execution \& Revision Memory}}
\end{minipage}\par\vspace{6pt}
\noindent\begin{minipage}{\linewidth}\centering
\makebox[\linewidth][c]{\includegraphics[width=1.18\linewidth,height=0.9\textheight,keepaspectratio]{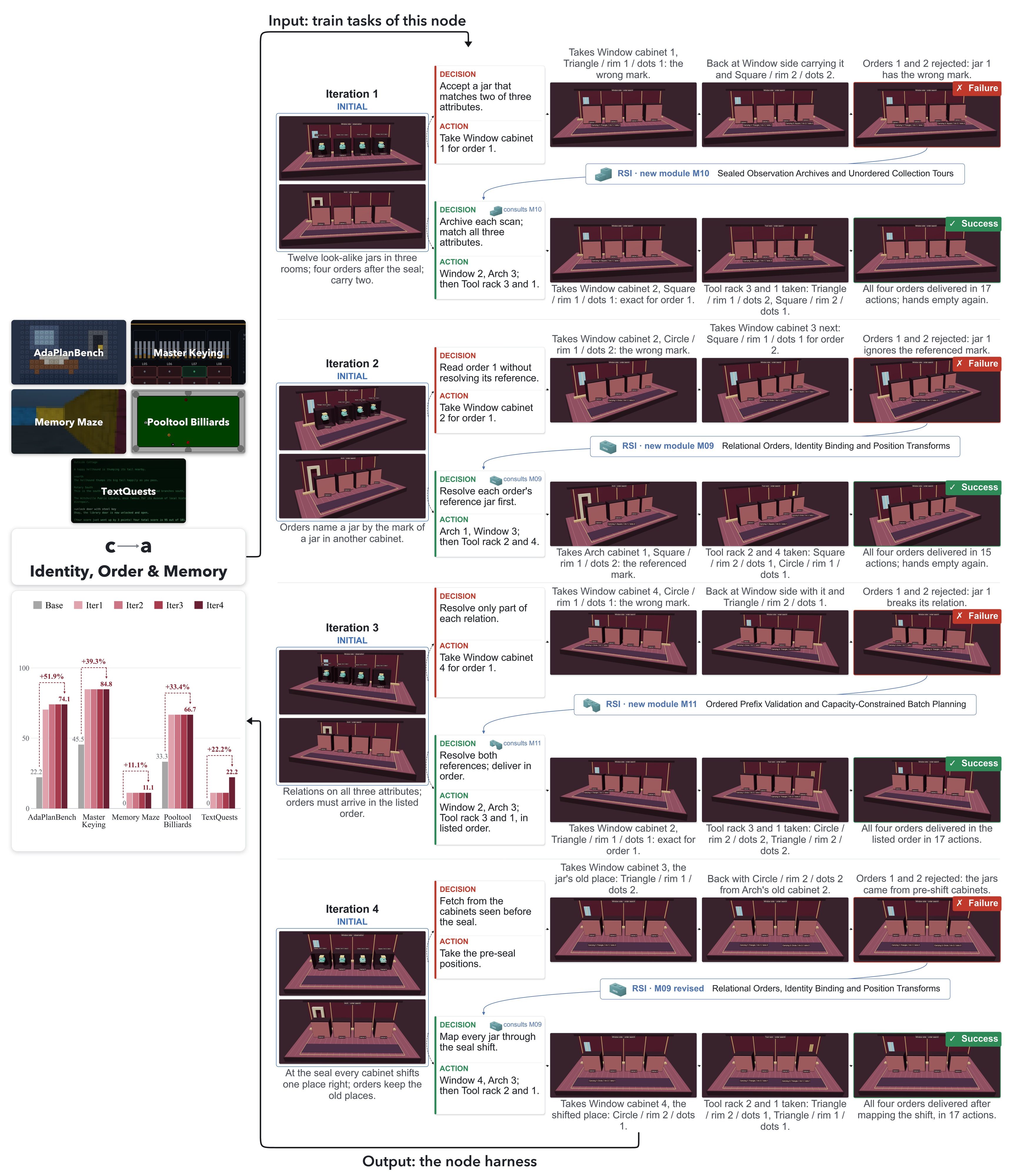}}
\captionof{figure}{\textbf{c$\to$a: Identity, Order \& Memory}}
\end{minipage}\par\vspace{6pt}
\noindent\begin{minipage}{\linewidth}\centering
\makebox[\linewidth][c]{\includegraphics[width=1.18\linewidth,height=0.9\textheight,keepaspectratio]{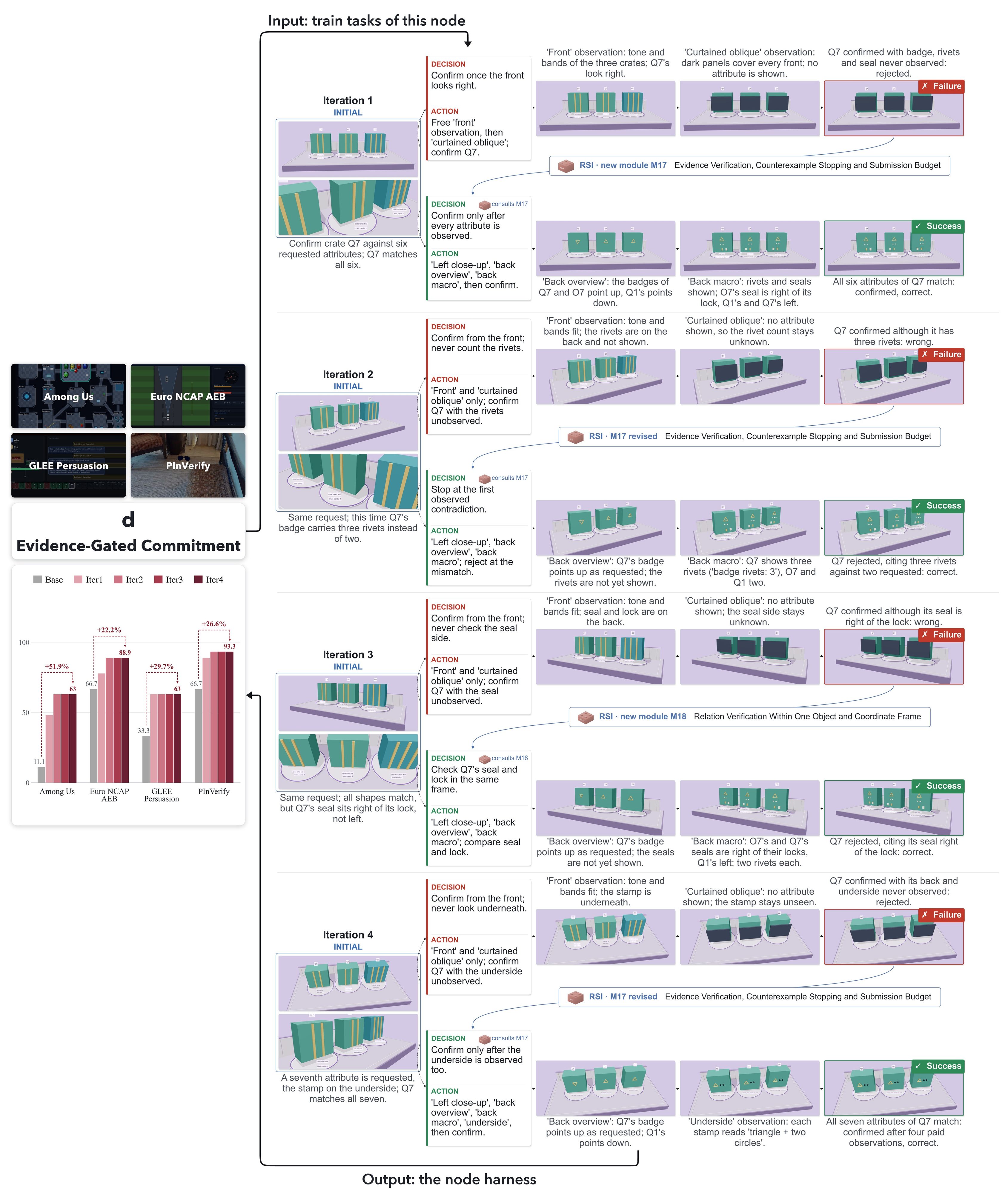}}
\captionof{figure}{\textbf{d: Evidence-Gated Commitment}}
\end{minipage}\par\vspace{6pt}
\noindent\begin{minipage}{\linewidth}\centering
\makebox[\linewidth][c]{\includegraphics[width=1.18\linewidth,height=0.9\textheight,keepaspectratio]{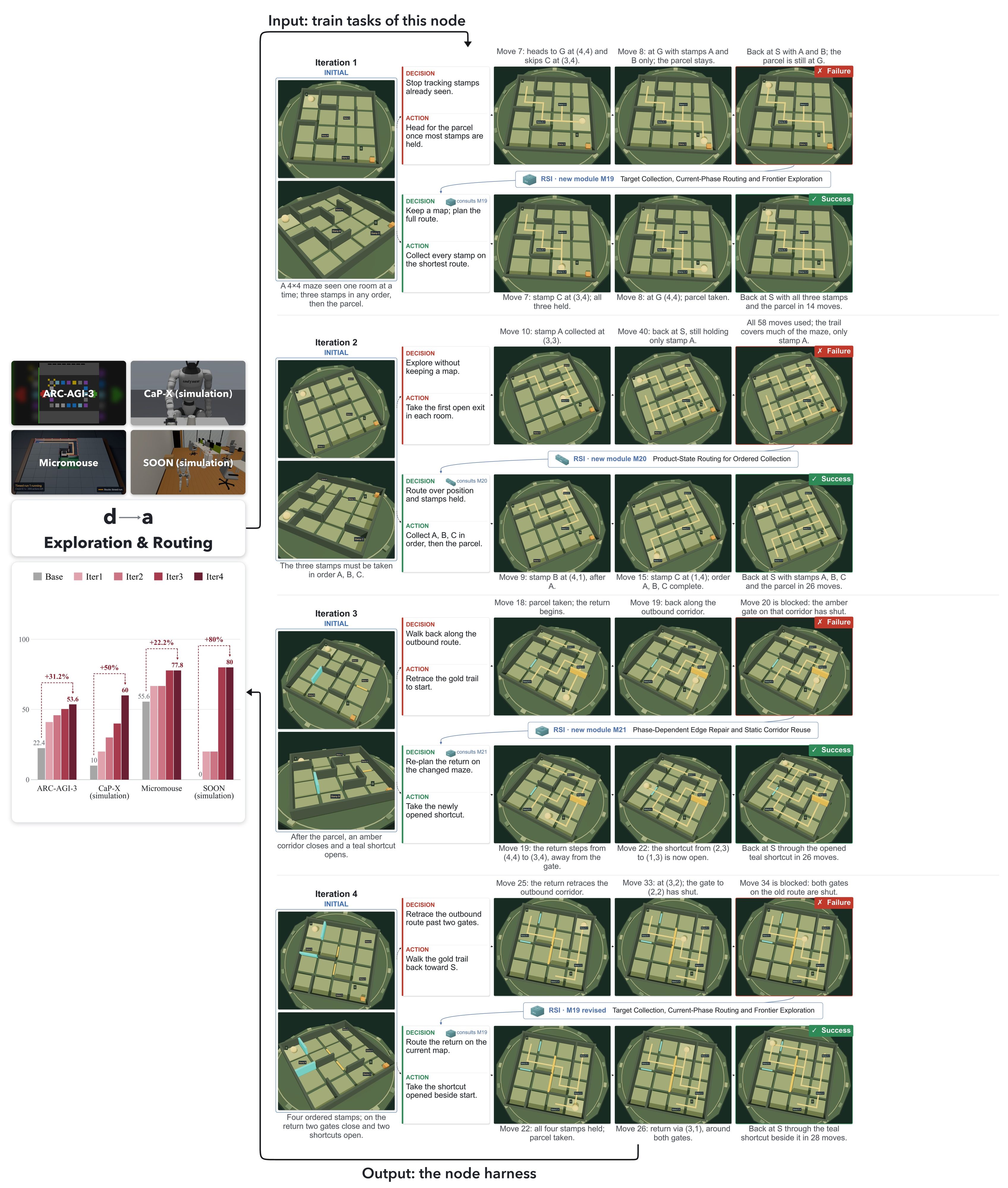}}
\captionof{figure}{\textbf{d$\to$a: Exploration \& Routing}}
\end{minipage}\par\vspace{6pt}
\noindent\begin{minipage}{\linewidth}\centering
\makebox[\linewidth][c]{\includegraphics[width=1.18\linewidth,height=0.9\textheight,keepaspectratio]{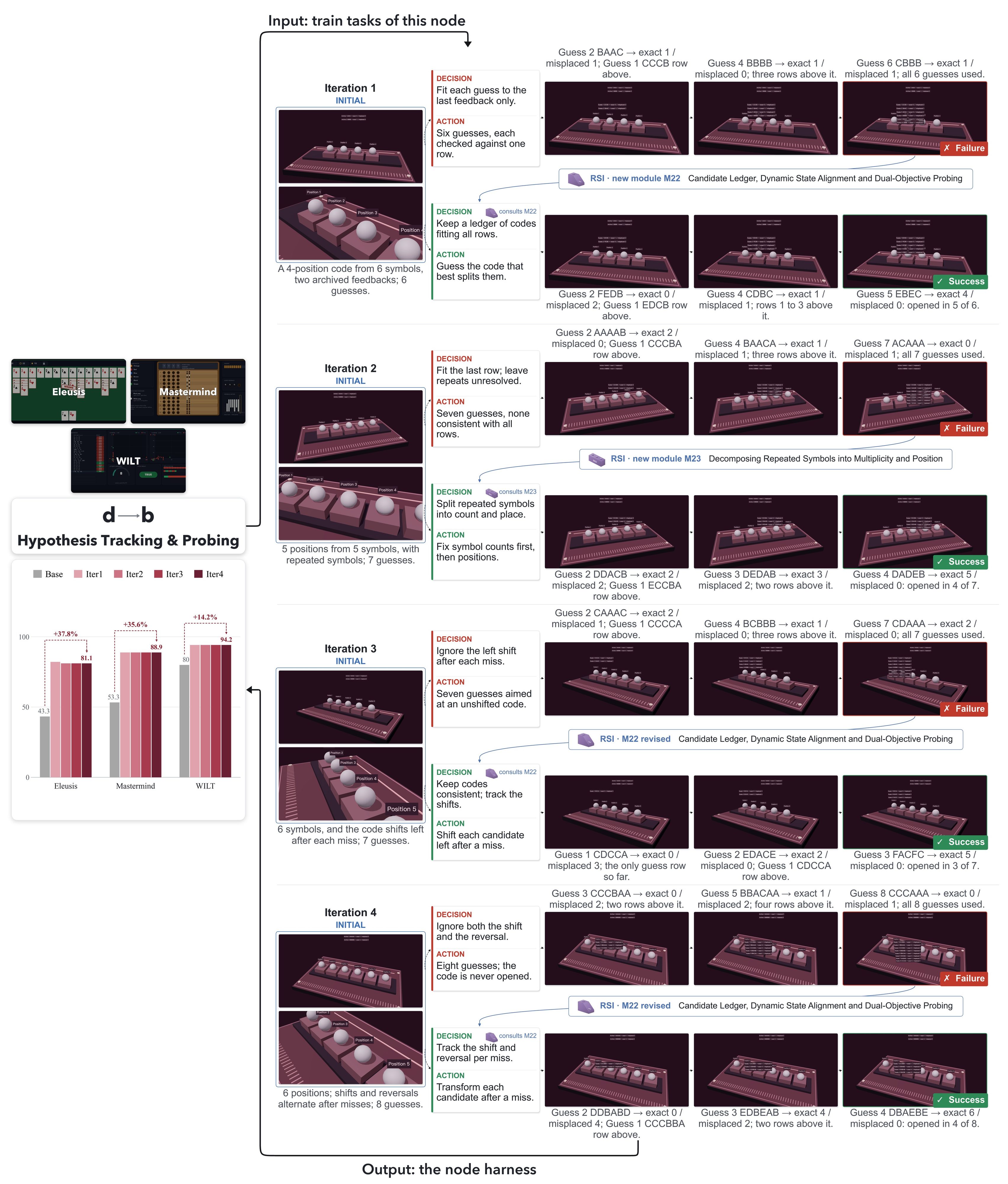}}
\captionof{figure}{\textbf{d$\to$b: Hypothesis Tracking \& Probing}}
\end{minipage}\par\vspace{6pt}
\noindent\begin{minipage}{\linewidth}\centering
\makebox[\linewidth][c]{\includegraphics[width=1.18\linewidth,height=0.9\textheight,keepaspectratio]{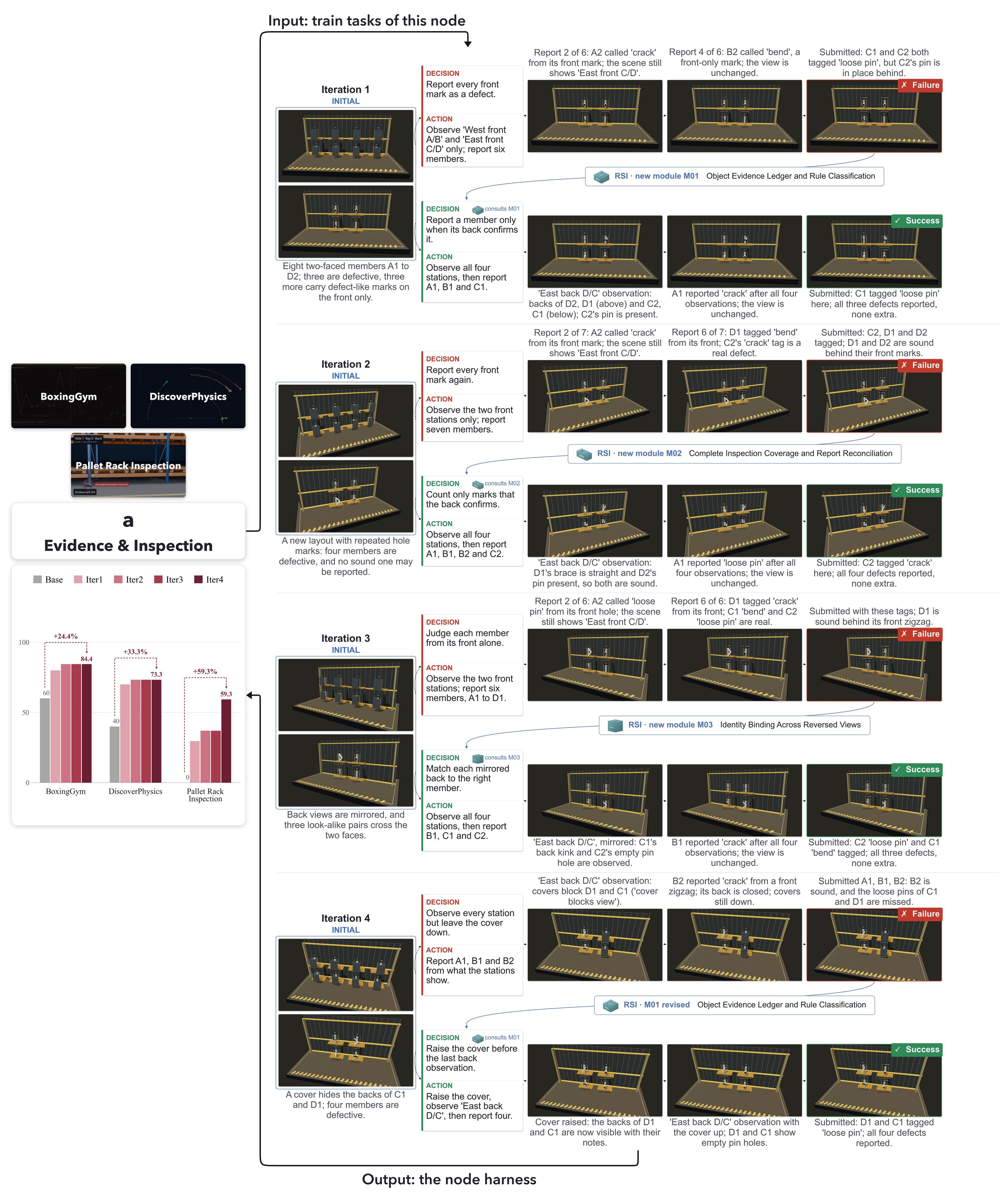}}
\captionof{figure}{\textbf{a: Evidence \& Inspection}}
\end{minipage}\par\vspace{6pt}
\noindent\begin{minipage}{\linewidth}\centering
\makebox[\linewidth][c]{\includegraphics[width=1.18\linewidth,height=0.9\textheight,keepaspectratio]{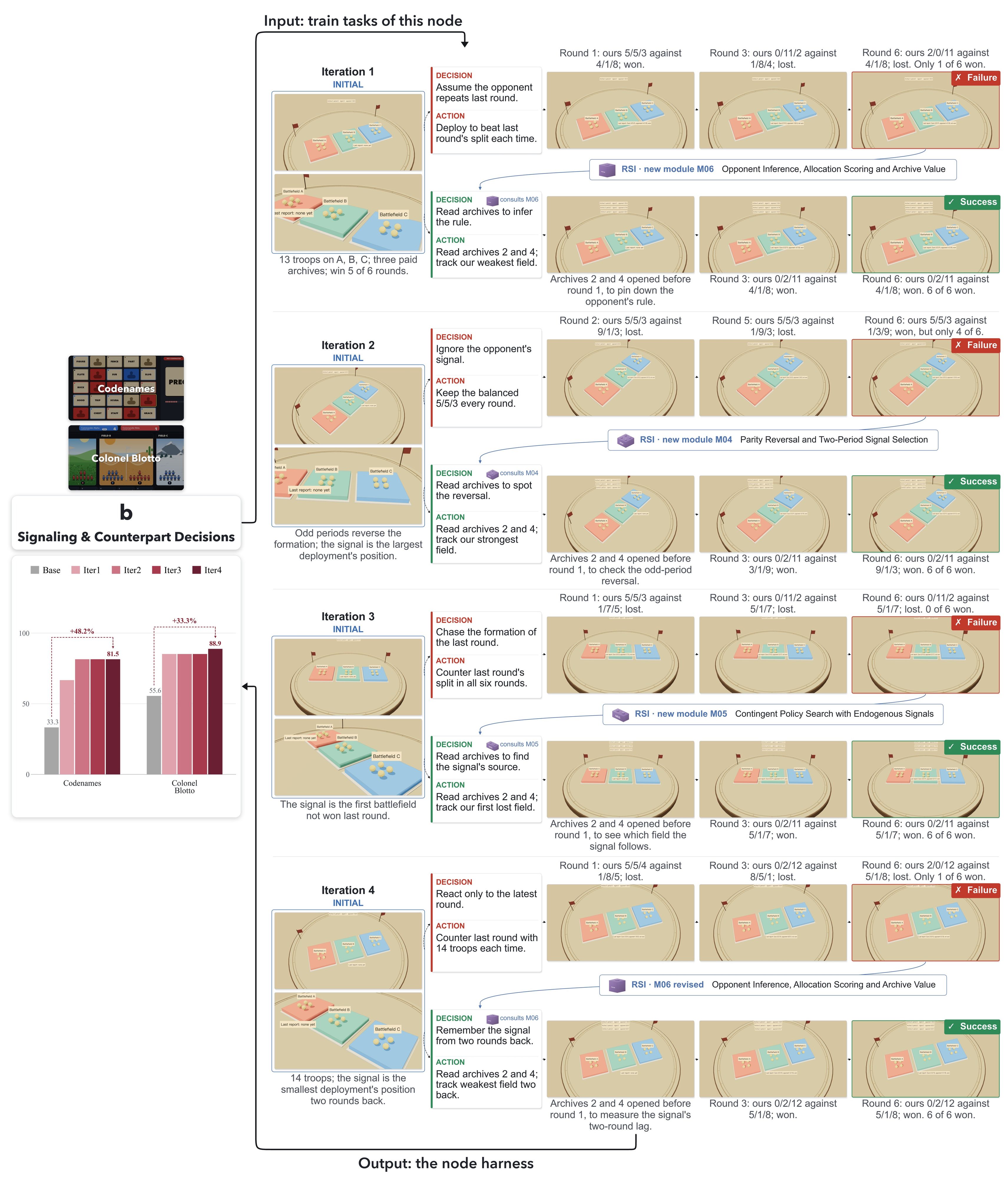}}
\captionof{figure}{\textbf{b: Signaling \& Counterpart Decisions}}
\end{minipage}\par\vspace{6pt}
\noindent\begin{minipage}{\linewidth}\centering
\makebox[\linewidth][c]{\includegraphics[width=1.18\linewidth,height=0.9\textheight,keepaspectratio]{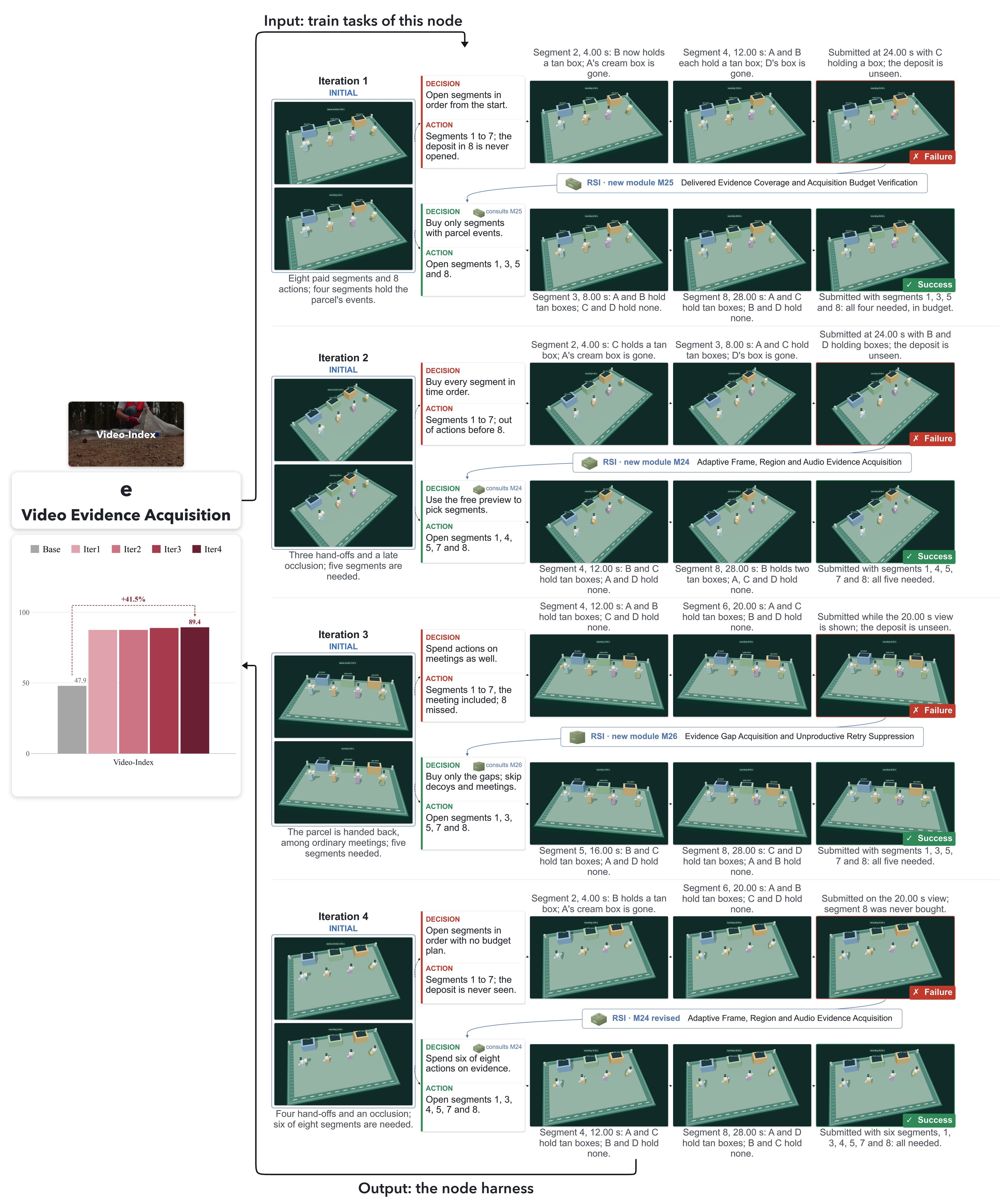}}
\captionof{figure}{\textbf{e: Video Evidence Acquisition}}
\end{minipage}\par\vspace{6pt}
\noindent\begin{minipage}{\linewidth}\centering
\makebox[\linewidth][c]{\includegraphics[width=1.18\linewidth,height=0.9\textheight,keepaspectratio]{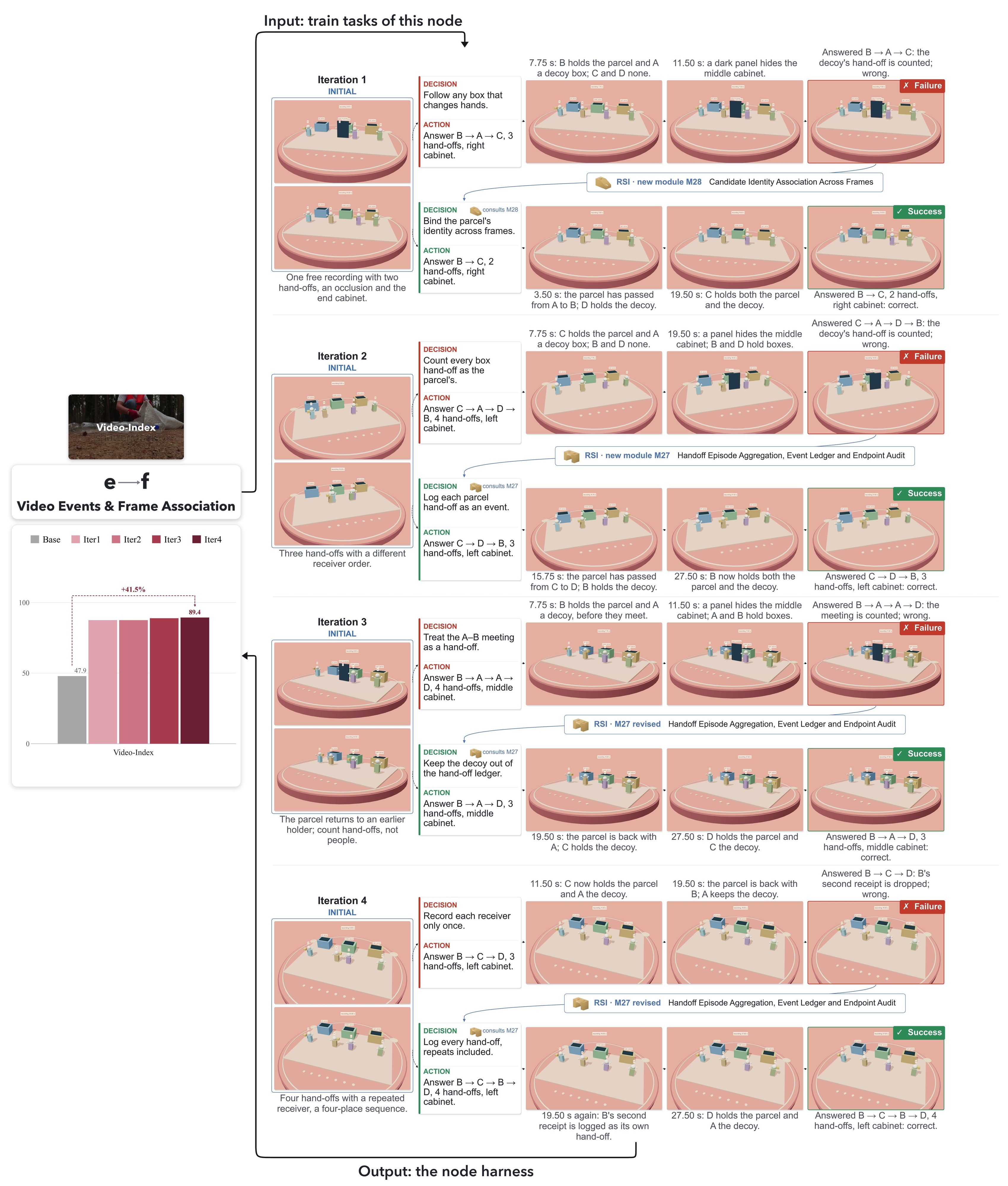}}
\captionof{figure}{\textbf{e$\to$f: Video Events \& Frame Association}}\label{fig:node-last}
\end{minipage}\par\vspace{6pt}
}

\FloatBarrier
\subsection{Embodied Benchmarks}
\label{app:embodied}
\begin{figure}[!htbp]
\centering
% 2x2: SOON real (top left) and its simulation from the same viewpoint (bottom left); CaP-X real (top right) and AlohaMini2 simulation (bottom right)
\IfFileExists{appendix/figures/embodied/real-sim-scenes.jpg}{\includegraphics[width=0.9\linewidth]{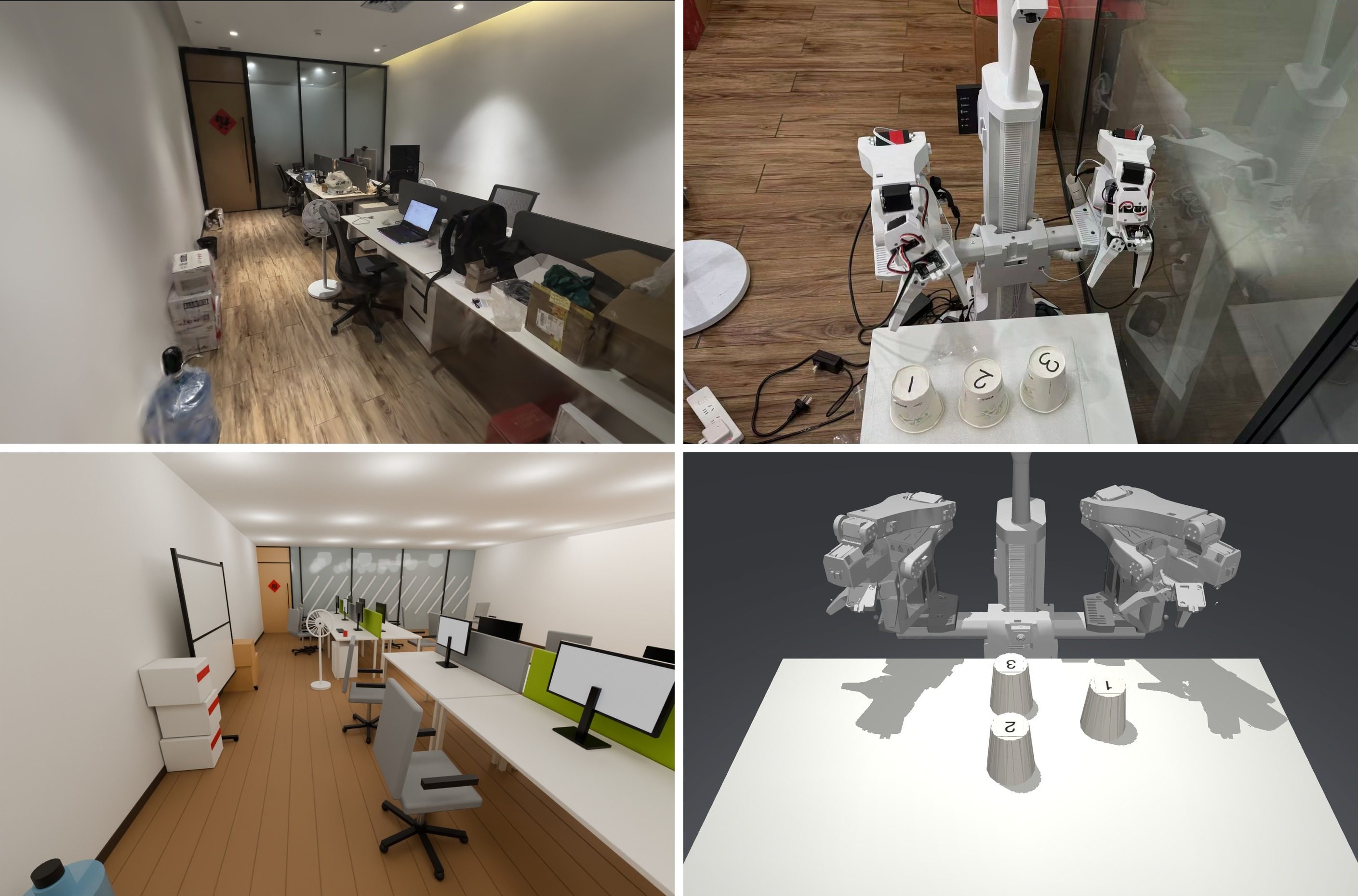}}{%
{\setlength{\fboxsep}{0pt}\color{black!8}\rule{0.49\linewidth}{0.3\linewidth}}\hfill{\color{black!8}\rule{0.49\linewidth}{0.3\linewidth}}\par\vspace{4pt}
{\color{black!8}\rule{0.49\linewidth}{0.3\linewidth}}\hfill{\color{black!8}\rule{0.49\linewidth}{0.3\linewidth}}}
\caption{\textbf{Embodied environments.} Real scenes appear above their simulated counterparts, with SOON on the left and CaP-X on the right.}
\label{fig:real-sim}
\end{figure}

The simulation of each embodied benchmark runs in an environment built from its real scene, which closely reproduces the layout and furnishings of the site, with a model of the same robot as the real runs: a Unitree G1 for SOON and an AlohaMini2 for CaP-X. For SOON we rebuilt the office from its floor plan and from photographs and video of the site: the side room where the robot starts, the doorway into the office, the rows of desks with their chairs and cabinets, and the white pedestal fan beside the target desk. The office is a 3D scene in which the walls and furniture are collision obstacles: each action moves the robot by the nominal distance or angle of the real robot's motion profile, a move that hits an obstacle counts as a collision and ends the episode as a failure, and after every action the agent receives the head-camera image rendered from the robot's new pose with the height, downward pitch, and field of view of the real camera. On the real robot a person judges every episode on site, and an episode in which the robot collides with an obstacle is a failure. The G1 starts in the side room, walks through the doorway into the office, and must stop at the desk next to the white fan and touch it within 30 movement actions. For CaP-X we built the table scene in the MuJoCo physics simulator~\citep{todorov2012mujoco} with the official AlohaMini2 model, and with objects of the same size and colour as the real ones: three paper cups, 9\,cm tall, 7\,cm across the mouth, and 5\,cm across the closed bottom, stand upside down on a white table, and in most layouts a 2\,cm cube, red, green, or blue, lies under one of them. The AlohaMini2 stands at the near edge of the table with its two six-joint arms; its mobile base and lift stay still. After every action the agent sees the head camera, the wrist cameras on both grippers, and a fixed side camera; it moves a gripper in small steps while the host solves the arm's inverse kinematics, and within 30 actions it must report whether a cube is present and, if it is, its colour and the cup it is under. Figure~\ref{fig:real-sim} compares the real and simulated scenes directly.

\begin{figure}[!htbp]
\centering
\includegraphics[width=0.85\linewidth]{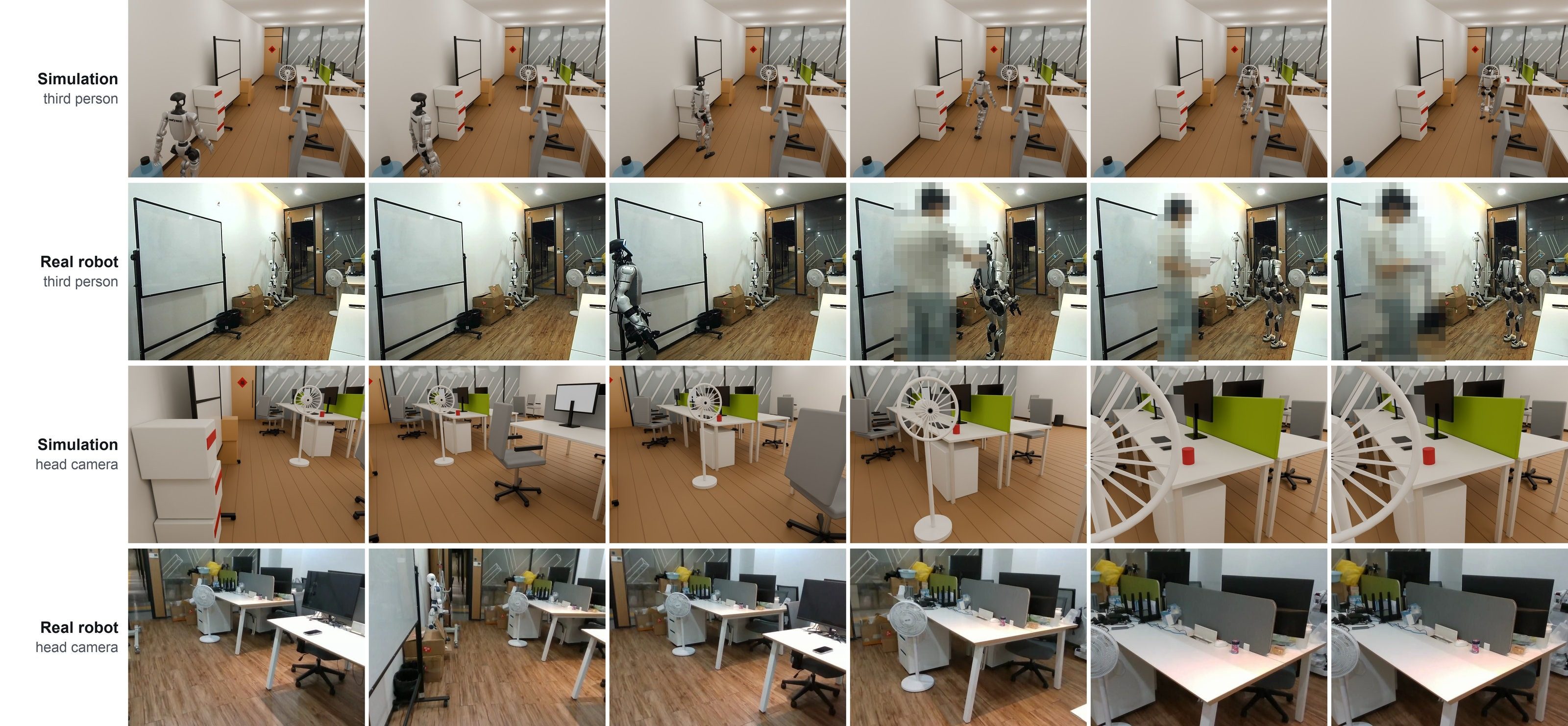}
\caption{\textbf{SOON with the Base Harness.} Rows show simulation and real-robot views from the fixed office camera, followed by the head camera. The robot walks from the side room along the aisle, looks around, and turns towards the desk beside the fan, but reaches out before the desk is within reach in both settings.}
\label{fig:soon-h0}
\end{figure}

\begin{figure}[!htbp]
\centering
\includegraphics[width=0.85\linewidth]{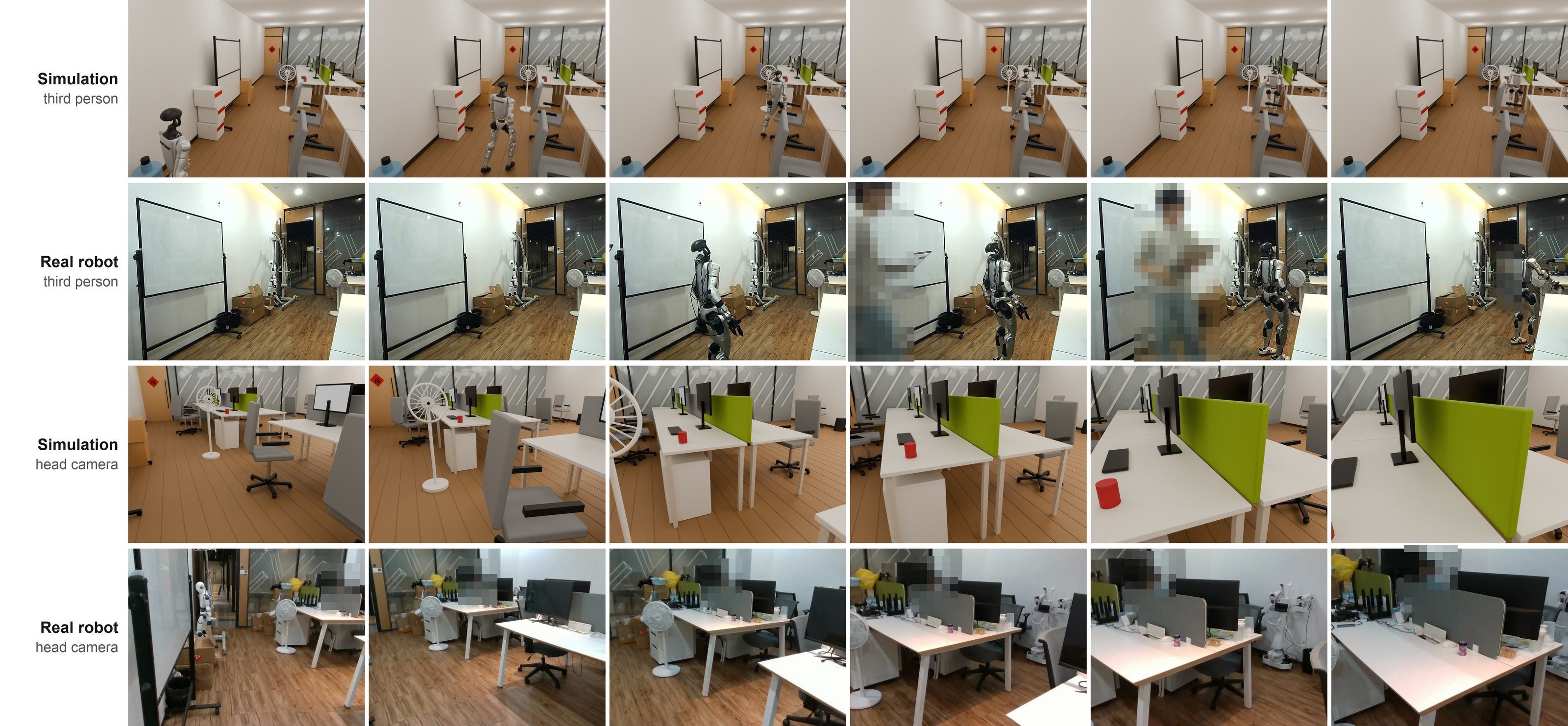}
\caption{\textbf{SOON with the evolved Harness.} Rows follow Figure~\ref{fig:soon-h0}. In simulation the robot turns right past the doorway, walks diagonally towards the desk beside the fan, and slows to small steps before touching it. The real robot also approaches in small steps and completes the task.}
\label{fig:soon-h4}
\end{figure}

\begin{figure}[!htbp]
\centering
\includegraphics[width=0.8\linewidth]{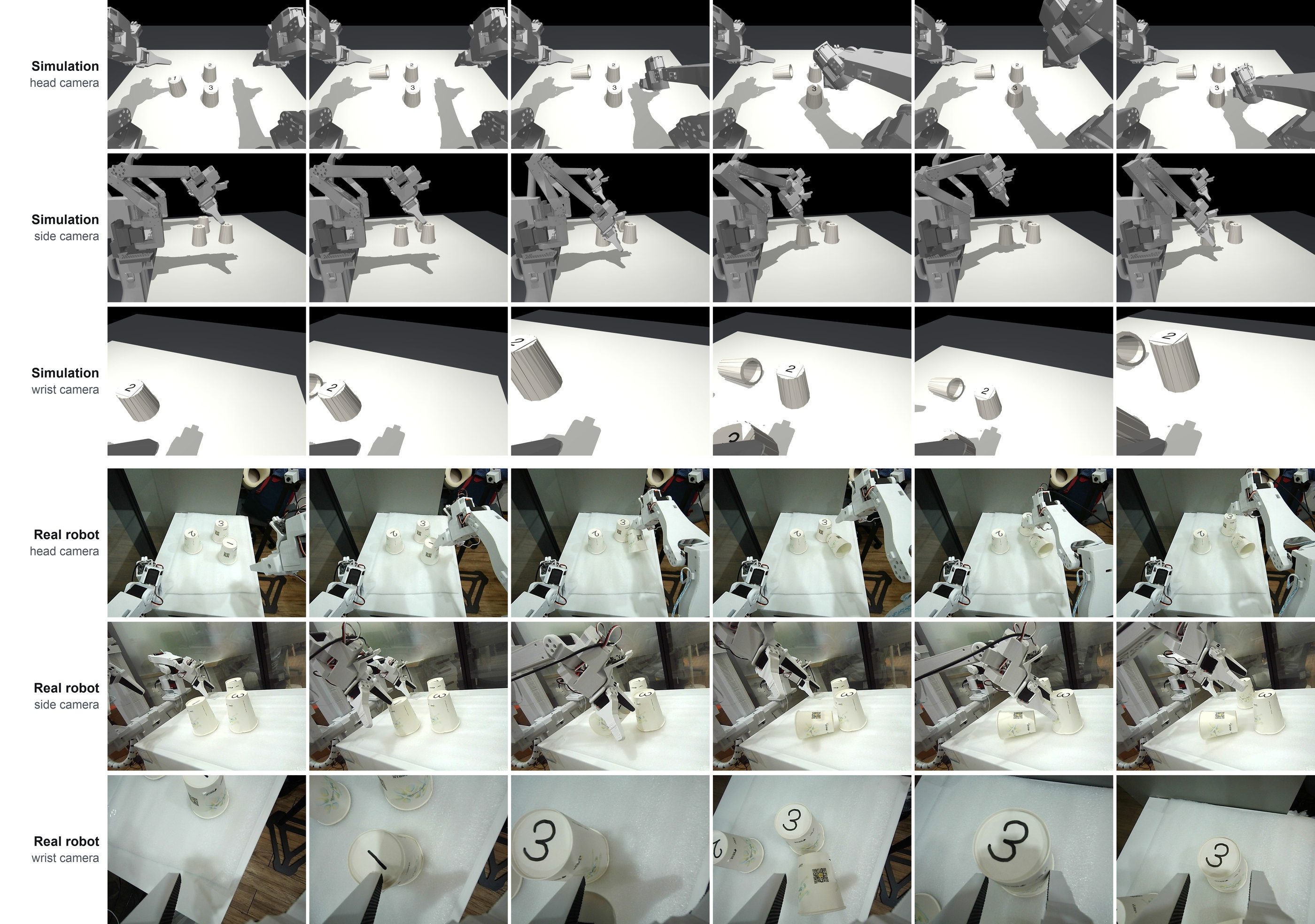}
\caption{\textbf{CaP-X with the Base Harness.} Rows show head, side, and wrist cameras in simulation, followed by the real robot. In both settings the robot manipulates other cups without inspecting the target cup and incorrectly reports that no cube is present.}
\label{fig:capx-h0}
\end{figure}

\begin{figure}[!htbp]
\centering
\includegraphics[width=0.8\linewidth]{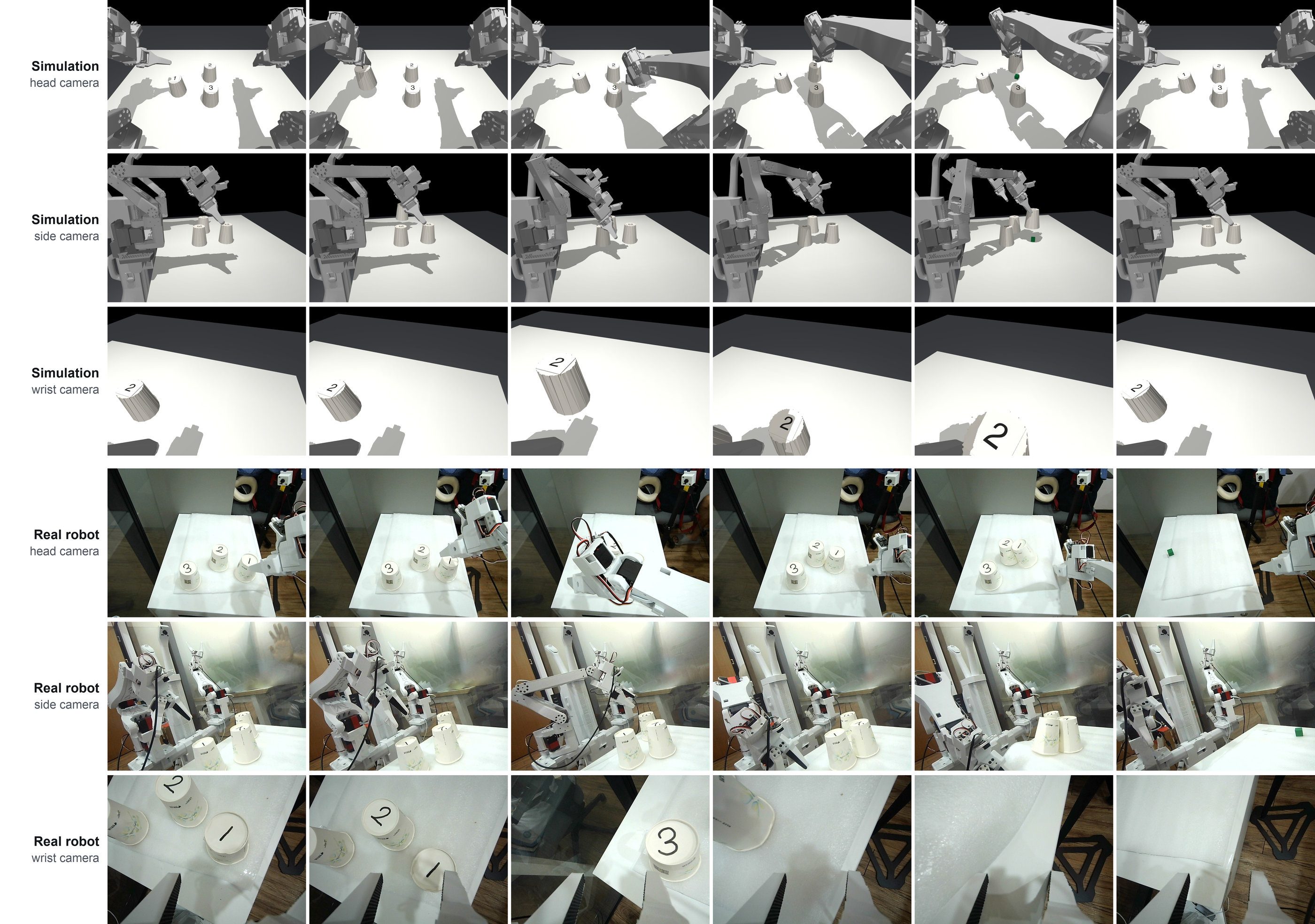}
\caption{\textbf{CaP-X with the evolved Harness.} Rows follow Figure~\ref{fig:capx-h0}. In simulation the robot checks the cups and retries a missed grasp until the cube is visible. On the real robot it sweeps the cups aside to uncover the cube; both runs then report its colour and location.}
\label{fig:capx-h4}
\end{figure}

Figures~\ref{fig:soon-h0}--\ref{fig:capx-h4} compare H$_0$ and H$^{(4)}$ under the same task settings. In SOON, the robot must judge the distance to the target desk from head-camera images alone. H$_0$ looks around and reaches out as soon as the desk looks close, stopping about 0.5\,m short in simulation and out of reach on the real robot. H$^{(4)}$ turns towards the desk right after the doorway and advances in small steps until the desk is directly ahead, so that in both settings the hand ends within the range judged as a successful touch. In CaP-X, the cube is hidden under upside-down cups and can only be found by uncovering them. H$_0$ grasps clumsily, knocks cup 1 over, tries cup 3, and then answers that there is no cube without ever examining cup 2. H$^{(4)}$ checks the cups one by one and answers only after seeing the cube: in simulation it lifts cup 1, finds it empty, retries a missed grasp on cup 2, and lifts it to reveal the green cube; on the real robot it pushes the cups aside until the cube lies uncovered. In both benchmarks, H$^{(4)}$ seeks direct evidence before committing and thus completes the task.

%% file: appendix/tables/defects.tex
\vspace{6pt}\begin{center}\footnotesize\setlength{\tabcolsep}{5pt}
\captionof{table}{\textbf{Capability and defect taxonomy.} Columns count the training and unseen benchmarks whose trajectories expose each defect type; unseen counts come from H$_0$ runs recorded after all Harness training.}\label{tab:defects}
\begin{tabular}{@{}p{0.70\linewidth}rr@{}}
 & Training & Unseen \\
\midrule
\textcolor{trinityred}{\textbf{a \ Perception \& State Tracking}} & & \\
\quad F01 \ Host action-protocol and output-format fidelity & 4 & 5 \\
\quad F02 \ Observation-to-state grounding & 3 & 4 \\
\quad F03 \ Action and parameter effect calibration & 4 & 3 \\
\quad F04 \ Instruction-to-action grounding & 2 & 0 \\
\quad F05 \ Outcome ledger: no re-issue of refuted or inert items & 5 & 5 \\
\quad F06 \ Re-acquisition of out-of-window facts & 7 & 0 \\
\quad F07 \ Exploration coverage and route reuse & 4 & 4 \\[4pt]
\textcolor{trinityred}{\textbf{b \ Hypothesis \& Counterpart Inference}} & & \\
\quad F08 \ Evidence-consistent hypothesis maintenance & 3 & 1 \\
\quad F09 \ Discriminating probe design & 3 & 2 \\
\quad F10 \ Data-to-model inference with assumption checks & 1 & 1 \\
\quad F11 \ Counterpart modelling from revealed history & 4 & 2 \\
\quad F12 \ Receiver-aware signalling & 3 & 1 \\[4pt]
\textcolor{trinityred}{\textbf{c \ Constraint Checking \& Feedback Repair}} & & \\
\quad F13 \ Pre-commit one-step constraint check & 6 & 5 \\
\quad F14 \ Multi-step constraint propagation to forced moves & 1 & 0 \\
\quad F15 \ Reachability lookahead before irreversible action & 2 & 1 \\
\quad F16 \ Feasibility-checked complete plan & 1 & 2 \\
\quad F17 \ Feedback-localized repair & 2 & 2 \\
\quad F18 \ Non-convergence detection and method switching & 9 & 9 \\[4pt]
\textcolor{trinityred}{\textbf{d \ Goal Commitment \& Value Trade-off}} & & \\
\quad F19 \ Evidence-gated commitment & 4 & 5 \\
\quad F20 \ Goal-discrepancy-directed action selection & 4 & 0 \\
\quad F21 \ Goal-completing opportunity recognition & 3 & 1 \\
\quad F22 \ Goal-predicate verification before terminal action & 4 & 1 \\
\quad F23 \ Budget pacing and terminal-action reservation & 5 & 4 \\
\quad F24 \ Scored-objective valuation of options and policies & 6 & 6 \\[4pt]
\textcolor{trinityred}{\textbf{e \ Video Evidence Acquisition}} & & \\
\quad V01 \ Question-guided evidence acquisition & 1 & 1 \\
\quad V02 \ Coverage-adequate frame sampling & 1 & 0 \\
\quad V03 \ Non-visual channel access & 1 & 1 \\[4pt]
\textcolor{trinityred}{\textbf{f \ Video Comprehension}} & & \\
\quad V04 \ Frame-grounded perception and order reading & 1 & 1 \\
\quad V05 \ Motion, counting and 3D dynamics across frames & 1 & 1 \\
\quad V06 \ Cross-frame event integration and inference & 1 & 1 \\
\end{tabular}
\end{center}

%% file: appendix/tables/results.tex
\begin{scriptsize}\setlength{\tabcolsep}{1.5pt}\setlength{\LTleft}{\fill}\setlength{\LTright}{\fill}\setlength{\LTpre}{10pt}\setlength{\LTpost}{0pt}
\setlength{\LTcapwidth}{\linewidth}
\begin{longtable}{@{}llrrrrrrr@{}}
\caption{\textbf{Full benchmark results.} Scores on a 0--100 scale for the Base Harness H$_0$, the Harnesses H$^{(1)}$--H$^{(4)}$ after each RSI iteration, and the Codex and DeepSeek Harnesses on the unseen benchmarks; -- marks runs that were not performed.}\label{tab:results}\\
Benchmark & Measure, instances, episodes & Base H$_0$ & H$^{(1)}$ & H$^{(2)}$ & H$^{(3)}$ & H$^{(4)}$ & Codex & DeepSeek \\
\midrule
\endfirsthead
Benchmark & Measure, instances, episodes & Base H$_0$ & H$^{(1)}$ & H$^{(2)}$ & H$^{(3)}$ & H$^{(4)}$ & Codex & DeepSeek \\
\midrule
\endhead
\midrule\multicolumn{9}{r}{\emph{continued on the next page}}\\
\endfoot
\endlastfoot
\multicolumn{9}{@{}l}{\textbf{Train}} \\*
AdaPlanBench~\citep{liu2026adaplanbench} & accuracy, 3 queries, 27 episodes & 22.2\phantom{0} & 70.4\phantom{0} & \textbf{74.1}\phantom{0} & \textbf{74.1}\phantom{0} & \textbf{74.1}\phantom{0} & \hmiss & \hmiss \\
Among Us~\citep{golechha2025amongus} & win rate, 3 games, 27 episodes & 11.1\phantom{0} & 48.1\phantom{0} & \textbf{63}\phantom{.00} & \textbf{63}\phantom{.00} & \textbf{63}\phantom{.00} & \hmiss & \hmiss \\
ARC-AGI-3~\citep{arcprize2026arcagi3} & success rate, 25 games, 183 levels & 22.4\phantom{0} & 41\phantom{.00} & 45.9\phantom{0} & 50.3\phantom{0} & \textbf{53.6}\phantom{0} & \hmiss & \hmiss \\
BoxingGym~\citep{gandhi2025boxinggym} & success rate, 3 environments, 45 episodes & 60\phantom{.00} & 80\phantom{.00} & \textbf{84.4}\phantom{0} & \textbf{84.4}\phantom{0} & \textbf{84.4}\phantom{0} & \hmiss & \hmiss \\
Chess Tactics*~\citep{lichess2026puzzles} & mate found, 3 positions, 27 episodes & 66.7\phantom{0} & 92.6\phantom{0} & 96.3\phantom{0} & 96.3\phantom{0} & \textbf{100}\phantom{.00} & \hmiss & \hmiss \\
Codenames~\citep{stephenson2025codenames} & win rate, 3 games, 27 episodes & 33.3\phantom{0} & 66.7\phantom{0} & \textbf{81.5}\phantom{0} & \textbf{81.5}\phantom{0} & \textbf{81.5}\phantom{0} & \hmiss & \hmiss \\
Colonel Blotto~\citep{guertler2025textarena} & win rate, 3 matches, 27 episodes & 55.6\phantom{0} & 85.2\phantom{0} & 85.2\phantom{0} & 85.2\phantom{0} & \textbf{88.9}\phantom{0} & \hmiss & \hmiss \\
DiscoverPhysics~\citep{wiemann2026discoverphysics} & pass rate, 2 worlds, 30 episodes & 40\phantom{.00} & 70\phantom{.00} & \textbf{73.3}\phantom{0} & \textbf{73.3}\phantom{0} & \textbf{73.3}\phantom{0} & \hmiss & \hmiss \\
Eleusis*~\citep{golden2020eleusisexpress} & rule found, 6 rules, 90 episodes & 43.3\phantom{0} & \textbf{82.2}\phantom{0} & 81.1\phantom{0} & 81.1\phantom{0} & 81.1\phantom{0} & \hmiss & \hmiss \\
Euro NCAP AEB*~\citep{euroncap2023aeb} & success rate, 3 scenarios, 9 episodes & 66.7\phantom{0} & 77.8\phantom{0} & \textbf{88.9}\phantom{0} & \textbf{88.9}\phantom{0} & \textbf{88.9}\phantom{0} & \hmiss & \hmiss \\
GLEE Persuasion~\citep{shapira2024glee} & revenue $\geq$ 50\%, 3 games, 27 episodes & 33.3\phantom{0} & \textbf{63}\phantom{.00} & \textbf{63}\phantom{.00} & \textbf{63}\phantom{.00} & \textbf{63}\phantom{.00} & \hmiss & \hmiss \\
Master Keying*~\citep{pulford2007locks} & success rate, 3 designs, 33 episodes & 45.5\phantom{0} & \textbf{84.8}\phantom{0} & \textbf{84.8}\phantom{0} & \textbf{84.8}\phantom{0} & \textbf{84.8}\phantom{0} & \hmiss & \hmiss \\
Mastermind~\citep{golde2025mastermindeval} & solve rate, 3 codes, 45 episodes & 53.3\phantom{0} & \textbf{88.9}\phantom{0} & \textbf{88.9}\phantom{0} & \textbf{88.9}\phantom{0} & \textbf{88.9}\phantom{0} & \hmiss & \hmiss \\
Memory Maze~\citep{pasukonis2022memorymaze} & success rate, 3 mazes, 9 episodes & 0\phantom{.00} & \textbf{11.1}\phantom{0} & \textbf{11.1}\phantom{0} & \textbf{11.1}\phantom{0} & \textbf{11.1}\phantom{0} & \hmiss & \hmiss \\
Micromouse*~\citep{ukmarsMicromouseRules} & success rate, 3 mazes, 27 episodes & 55.6\phantom{0} & 66.7\phantom{0} & 66.7\phantom{0} & \textbf{77.8}\phantom{0} & \textbf{77.8}\phantom{0} & \hmiss & \hmiss \\
Pallet Rack Inspection*~\citep{en15635} & success rate, 3 racks, 27 episodes & 0\phantom{.00} & 29.6\phantom{0} & 37\phantom{.00} & 37\phantom{.00} & \textbf{59.3}\phantom{0} & \hmiss & \hmiss \\
PInVerify~\citep{jiang2026pinverify} & accuracy, 3 pairs, 45 episodes & 66.7\phantom{0} & 88.9\phantom{0} & \textbf{93.3}\phantom{0} & \textbf{93.3}\phantom{0} & \textbf{93.3}\phantom{0} & \hmiss & \hmiss \\
Pooltool Billiards~\citep{xu2025deepphy,kiefl2024pooltool} & success rate, 3 seeds, 27 episodes & 33.3\phantom{0} & \textbf{66.7}\phantom{0} & \textbf{66.7}\phantom{0} & \textbf{66.7}\phantom{0} & \textbf{66.7}\phantom{0} & \hmiss & \hmiss \\
puzz.link / Penpa+~\citep{robx2026pzprjs,penpa2026} & solve rate, 3 puzzles, 9 episodes & 33.3\phantom{0} & \textbf{44.4}\phantom{0} & \textbf{44.4}\phantom{0} & \textbf{44.4}\phantom{0} & \textbf{44.4}\phantom{0} & \hmiss & \hmiss \\
PuzzleScript~\citep{lavelle2026puzzlescript} & solve rate, 3 levels, 9 episodes & 66.7\phantom{0} & \textbf{77.8}\phantom{0} & \textbf{77.8}\phantom{0} & \textbf{77.8}\phantom{0} & \textbf{77.8}\phantom{0} & \hmiss & \hmiss \\
Saboteur*~\citep{moyersoen2023saboteur} & win rate, 3 games, 27 episodes & 55.6\phantom{0} & \textbf{77.8}\phantom{0} & \textbf{77.8}\phantom{0} & \textbf{77.8}\phantom{0} & \textbf{77.8}\phantom{0} & \hmiss & \hmiss \\
TextQuests~\citep{phan2025textquests} & success rate, 3 games, 9 episodes & 0\phantom{.00} & 11.1\phantom{0} & 11.1\phantom{0} & 11.1\phantom{0} & \textbf{22.2}\phantom{0} & \hmiss & \hmiss \\
The Long Dark*~\citep{hinterland2017longdark} & success rate, 3 runs, 27 episodes & 77.8\phantom{0} & \textbf{88.9}\phantom{0} & \textbf{88.9}\phantom{0} & \textbf{88.9}\phantom{0} & \textbf{88.9}\phantom{0} & \hmiss & \hmiss \\
Video-Index~\citep{song2026videoindex} & accuracy, 777 questions, 777 episodes & 47.9\phantom{0} & 87.5\phantom{0} & 87.5\phantom{0} & 88.9\phantom{0} & \textbf{89.4}\phantom{0} & \hmiss & \hmiss \\
WCA Cube Solving*~\citep{wca2025regulations} & solve rate, 3 scrambles, 45 episodes & 13.3\phantom{0} & 51.1\phantom{0} & \textbf{80}\phantom{.00} & \textbf{80}\phantom{.00} & \textbf{80}\phantom{.00} & \hmiss & \hmiss \\
WILT~\citep{banatt2024wilt} & accuracy, 8 rules, 120 episodes & 80\phantom{.00} & \textbf{94.2}\phantom{0} & \textbf{94.2}\phantom{0} & \textbf{94.2}\phantom{0} & \textbf{94.2}\phantom{0} & \hmiss & \hmiss \\
SOON (simulation)~\citep{zhu2021soon} & success rate, 1 route, 10 episodes & 0\phantom{.00} & 20\phantom{.00} & 20\phantom{.00} & \textbf{80}\phantom{.00} & \textbf{80}\phantom{.00} & \hmiss & \hmiss \\
CaP-X (simulation)~\citep{fu2026capx} & success rate, 10 layouts, 10 episodes & 10\phantom{.00} & 20\phantom{.00} & 30\phantom{.00} & 40\phantom{.00} & \textbf{60}\phantom{.00} & \hmiss & \hmiss \\

\pagebreak
\addlinespace[6pt]\multicolumn{9}{@{}l}{\textbf{Unseen}} \\*
ActiView~\citep{wang2025actiview} & accuracy, 1,625 questions, 1,625 episodes & 73.5\phantom{0} & 93.7\phantom{0} & 94.6\phantom{0} & 95.7\phantom{0} & \textbf{97.4}\phantom{0} & 77.2\phantom{0} & 73.7\phantom{0} \\
ALE-Bench~\citep{imajuku2025alebench} & percentile, 26 problems, 26 episodes & 81.3\phantom{0} & 85.4\phantom{0} & 86.4\phantom{0} & 87.4\phantom{0} & \textbf{91.9}\phantom{0} & 85\phantom{.00} & 91.4\phantom{0} \\
BuildArena~\citep{xia2026buildarena} & pass rate, 9 levels, 45 episodes & 73.3\phantom{0} & 93.3\phantom{0} & 95.6\phantom{0} & 95.6\phantom{0} & \textbf{97.8}\phantom{0} & 77.8\phantom{0} & 80\phantom{.00} \\
Coup~\citep{guertler2025textarena} & win rate, 40 games, 40 episodes & 45\phantom{.00} & 57.5\phantom{0} & 57.5\phantom{0} & \textbf{65}\phantom{.00} & \textbf{65}\phantom{.00} & 32.5\phantom{0} & 35\phantom{.00} \\
Crafter~\citep{hafner2022crafter} & progress, 10 seeds, 10 episodes & 51\phantom{.00} & 56\phantom{.00} & 57\phantom{.00} & 58\phantom{.00} & 60\phantom{.00} & \textbf{82.7}\phantom{0} & 70\phantom{.00} \\
DiG-bench~\citep{battleday2026digbench} & levels passed, 13 games, 13 episodes & 32.5\phantom{0} & 33.3\phantom{0} & 34.2\phantom{0} & 35.1\phantom{0} & 36.8\phantom{0} & 40.4\phantom{0} & \textbf{54.4}\phantom{0} \\
Entity Deduction Arena~\citep{zhang2024eda} & success rate, 200 entities, 200 episodes & 64.5\phantom{0} & \textbf{95}\phantom{.00} & \textbf{95}\phantom{.00} & 94.5\phantom{0} & \textbf{95}\phantom{.00} & 72.5\phantom{0} & 67.5\phantom{0} \\
Hanabi~\citep{bard2020hanabi} & success rate, 40 games, 40 episodes & 0\phantom{.00} & 5\phantom{.00} & 10\phantom{.00} & 15\phantom{.00} & \textbf{20}\phantom{.00} & 0\phantom{.00} & 0\phantom{.00} \\
Heads-Up NL Hold'em~\citep{deepmind2026gamearena,lanctot2019openspiel} & matches ahead, 20 matches, 20 episodes & 60\phantom{.00} & 80\phantom{.00} & 80\phantom{.00} & 80\phantom{.00} & \textbf{85}\phantom{.00} & 30\phantom{.00} & 35\phantom{.00} \\
MiniHack Boxoban~\citep{samvelyan2021minihack} & solve rate, 10 levels, 10 episodes & 0\phantom{.00} & 10\phantom{.00} & 20\phantom{.00} & 20\phantom{.00} & \textbf{30}\phantom{.00} & 10\phantom{.00} & 0\phantom{.00} \\
MiniHack Maze / Corridor~\citep{samvelyan2021minihack} & success rate, 20 levels, 20 episodes & 55\phantom{.00} & 80\phantom{.00} & 80\phantom{.00} & 80\phantom{.00} & \textbf{85}\phantom{.00} & 60\phantom{.00} & 75\phantom{.00} \\
NetHack~\citep{kuttler2020nethack} & progress, 5 seeds, 5 episodes, 500 steps & 2.2\phantom{0} & 3\phantom{.00} & 3\phantom{.00} & 3\phantom{.00} & \textbf{4}\phantom{.00} & 3.7\phantom{0} & 2.3\phantom{0} \\
Pencil Puzzle Bench~\citep{waugh2026ppbench} & solve rate, 30 puzzles, 30 episodes & 20\phantom{.00} & 26.7\phantom{0} & 26.7\phantom{0} & 26.7\phantom{0} & \textbf{30}\phantom{.00} & 23.3\phantom{0} & 23.3\phantom{0} \\
PHYRE~\citep{bakhtin2019phyre} & solved, 50 tasks, 50 episodes & 42\phantom{.00} & 82\phantom{.00} & 86\phantom{.00} & 86\phantom{.00} & \textbf{90}\phantom{.00} & 84\phantom{.00} & 52\phantom{.00} \\
SciGym~\citep{duan2025scigym} & recovered, 126 systems, 126 episodes & 25.4\phantom{0} & 45.2\phantom{0} & 47.6\phantom{0} & 50\phantom{.00} & \textbf{57.1}\phantom{0} & 26.2\phantom{0} & 33.3\phantom{0} \\
Video-MME-v2~\citep{fu2026videommev2} & accuracy, 1,600 questions, 1,600 episodes & 48.4\phantom{0} & 77.4\phantom{0} & 79.5\phantom{0} & \textbf{80}\phantom{.00} & \textbf{80}\phantom{.00} & 54.5\phantom{0} & 45.9\phantom{0} \\
SOON (robot)~\citep{zhu2021soon} & success rate, 1 route, 10 episodes & 0\phantom{.00} & 20\phantom{.00} & 20\phantom{.00} & \textbf{60}\phantom{.00} & \textbf{60}\phantom{.00} & \hmiss & \hmiss \\
CaP-X (robot)~\citep{fu2026capx} & success rate, 10 layouts, 10 episodes & 0\phantom{.00} & 20\phantom{.00} & 20\phantom{.00} & 40\phantom{.00} & \textbf{60}\phantom{.00} & \hmiss & \hmiss \\
\midrule
\multicolumn{2}{@{}l}{Train mean (26 benchmarks, excluding SOON and CaP-X)} & 41.68 & 67.17 & 71.03 & 71.68 & \textbf{73.40} & \hmiss & \hmiss \\
\multicolumn{2}{@{}l}{Train mean (28 benchmarks)} & 39.06 & 63.80 & 67.75 & 70.85 & \textbf{73.16} & \hmiss & \hmiss \\
\multicolumn{2}{@{}l}{Unseen mean (16 benchmarks, excluding SOON and CaP-X)} & 42.13 & 57.72 & 59.57 & 60.75 & \textbf{64.06} & 47.49 & 46.18 \\
\multicolumn{2}{@{}l}{Unseen mean (18 benchmarks)} & 37.45 & 53.53 & 55.17 & 59.56 & \textbf{63.61} & \hmiss & \hmiss \\
\end{longtable}
\end{scriptsize}

%% file: references.bib
@incollection{schmidhuber2007godel,
  title = {{G{\"o}del} Machines: Fully Self-Referential Optimal Universal Self-Improvers},
  author = {Schmidhuber, J{\"u}rgen},
  booktitle = {Artificial General Intelligence},
  editor = {Goertzel, Ben and Pennachin, Cassio},
  series = {Cognitive Technologies},
  pages = {199--226},
  year = {2007},
  publisher = {Springer}
}

@article{achiam2023gpt,
  title = {{GPT-4} Technical Report},
  author = {{OpenAI}},
  journal = {arXiv preprint arXiv:2303.08774},
  year = {2023}
}

@article{team2023gemini,
  title = {{Gemini}: A Family of Highly Capable Multimodal Models},
  author = {{Gemini Team} and Anil, Rohan and Borgeaud, Sebastian and Alayrac, Jean-Baptiste and Yu, Jiahui and Soricut, Radu and Schalkwyk, Johan and Dai, Andrew M. and Hauth, Anja and Millican, Katie and others},
  journal = {arXiv preprint arXiv:2312.11805},
  year = {2023}
}

@inproceedings{alayrac2022flamingo,
  title = {{Flamingo}: a Visual Language Model for Few-Shot Learning},
  author = {Alayrac, Jean-Baptiste and Donahue, Jeff and Luc, Pauline and Miech, Antoine and Barr, Iain and Hasson, Yana and Lenc, Karel and Mensch, Arthur and Millican, Katie and Reynolds, Malcolm and others},
  booktitle = {NeurIPS},
  year = {2022}
}

@inproceedings{liu2023visual,
  title = {Visual Instruction Tuning},
  author = {Liu, Haotian and Li, Chunyuan and Wu, Qingyang and Lee, Yong Jae},
  booktitle = {NeurIPS},
  year = {2023}
}

@inproceedings{madaan2023self,
  title = {{Self-Refine}: Iterative Refinement with Self-Feedback},
  author = {Madaan, Aman and Tandon, Niket and Gupta, Prakhar and Hallinan, Skyler and Gao, Luyu and Wiegreffe, Sarah and Alon, Uri and Dziri, Nouha and Prabhumoye, Shrimai and Yang, Yiming and others},
  booktitle = {NeurIPS},
  year = {2023}
}

@inproceedings{shinn2023reflexion,
  title = {{Reflexion}: Language Agents with Verbal Reinforcement Learning},
  author = {Shinn, Noah and Cassano, Federico and Gopinath, Ashwin and Narasimhan, Karthik and Yao, Shunyu},
  booktitle = {NeurIPS},
  year = {2023}
}

@inproceedings{zhao2024expel,
  title = {{ExpeL}: {LLM} Agents Are Experiential Learners},
  author = {Zhao, Andrew and Huang, Daniel and Xu, Quentin and Lin, Matthieu and Liu, Yong-Jin and Huang, Gao},
  booktitle = {AAAI},
  year = {2024},
  volume = {38},
  number = {17},
  pages = {19632--19642}
}

@article{wang2023voyager,
  title = {{Voyager}: An Open-Ended Embodied Agent with Large Language Models},
  author = {Wang, Guanzhi and Xie, Yuqi and Jiang, Yunfan and Mandlekar, Ajay and Xiao, Chaowei and Zhu, Yuke and Fan, Linxi and Anandkumar, Anima},
  journal = {TMLR},
  year = {2024}
}

@inproceedings{zelikman2023self,
  title = {Self-Taught Optimizer ({STOP}): Recursively Self-Improving Code Generation},
  author = {Zelikman, Eric and Lorch, Eliana and Mackey, Lester and Kalai, Adam Tauman},
  booktitle = {COLM},
  year = {2024}
}

@inproceedings{yin2024g,
  title = {{G{\"o}del Agent}: A Self-Referential Agent Framework for Recursively Self-Improvement},
  author = {Yin, Xunjian and Wang, Xinyi and Pan, Liangming and Lin, Li and Wan, Xiaojun and Wang, William Yang},
  booktitle = {ACL},
  year = {2025},
  pages = {27890--27913}
}

@inproceedings{zhang2026darwin,
  title = {{Darwin G{\"o}del Machine}: Open-Ended Evolution of Self-Improving Agents},
  author = {Zhang, Jenny and Hu, Shengran and Lu, Cong and Lange, Robert and Clune, Jeff},
  booktitle = {ICLR},
  year = {2026}
}

@article{zhang2026self,
  title = {{Self-Harness}: Harnesses That Improve Themselves},
  author = {Zhang, Hangfan and Zhang, Shao and Li, Kangcong and Zhang, Chen and Chen, Yang and Zhang, Yiqun and Bai, Lei and Hu, Shuyue},
  journal = {arXiv preprint arXiv:2606.09498},
  year = {2026}
}

@article{lee2026meta,
  title = {{Meta-Harness}: End-to-End Optimization of Model Harnesses},
  author = {Lee, Yoonho and Nair, Roshen and Zhang, Qizheng and Lee, Kangwook and Khattab, Omar and Finn, Chelsea},
  journal = {arXiv preprint arXiv:2603.28052},
  year = {2026}
}

@article{lin2026agentic,
  title = {Agentic Harness Engineering: Observability-Driven Automatic Evolution of Coding-Agent Harnesses},
  author = {Lin, Jiahang and Liu, Shichun and Pan, Chengjun and Lin, Lizhi and Dou, Shihan and Xi, Zhiheng and Huang, Xuanjing and Yan, Hang and Han, Zhenhua and Gui, Tao and others},
  journal = {arXiv preprint arXiv:2604.25850},
  year = {2026}
}

@inproceedings{hu2025automated,
  title = {Automated Design of Agentic Systems},
  author = {Hu, Shengran and Lu, Cong and Clune, Jeff},
  booktitle = {ICLR},
  year = {2025}
}

@inproceedings{shang2025agentsquare,
  title = {{AgentSquare}: Automatic {LLM} Agent Search in Modular Design Space},
  author = {Shang, Yu and Li, Yu and Zhao, Keyu and Ma, Likai and Liu, Jiahe and Xu, Fengli and Li, Yong},
  booktitle = {ICLR},
  year = {2025}
}

@inproceedings{zhang2025aflow,
  title = {{AFlow}: Automating Agentic Workflow Generation},
  author = {Zhang, Jiayi and Xiang, Jinyu and Yu, Zhaoyang and Teng, Fengwei and Chen, Xiong-Hui and Chen, Jiaqi and Zhuge, Mingchen and Cheng, Xin and Hong, Sirui and Wang, Jinlin and others},
  booktitle = {ICLR},
  year = {2025}
}

@inproceedings{yang2024large,
  title = {Large Language Models as Optimizers},
  author = {Yang, Chengrun and Wang, Xuezhi and Lu, Yifeng and Liu, Hanxiao and Le, Quoc V. and Zhou, Denny and Chen, Xinyun},
  booktitle = {ICLR},
  year = {2024}
}

@inproceedings{fernando2023promptbreeder,
  title = {{Promptbreeder}: Self-Referential Self-Improvement via Prompt Evolution},
  author = {Fernando, Chrisantha and Banarse, Dylan Sunil and Michalewski, Henryk and Osindero, Simon and Rockt{\"a}schel, Tim},
  booktitle = {ICML},
  year = {2024},
  pages = {13481--13544}
}

@article{yuksekgonul2024textgrad,
  title = {{TextGrad}: Automatic ``Differentiation'' via Text},
  author = {Yuksekgonul, Mert and Bianchi, Federico and Boen, Joseph and Liu, Sheng and Huang, Zhi and Guestrin, Carlos and Zou, James},
  journal = {arXiv preprint arXiv:2406.07496},
  year = {2024}
}

@article{wu2026modularrsi,
  title = {{ModularRSI}: Modular and Generalizable Recursive Harness Self-Improvement},
  author = {Wu, Siwei and Ren, Jincheng and Li, Yizhi and Li, Haau-Sing and Yang, Chengran and Ma, Xudong and Zhang, Yuxuan and Gu, Weicheng and Yang, Jian and Batista-Navarro, Riza and others},
  journal = {arXiv preprint arXiv:2609.14857},
  year = {2026}
}

@article{zhang2026jit,
  title = {{JIT-Agent}: Scaling Harness Intelligence via Just-in-Time Harness Evolution},
  author = {Zhang, Guibin and Lu, Leo and Xie, Fangzhou and Zhu, Kang and Wang, Junhao and Xie, Zhifei and Yu, Zhaochen and Liu, Zihang and Sun, Zhongxiang and Li, Qiankun and others},
  journal = {arXiv preprint arXiv:2608.25593},
  year = {2026}
}

@article{kang2026simworld,
  title = {{SimWorld Studio}: Automatic Environment Generation with Evolving Coding Agent for Embodied Agent Learning},
  author = {Kang, Haoqiang and Ye, Xiaokang and Liu, Yuhan and Mantri, Siddhant Hitesh and Mao, Lingjun and Fleming, James and Regmi, Drishti and Qin, Lianhui},
  journal = {arXiv preprint arXiv:2605.09423},
  year = {2026}
}

@inproceedings{zala2024envgen,
  title = {{EnvGen}: Generating and Adapting Environments via {LLMs} for Training Embodied Agents},
  author = {Zala, Abhay and Cho, Jaemin and Lin, Han and Yoon, Jaehong and Bansal, Mohit},
  booktitle = {COLM},
  year = {2024}
}

@inproceedings{xie2024osworld,
  title = {{OSWorld}: Benchmarking Multimodal Agents for Open-Ended Tasks in Real Computer Environments},
  author = {Xie, Tianbao and Zhang, Danyang and Chen, Jixuan and Li, Xiaochuan and Zhao, Siheng and Cao, Ruisheng and Hua, Toh Jing and Cheng, Zhoujun and Shin, Dongchan and Lei, Fangyu and others},
  booktitle = {NeurIPS},
  year = {2024}
}

@article{zuo2026qwen,
  title = {{Qwen-AgentWorld}: Language World Models for General Agents},
  author = {Zuo, Yuxin and Xiao, Zikai and Sheng, Li and Huang, Fei and Tu, Jianhong and Liu, Yuxuan and Tang, Tianyi and Hu, Xiaomeng and Su, Yang and Lan, Qingfeng and others},
  journal = {arXiv preprint arXiv:2606.24597},
  year = {2026}
}

@inproceedings{brant2017minimal,
  title = {Minimal Criterion Coevolution: A New Approach to Open-Ended Search},
  author = {Brant, Jonathan C. and Stanley, Kenneth O.},
  booktitle = {GECCO},
  year = {2017},
  pages = {67--74}
}

@inproceedings{wang2019paired,
  title = {{POET}: Open-Ended Coevolution of Environments and Their Optimized Solutions},
  author = {Wang, Rui and Lehman, Joel and Clune, Jeff and Stanley, Kenneth O.},
  booktitle = {GECCO},
  year = {2019},
  pages = {142--151}
}

@inproceedings{wang2020enhanced,
  title = {Enhanced {POET}: Open-Ended Reinforcement Learning through Unbounded Invention of Learning Challenges and their Solutions},
  author = {Wang, Rui and Lehman, Joel and Rawal, Aditya and Zhi, Jiale and Li, Yulun and Clune, Jeffrey and Stanley, Kenneth},
  booktitle = {ICML},
  year = {2020},
  pages = {9940--9951}
}

@inproceedings{parker2022evolving,
  title = {Evolving Curricula with Regret-Based Environment Design},
  author = {Parker-Holder, Jack and Jiang, Minqi and Dennis, Michael and Samvelyan, Mikayel and Foerster, Jakob and Grefenstette, Edward and Rockt{\"a}schel, Tim},
  booktitle = {ICML},
  year = {2022},
  pages = {17473--17498}
}

@inproceedings{bengio2009curriculum,
  title = {Curriculum Learning},
  author = {Bengio, Yoshua and Louradour, J{\'e}r{\^o}me and Collobert, Ronan and Weston, Jason},
  booktitle = {ICML},
  year = {2009},
  pages = {41--48}
}

@inproceedings{graves2017automated,
  title = {Automated Curriculum Learning for Neural Networks},
  author = {Graves, Alex and Bellemare, Marc G. and Menick, Jacob and Munos, R{\'e}mi and Kavukcuoglu, Koray},
  booktitle = {ICML},
  year = {2017},
  pages = {1311--1320}
}

@inproceedings{brohan2023rt,
  title = {{RT-2}: Vision-Language-Action Models Transfer Web Knowledge to Robotic Control},
  author = {Zitkovich, Brianna and Yu, Tianhe and Xu, Sichun and Xu, Peng and Xiao, Ted and Xia, Fei and Wu, Jialin and Wohlhart, Paul and Welker, Stefan and Wahid, Ayzaan and others},
  booktitle = {CoRL},
  year = {2023},
  pages = {2165--2183}
}

@inproceedings{driess2023palm,
  title = {{PaLM-E}: An Embodied Multimodal Language Model},
  author = {Driess, Danny and Xia, Fei and Sajjadi, Mehdi S. M. and Lynch, Corey and Chowdhery, Aakanksha and Ichter, Brian and Wahid, Ayzaan and Tompson, Jonathan and Vuong, Quan and Yu, Tianhe and others},
  booktitle = {ICML},
  year = {2023},
  pages = {8469--8488}
}

@article{zhou2026hierarchical,
  title = {Hierarchical Self-Improvement: A Framework for Task-Specific Evolvable Agent Harnesses},
  author = {Zhou, Tailin},
  journal = {arXiv preprint arXiv:2608.08466},
  year = {2026}
}

@article{liu2026spade,
  title = {{SPADE}: Self-Play in Adaptive Synthetic Executable Environments},
  author = {Liu, Bo and Yu, Simon and Jiang, Yiding and Qu, Ao and Zhao, Andrew and Liu, Zichen and Kim, Junsu and Zhou, Zijian and Kim, Seungone and Ren, Tongzheng and others},
  journal = {arXiv preprint arXiv:2608.19197},
  year = {2026}
}

@inproceedings{yuan2026pace,
  title = {{PACE}: Parameter Change for Unsupervised Environment Design},
  author = {Yuan, Fang and Zeng, Junjie and Li, Qinglun and Qin, Long and Yin, Quanjun and Shen, Siqi and Xie, Yuxiang and Yang, Junqiang},
  booktitle = {ICML},
  year = {2026},
  pages = {152219--152240}
}

@article{wang2025embodied,
  title = {Embodied Co-Design for Rapidly Evolving Agents: Taxonomy, Frontiers, and Challenges},
  author = {Wang, Yuxing and Chen, Zhiyu and Zhang, Tiantian and Yin, Qiyue and Chang, Yongzhe and Li, Zhiheng and Wang, Liang and Wang, Xueqian},
  journal = {arXiv preprint arXiv:2512.04770},
  year = {2025}
}

@article{bai2025qwen3,
  title = {{Qwen3-VL} Technical Report},
  author = {Bai, Shuai and Cai, Yuxuan and Chen, Ruizhe and Chen, Keqin and Chen, Xionghui and Cheng, Zesen and Deng, Lianghao and Ding, Wei and Gao, Chang and Ge, Chunjiang and others},
  journal = {arXiv preprint arXiv:2511.21631},
  year = {2025}
}

@article{lu2026muse,
  title = {{MUSE}: A Unified Agentic Harness for {MLLMs}},
  author = {Lu, Jianglin and Wang, Hailing and Ma, Xu and Dong, Qihua and Zhang, Mingyuan and Wang, Yizhou and Fu, Yun},
  journal = {arXiv preprint arXiv:2606.03005},
  year = {2026}
}

@article{kang2026harness,
  title = {Harness Continual Learning: Continual Adaptation Beyond Model Parameters},
  author = {Kang, Borui and Gu, Jinrui and Lv, Junhan and Li, Wenbin and Wang, Lei and Gao, Yang},
  journal = {arXiv preprint arXiv:2608.19013},
  year = {2026}
}

@article{wu2026harnessdev,
  title = {{HarnessDev}: Can {LLMs} Create and Evolve Their Own Agent Harness?},
  author = {Wu, Yuhao and Zhang, Jingyuan and Shi, Jiajun and Lei, Xinping and Gu, Qingshui and Zhang, Yuxuan and Wang, Zexuan and He, Chen and Huang, Chen and Song, Maojia and others},
  journal = {arXiv preprint arXiv:2609.01437},
  year = {2026}
}

@article{tan2026metarsi,
  title = {{MetaRSI} / {RSI2}: A Meta-Recursive Self-Improving System for Recursive Self-Improving Systems Themselves},
  author = {Tan, Zihan and Sun, Leixin and Shi, Zitong and Liu, Yitao and Wu, Jiajun and Brooks, Nathaniel and Qian, Jiaru and Shang, Xiaoran and Huang, Suyuan and Ding, Yi and others},
  journal = {arXiv preprint arXiv:2609.06396},
  year = {2026}
}

@article{wang2026metaskill,
  title = {{MetaSkill-Evolve}: Recursive Self-Improvement of {LLM} Agents via Two-Timescale Meta-Skill Evolution},
  author = {Wang, Zefeng and Yan, Minxi and Bi, Jinhe and Yan, Sikuan and Tresp, Volker and Ma, Yunpu},
  journal = {arXiv preprint arXiv:2607.05297},
  year = {2026}
}

@article{jia2026gamegen,
  title = {{GameGen-Verifier}: Parallel Keypoint-Based Verification for {LLM}-Generated Games via Runtime State Injection},
  author = {Jia, Chaobo and Wan, Ruipeng and Sun, Ting and Tan, Weihao and Wan, Borui and Tong, Yuxuan and Sheng, Guangming and Xu, Hong},
  journal = {arXiv preprint arXiv:2605.07442},
  year = {2026}
}

@article{dong2026agent,
  title = {{Agent-World}: Scaling Real-World Environment Synthesis for Evolving General Agent Intelligence},
  author = {Dong, Guanting and Lu, Junting and Huang, Junjie and Zhong, Wanjun and Liu, Longxiang and Huang, Shijue and Li, Zhenyu and Zhao, Yang and Song, Xiaoshuai and Li, Xiaoxi and others},
  journal = {arXiv preprint arXiv:2604.18292},
  year = {2026}
}

@article{lee2026recursive,
  title = {Recursive Harness Self-Improvement},
  author = {Lee, Hyunin and Xu, Jinglue and Seely, Jeffrey and Lee, Donghyun and Zaharia, Matei and Tang, Yujin},
  journal = {arXiv preprint arXiv:2607.15524},
  year = {2026}
}

@article{hurst2024gpt,
  title = {{GPT-4o} System Card},
  author = {{OpenAI}},
  journal = {arXiv preprint arXiv:2410.21276},
  year = {2024}
}

@article{singh2025openai,
  title = {{OpenAI GPT-5} System Card},
  author = {{OpenAI}},
  journal = {arXiv preprint arXiv:2601.03267},
  year = {2025}
}

@article{lou2026autoharness,
  title = {{AutoHarness}: Improving {LLM} Agents by Automatically Synthesizing a Code Harness},
  author = {Lou, Xinghua and L{\'a}zaro-Gredilla, Miguel and Dedieu, Antoine and Wendelken, Carter and Lehrach, Wolfgang and Murphy, Kevin P.},
  journal = {arXiv preprint arXiv:2603.03329},
  year = {2026}
}

@article{chen2026harnessx,
  title = {{HarnessX}: A Composable, Adaptive, and Evolvable Agent Harness Foundry},
  author = {Chen, Tingyang and Lu, Shuo and Zhao, Kang and Meng, Weicheng and Teng, Hanlin and Li, Tianhao and Li, Chao and Liu, Xule and Liang, Jian and Zhang, Zhizhong and others},
  journal = {arXiv preprint arXiv:2606.14249},
  year = {2026}
}

@article{shao2026harness,
  title = {{Harness-R1}: Learning to Edit Executable Runtime Harnesses from Agent Failure Trajectories},
  author = {Shao, Shuai and Zhang, Kangning and Li, Qingyao and Wang, Shijian and Wang, Hao and Jiao, Wenxiang and Lu, Yuan and Guo, Yi and Liu, Weiwen and Zhang, Weinan},
  journal = {arXiv preprint arXiv:2608.02276},
  year = {2026}
}

@article{chen2026failed,
  title = {From Failed Trajectories to Reliable {LLM} Agents: Diagnosing and Repairing Harness Flaws},
  author = {Chen, Mengzhuo and Wang, Junjie and Liu, Zhe and Wang, Yawen and Zheng, Haiming and Wang, Qing},
  journal = {arXiv preprint arXiv:2606.06324},
  year = {2026}
}

@article{liu2026adaptive,
  title = {{Adaptive Auto-Harness}: Sustained Self-Improvement for Agentic System Deployment on Open-Ended Task Streams},
  author = {Liu, Zewen and Shi, Zhan and Sang, Yisi and He, Bing and Lin, Minhua and Wei, Tianxin and Wang, Dakuo and Dumoulin, Benoit and Jin, Wei and Lu, Hanqing},
  journal = {arXiv preprint arXiv:2606.01770},
  year = {2026}
}

@article{nie2026tthe,
  title = {{TTHE}: Test-Time Harness Evolution},
  author = {Nie, Jun and Zhang, Yonggang and Song, Jun and Cai, Qianshu and Yu, Dahai and Guo, Yike and Tian, Xinmei and Han, Bo},
  journal = {arXiv preprint arXiv:2607.08124},
  year = {2026}
}

@article{yao2026harness,
  title = {{Harness-Bench}: Evaluating Agent Harnesses across Models and Realistic Workflows},
  author = {Yao, Yilun and Tan, Xinyu and Liu, Chao-Hsuan and Li, Yaoming and Wang, Zhengyang and Lv, Zongwei and Yu, Wenhan and Tan, Zhewen and Tian, Yuxuan and Sun, Lin and others},
  journal = {arXiv preprint arXiv:2605.27922},
  year = {2026}
}

@article{ren2026self,
  title = {Self-Improvements in Modern Agentic Systems: A Survey},
  author = {Ren, Zhe and Chen, Yimeng and Guo, Dandan and Rong, Guowei and Li, Tonghui and Xiong, R. B. and Lan, Qingfeng and Wang, Wenyi and Li, Nanbo and Yang, Yibo and others},
  journal = {arXiv preprint arXiv:2607.13104},
  year = {2026}
}

@article{hu2026seal,
  title = {{SEAL}: Synergistic Co-Evolution of Agents and Learning Environments},
  author = {Hu, Yihao and Wen, Zhihao and Liu, Xiujin and Wang, Pan and Zhang, Xin and Wu, Wei},
  journal = {arXiv preprint arXiv:2605.24426},
  year = {2026}
}

@article{liu2026escher,
  title = {{Escher-Loop}: Mutual Evolution by Closed-Loop Self-Referential Optimization},
  author = {Liu, Ziyang and Guo, Xinyan and Wei, Xuchen and Hao, Han and Yang, Liu},
  journal = {arXiv preprint arXiv:2604.23472},
  year = {2026}
}

@article{chen2026evotrainer,
  title = {{EvoTrainer}: Co-Evolving {LLM} Policies and Training Harnesses for Autonomous Agentic Reinforcement Learning},
  author = {Chen, Guhong and Shi, Yingcheng and Li, Yongbin and Li, Binhua and Xu, Xander and Wei, Hu and Ni, Shiwen and Yang, Min and Ye, Jieping},
  journal = {arXiv preprint arXiv:2606.03108},
  year = {2026}
}

@article{gandhi2026endless,
  title = {Endless Terminals: Scaling {RL} Environments for Terminal Agents},
  author = {Gandhi, Kanishk and Garg, Shivam and Goodman, Noah D. and Papailiopoulos, Dimitris},
  journal = {arXiv preprint arXiv:2601.16443},
  year = {2026}
}

@article{cheng2026terminal,
  title = {{Terminal-World}: Scaling Terminal-Agent Environments via Agent Skills},
  author = {Cheng, Zihao and Wang, Hongru and Liu, Zeming and Wang, Xinyi and Zhu, Xiangrong and Guo, Yuhang and Lin, Wei and Pan, Jeff Z. and Wang, Yunhong},
  journal = {arXiv preprint arXiv:2605.20876},
  year = {2026}
}

@article{hebbar2026sia,
  title = {{SIA}: Self Improving {AI} with Harness \& Weight Updates},
  author = {Hebbar, Prannay and Manawat, Yogendra and Verboomen, Samuel and Ivanova, Alesia and Palanimalai, Selvam and Bhatia, Kunal and Baskaran, Vignesh},
  journal = {arXiv preprint arXiv:2605.27276},
  year = {2026}
}

@article{zhou2026agentrouter,
  title = {{Agent-as-a-Router}: Agentic Model Routing for Coding Tasks},
  author = {Zhou, Pengfei and Tang, Zhiwei and Ma, Yixing and Tang, Jiasheng and Han, Yizeng and Wan, Zhenglin and Meng, Fanqing and Wang, Wei and Zhuang, Bohan and Zhao, Wangbo and others},
  journal = {arXiv preprint arXiv:2606.22902},
  year = {2026}
}

@article{arcprize2026arcagi3,
  title = {{ARC-AGI-3}: A New Challenge for Frontier Agentic Intelligence},
  author = {{ARC Prize Foundation}},
  journal = {arXiv preprint arXiv:2603.24621},
  year = {2026}
}

@misc{ukmarsMicromouseRules,
  title = {{UK Micromouse Classic Rules}},
  author = {{United Kingdom Micromouse and Robotics Society (UKMARS)}},
  howpublished = {\url{https://ukmars.org/contests/contest-rules/micromouse-classic/}},
  note = {Accessed 2026-10-26}
}

@misc{openai2025codex,
  title = {{Codex CLI}: Lightweight Coding Agent That Runs in Your Terminal},
  author = {{OpenAI}},
  year = {2025},
  howpublished = {\url{https://github.com/openai/codex}},
  note = {Accessed 2026-10-26}
}

@misc{deepseek2026harness,
  title = {{DeepSeek Harness}: Everything is a Plugin},
  author = {{DeepSeek-AI}},
  year = {2026},
  howpublished = {\url{https://github.com/deepseek-ai/deepseek-harness}},
  note = {Accessed 2026-10-26}
}

@inproceedings{paglieri2025balrog,
  title = {{BALROG}: Benchmarking Agentic {LLM} and {VLM} Reasoning on Games},
  author = {Paglieri, Davide and Cupia{\l}, Bart{\l}omiej and Coward, Samuel and Piterbarg, Ulyana and Wolczyk, Maciej and Khan, Akbir and Pignatelli, Eduardo and Kuci{\'n}ski, {\L}ukasz and Pinto, Lerrel and Fergus, Rob and others},
  booktitle = {ICLR},
  year = {2025}
}

@inproceedings{bakhtin2019phyre,
  title = {{PHYRE}: A New Benchmark for Physical Reasoning},
  author = {Bakhtin, Anton and van der Maaten, Laurens and Johnson, Justin and Gustafson, Laura and Girshick, Ross},
  booktitle = {NeurIPS},
  year = {2019}
}

@article{battleday2026digbench,
  title = {{DiG-bench}: Discovery in Games},
  author = {Battleday, Ruairidh M. and Sandbrink, Kai and Cullen-Drohan, Jimi and Yan, Zihan and Muller, Timothy and Maguire, Clare and Kubicek, Ales and Greenlee-Scott, Fraser and Sumant, Sukrit and Dao, Tri and others},
  journal = {arXiv preprint arXiv:2608.12593},
  year = {2026}
}

@article{imajuku2025alebench,
  title = {{ALE-Bench}: A Benchmark for Long-Horizon Objective-Driven Algorithm Engineering},
  author = {Imajuku, Yuki and Horie, Kohki and Iwata, Yoichi and Aoki, Kensho and Takahashi, Naohiro and Akiba, Takuya},
  journal = {arXiv preprint arXiv:2506.09050},
  year = {2025}
}

@article{fu2026videommev2,
  title = {{Video-MME-v2}: Towards the Next Stage in Benchmarks for Comprehensive Video Understanding},
  author = {Fu, Chaoyou and Yuan, Haozhi and Dong, Yuhao and Zhang, Yi-Fan and Shen, Yunhang and Hu, Xiaoxing and Li, Xueying and Su, Jinsen and Long, Chengwu and Xie, Xiaoyao and others},
  journal = {arXiv preprint arXiv:2604.05015},
  year = {2026}
}

@inproceedings{xia2026buildarena,
  title = {{BuildArena}: A Physics-Aligned Interactive Benchmark of {LLMs} for Engineering Construction},
  author = {Xia, Tian and Gao, Tianrun and Deng, Wenhao and Wei, Long and Qian, Xiaowei and Yu, Chenglei and Wu, Tailin},
  booktitle = {ICML},
  year = {2026}
}

@inproceedings{todorov2012mujoco,
  title = {{MuJoCo}: A Physics Engine for Model-Based Control},
  author = {Todorov, Emanuel and Erez, Tom and Tassa, Yuval},
  booktitle = {IEEE/RSJ International Conference on Intelligent Robots and Systems (IROS)},
  year = {2012},
  pages = {5026--5033}
}

@misc{golden2020eleusisexpress,
  title = {{Eleusis Express}},
  author = {Golden, John},
  year = {2020},
  howpublished = {Gathering 4 Gardner, \url{https://www.gathering4gardner.org/wp-content/uploads/2020/04/G4G14-EleusisExpress-Instructions.pdf}},
  note = {Express variation of Robert Abbott's Eleusis; accessed 2026-10-26}
}

@misc{moyersoen2023saboteur,
  title = {{Saboteur}: Rules of Play, 20 Year Anniversary Edition},
  author = {Moyersoen, Frederic},
  year = {2023},
  howpublished = {AMIGO Spiel + Freizeit GmbH, \url{https://blog.amigo-spiele.de/content/ap/rule/02402-GB-AmigoRule.pdf}},
  note = {Accessed 2026-10-26}
}

@techreport{euroncap2023aeb,
  title = {Test Protocol -- {AEB} Car-to-Car Systems},
  author = {{Euro NCAP}},
  institution = {European New Car Assessment Programme},
  year = {2023},
  month = dec,
  number = {Version 4.3},
  url = {https://cdn.euroncap.com/cars/assets/euro_ncap_aeb_c2c_test_protocol_v43_1e6ed06def.pdf}
}

@misc{wca2025regulations,
  title = {{WCA} Regulations},
  author = {{World Cube Association}},
  year = {2025},
  howpublished = {\url{https://www.worldcubeassociation.org/regulations/}},
  note = {Release of 17 July 2025; accessed 2026-10-26}
}

@misc{hinterland2017longdark,
  title = {{The Long Dark}},
  author = {{Hinterland Studio}},
  year = {2017},
  howpublished = {Video game, \url{https://www.thelongdark.com/}},
  note = {Accessed 2026-10-26}
}

@misc{lichess2026puzzles,
  title = {Lichess Open Puzzle Database},
  author = {{Lichess}},
  year = {2026},
  howpublished = {\url{https://database.lichess.org/#puzzles}},
  note = {Accessed 2026-10-26}
}

@book{pulford2007locks,
  title = {High-Security Mechanical Locks: An Encyclopedic Reference},
  author = {Pulford, Graham W.},
  publisher = {Butterworth-Heinemann},
  year = {2007},
  isbn = {9780750684378}
}

@techreport{en15635,
  title = {{EN 15635:2008} -- Steel Static Storage Systems -- Application and Maintenance of Storage Equipment},
  author = {{European Committee for Standardization}},
  institution = {CEN},
  year = {2009},
  note = {BS EN 15635:2008, British Standards Institution}
}

@inproceedings{liu2026adaplanbench,
  title = {{AdaPlanBench}: Evaluating Adaptive Planning in Large Language Model Agents under World and User Constraints},
  author = {Liu, Jiayu and Qian, Cheng and Wang, Zhenhailong and Li, Bingxuan and Liu, Jiateng and Zong, Qing and Wang, Heng and Kim, Jeonghwan and Wang, Yumeng and He, Bingxiang and others},
  booktitle = {COLM},
  year = {2026}
}

@inproceedings{golechha2025amongus,
  title = {{Among Us}: A Sandbox for Measuring and Detecting Agentic Deception},
  author = {Golechha, Satvik and Garriga-Alonso, Adri{\`a}},
  booktitle = {NeurIPS},
  year = {2025}
}

@article{gandhi2025boxinggym,
  title = {{BoxingGym}: Benchmarking Progress in Automated Experimental Design and Model Discovery},
  author = {Gandhi, Kanishk and Li, Michael Y. and Goodyear, Lyle and Bhatia, Agam and Li, Louise and Bhaskar, Aditi and Zaman, Mohammed and Goodman, Noah D.},
  journal = {arXiv preprint arXiv:2501.01540},
  year = {2025}
}

@article{stephenson2025codenames,
  title = {Codenames as a Benchmark for Large Language Models},
  author = {Stephenson, Matthew and Sidji, Matthew and Ronval, Beno{\^i}t},
  journal = {IEEE Transactions on Games},
  year = {2025},
  pages = {1--12}
}

@article{guertler2025textarena,
  title = {{TextArena}},
  author = {Guertler, Leon and Cheng, Bobby and Yu, Simon and Liu, Bo and Choshen, Leshem and Tan, Cheston},
  journal = {arXiv preprint arXiv:2504.11442},
  year = {2025}
}

@article{wiemann2026discoverphysics,
  title = {{DiscoverPhysics}: Benchmarking {LLMs} for Out-of-the-Box Scientific Thinking},
  author = {Wiemann, Matt L. and Smith, Lindsay M. and Melchior, Peter and Mishra-Sharma, Siddharth and Wilson, Andrew Gordon and Izmailov, Pavel and Cuesta-L{\'a}zaro, Carolina},
  journal = {arXiv preprint arXiv:2605.26087},
  year = {2026}
}

@article{shapira2024glee,
  title = {{GLEE}: A Unified Framework and Benchmark for Language-based Economic Environments},
  author = {Shapira, Eilam and Madmon, Omer and Reinman, Itamar and Amouyal, Samuel Joseph and Reichart, Roi and Tennenholtz, Moshe},
  journal = {arXiv preprint arXiv:2410.05254},
  year = {2024}
}

@article{golde2025mastermindeval,
  title = {{MastermindEval}: A Simple But Scalable Reasoning Benchmark},
  author = {Golde, Jonas and Haller, Patrick and Barth, Fabio and Akbik, Alan},
  journal = {arXiv preprint arXiv:2503.05891},
  year = {2025}
}

@article{pasukonis2022memorymaze,
  title = {Evaluating Long-Term Memory in {3D} Mazes},
  author = {Pasukonis, Jurgis and Lillicrap, Timothy and Hafner, Danijar},
  journal = {arXiv preprint arXiv:2210.13383},
  year = {2022}
}

@article{jiang2026pinverify,
  title = {{PInVerify}: An Offline Embodied Benchmark for Active Instance Verification},
  author = {Jiang, Yuhang},
  journal = {arXiv preprint arXiv:2605.30639},
  year = {2026}
}

@article{xu2025deepphy,
  title = {{DeepPHY}: Benchmarking Agentic {VLMs} on Physical Reasoning},
  author = {Xu, Xinrun and Bu, Pi and Wang, Ye and Karlsson, B{\"o}rje F. and Wang, Ziming and Song, Tengtao and Zhu, Qi and Song, Jun and Ding, Zhiming and Zheng, Bo},
  journal = {arXiv preprint arXiv:2508.05405},
  year = {2025}
}

@article{kiefl2024pooltool,
  title = {Pooltool: A {Python} Package for Realistic Billiards Simulation},
  author = {Kiefl, Evan},
  journal = {Journal of Open Source Software},
  year = {2024},
  volume = {9},
  number = {101},
  pages = {7301}
}

@misc{robx2026pzprjs,
  title = {pzprjs: Playing and Editing Puzzles by {JavaScript}},
  author = {{robx}},
  year = {2026},
  howpublished = {\url{https://github.com/robx/pzprjs}},
  note = {Accessed 2026-10-26}
}

@misc{penpa2026,
  title = {{Penpa+}: Web Application to Create and Solve Sudokus and Puzzles},
  author = {{swaroopg92}},
  year = {2026},
  howpublished = {\url{https://github.com/swaroopg92/penpa-edit}},
  note = {Accessed 2026-10-26}
}

@misc{lavelle2026puzzlescript,
  title = {{PuzzleScript}: Open Source {HTML5} Puzzle Game Engine},
  author = {Lavelle, Stephen},
  year = {2026},
  howpublished = {\url{https://github.com/increpare/PuzzleScript}},
  note = {Accessed 2026-10-26}
}

@article{phan2025textquests,
  title = {{TextQuests}: How Good are {LLMs} at Text-Based Video Games?},
  author = {Phan, Long and Mazeika, Mantas and Zou, Andy and Hendrycks, Dan},
  journal = {arXiv preprint arXiv:2507.23701},
  year = {2025}
}

@article{song2026videoindex,
  title = {{Video-Index}: A Curated Meta-Benchmark for Video Understanding},
  author = {Song, Enxin and Xu, Yinuo and Yang, Shusheng and Chai, Wenhao and Gu, Jiatao},
  journal = {arXiv preprint arXiv:2610.00960},
  year = {2026}
}

@article{banatt2024wilt,
  title = {{WILT}: A Multi-Turn, Memorization-Robust Inductive Logic Benchmark for {LLMs}},
  author = {Banatt, Eryk and Cheng, Jonathan and Vaidyanath, Skanda and Hwu, Tiffany},
  journal = {arXiv preprint arXiv:2410.10998},
  year = {2024}
}

@inproceedings{zhu2021soon,
  title = {{SOON}: Scenario Oriented Object Navigation with Graph-based Exploration},
  author = {Zhu, Fengda and Liang, Xiwen and Zhu, Yi and Yu, Qizhi and Chang, Xiaojun and Liang, Xiaodan},
  booktitle = {CVPR},
  year = {2021},
  pages = {12689--12699}
}

@inproceedings{fu2026capx,
  title = {{CaP-X}: A Framework for Benchmarking and Improving Coding Agents for Robot Manipulation},
  author = {Fu, Letian and Yu, Justin and El-Refai, Karim and Kou, Ethan and Xue, Haoru and Huang, Huang and Xiao, Wenli and Fei-Fei, Li and Shi, Guanya and Wu, Jiajun and others},
  booktitle = {ICML},
  year = {2026}
}

@inproceedings{wang2025actiview,
  title = {{ActiView}: Evaluating Active Perception Ability for Multimodal Large Language Models},
  author = {Wang, Ziyue and Chen, Chi and Luo, Fuwen and Dong, Yurui and Zhang, Yuanchi and Xu, Yuzhuang and Wang, Xiaolong and Li, Peng and Liu, Yang},
  booktitle = {ACL},
  year = {2025},
  pages = {7605--7633}
}

@inproceedings{hafner2022crafter,
  title = {Benchmarking the Spectrum of Agent Capabilities},
  author = {Hafner, Danijar},
  booktitle = {ICLR},
  year = {2022}
}

@inproceedings{zhang2024eda,
  title = {Probing the Multi-turn Planning Capabilities of {LLMs} via 20 Question Games},
  author = {Zhang, Yizhe and Lu, Jiarui and Jaitly, Navdeep},
  booktitle = {ACL},
  year = {2024},
  pages = {1495--1516}
}

@article{bard2020hanabi,
  title = {The {Hanabi} Challenge: A New Frontier for {AI} Research},
  author = {Bard, Nolan and Foerster, Jakob N. and Chandar, Sarath and Burch, Neil and Lanctot, Marc and Song, H. Francis and Parisotto, Emilio and Dumoulin, Vincent and Moitra, Subhodeep and Hughes, Edward and others},
  journal = {Artificial Intelligence},
  year = {2020},
  volume = {280},
  pages = {103216}
}

@misc{deepmind2026gamearena,
  title = {{Game Arena} Harness},
  author = {{Google DeepMind}},
  year = {2026},
  howpublished = {\url{https://github.com/google-deepmind/game_arena}},
  note = {Accessed 2026-10-26}
}

@article{lanctot2019openspiel,
  title = {{OpenSpiel}: A Framework for Reinforcement Learning in Games},
  author = {Lanctot, Marc and Lockhart, Edward and Lespiau, Jean-Baptiste and Zambaldi, Vinicius and Upadhyay, Satyaki and P{\'e}rolat, Julien and Srinivasan, Sriram and Timbers, Finbarr and Tuyls, Karl and Omidshafiei, Shayegan and others},
  journal = {arXiv preprint arXiv:1908.09453},
  year = {2019}
}

@inproceedings{samvelyan2021minihack,
  title = {{MiniHack} the Planet: A Sandbox for Open-Ended Reinforcement Learning Research},
  author = {Samvelyan, Mikayel and Kirk, Robert and Kurin, Vitaly and Parker-Holder, Jack and Jiang, Minqi and Hambro, Eric and Petroni, Fabio and K{\"u}ttler, Heinrich and Grefenstette, Edward and Rockt{\"a}schel, Tim},
  booktitle = {NeurIPS Datasets and Benchmarks Track},
  year = {2021}
}

@inproceedings{kuttler2020nethack,
  title = {The {NetHack} Learning Environment},
  author = {K{\"u}ttler, Heinrich and Nardelli, Nantas and Miller, Alexander H. and Raileanu, Roberta and Selvatici, Marco and Grefenstette, Edward and Rockt{\"a}schel, Tim},
  booktitle = {NeurIPS},
  year = {2020},
  pages = {7671--7684}
}

@article{waugh2026ppbench,
  title = {{Pencil Puzzle Bench}: A Benchmark for Multi-Step Verifiable Reasoning},
  author = {Waugh, Justin},
  journal = {arXiv preprint arXiv:2603.02119},
  year = {2026}
}

@inproceedings{duan2025scigym,
  title = {Measuring Scientific Capabilities of Language Models with a Systems Biology Dry Lab},
  author = {Duan, Haonan and Lu, Stephen Zhewen and Harrigan, Caitlin Fiona and Desai, Nishkrit and Lu, Jiarui and Koziarski, Micha{\l} and Cotta, Leonardo and Maddison, Chris J.},
  booktitle = {NeurIPS Datasets and Benchmarks Track},
  year = {2025}
}

@misc{anthropic2026opus5,
  title = {System Card: {Claude Opus 5}},
  author = {{Anthropic}},
  year = {2026},
  month = jul,
  howpublished = {\url{https://anthropic.com/claude-opus-5-system-card}},
  note = {Accessed 2026-10-26}
}

@inproceedings{shi2026codehacker,
  title     = {{CodeHacker}: Automated Test Case Generation for Detecting Vulnerabilities in Competitive Programming Solutions},
  author    = {Shi, Jingwei and Yin, Xinxiang and Huang, Jing and Tao, Shengyu and Zhao, Jinman},
  booktitle = {ACL},
  year      = {2026},
  pages     = {2352--2382},
  doi       = {10.18653/v1/2026.acl-long.108}
}

@misc{zhang2026valg,
  title         = {{VALG}: An Agentic System for ML Theory Research},
  author        = {Zhang, Dechen and Tang, Xuan and Yin, Xinxiang and Chen, Xingwu and Qian, Jian and Zou, Difan},
  year          = {2026},
  eprint        = {2608.13060},
  archivePrefix = {arXiv},
  primaryClass  = {cs.AI}
}

@misc{cui2026earthverse,
  title         = {{EarthVerse}: Benchmarking Scientific Agents Across Dynamic Earth Systems and Natural Hazards},
  author        = {Cui, Zhiqing and Yin, Xinxiang and Tang, Yihong and Zhang, Xinglang and Hu, Yuanzhe and Zhong, Siru and Tang, Weidong and Liang, Yuxuan and Li, Weijia and Jin, Ming and Pan, Shirui and Kang, Yuhao and Zhuang, Dingyi and Zhao, Jinhua},
  year          = {2026},
  eprint        = {2608.23525},
  archivePrefix = {arXiv},
  primaryClass  = {cs.AI}
}

@InProceedings{Tang_2026_CVPR,
    author    = {Tang, Weidong and Liang, Zhiyuan and Wan, Xinyan and Zhu, Chen and Xu, Zhaopan and Zhou, Pengfei and Song, Yan and You, Yang and Zhao, Wangbo},
    title     = {Efficient Video Object Segmentation and Tracking with Recurrent Dynamic Submodel},
    booktitle = {CVPR},
    month     = {June},
    year      = {2026},
    pages     = {20912-20921}
}

@inproceedings{tang2026grouptom,
  title     = {{GroupToM-Bench}: Benchmarking Group Theory of Mind and Nonlinear Social Emergence in {MLLMs}},
  author    = {Tang, Weidong and Li, Jierui and Hou, Yueling and Mei, Zihan and Zhang, Can and Wan, Xinyan and Liang, Zhiyuan and Zhou, Pengfei and You, Yang and Zhao, Wangbo},
  booktitle = {ACL},
  year      = {2026},
  pages     = {40007--40031},
  doi       = {10.18653/v1/2026.acl-long.1859}
}

@inproceedings{tang2026varscaling,
  title     = {Inference-Time Scaling for Visual AutoRegressive Modeling by Searching Representative Samples},
  author    = {Tang, Weidong and Wan, Xinyan and Li, Siyu and Wang, Xiumei},
  booktitle = {PRCV},
  year      = {2026},
  pages     = {406--419},
  doi       = {10.1007/978-981-95-5699-1_28}
}

@article{liu2026world,
  title={World Action Learning via Interaction-Centric Spectral Latent Guidance},
  author={Liu, Zhiming and Miao, Yikun and Chen, Ying and Yin, Hongrui and Zhu, Fangqi and Pang, Xiaoyi and Shou, Quanxin and Yan, Zhengyang and Wang, Haodong and Guo, Song},
  journal={arXiv preprint arXiv:2610.03607},
  year={2026}
}

@article{liu2026all,
  title={Do All Individual Layers Help? An Empirical Study of Task-Interfering Layers in Vision-Language Models},
  author={Liu, Zhiming and Wei, Yujie and Feng, Lei and Su, Xiu and Xia, Xiaobo and Guan, Weili and Xie, Zeke and Yang, Shuo},
  journal={arXiv preprint arXiv:2602.01167},
  year={2026}
}

@inproceedings{li2026coco,
  title={{{CoCo}}: Code as {CoT} for Text-to-Image Preview and Rare Concept Generation},
  author={Li, Haodong and Qing, Chunmei and Zhang, Huanyu and Jiang, Dongzhi and Zou, Yihang and Peng, Hongbo and Li, Dingming and Dai, Yuhong and Lin, ZePeng and Tian, Juanxi and others},
  booktitle={ECCV},
  pages={57--75},
  year={2026}
}

@article{li2026videococo,
  title={{{VideoCoCo}}: Code-as-{CoT} for Physically-Consistent Video Generation via an Agentic Dual-Engine System},
  author={Li, Haodong and Ren, Tianfei and Ma, Xiaoxiao and Qing, Chunmei and Fang, Zhen and He, Sipeng and Guo, Ziyu and Wu, Haoyu and Tian, Juanxi and Zou, Yihang and others},
  journal={arXiv preprint arXiv:2607.27380},
  year={2026}
}
